\documentclass[10pt,twocolumn,letterpaper]{article}

\usepackage[pagenumbers]{cvpr}      %

\usepackage[dvipsnames, table]{xcolor}
\usepackage{times}
\usepackage{epsfig}
\usepackage{graphicx}
\usepackage{amsmath}
\usepackage{amssymb}
\usepackage{amsthm}
\usepackage{mathtools}
\usepackage{bm}
\makeatletter
\@namedef{ver@everyshi.sty}{}
\makeatother
\usepackage[mode=buildnew]{standalone}
\usepackage{tikz}
\usetikzlibrary{positioning}
\usetikzlibrary{shapes} 
\usetikzlibrary{matrix}
\usetikzlibrary{calc} 
\usetikzlibrary{intersections}
\usepackage[accsupp]{axessibility}  %
\usepackage{tikzscale}
\usepackage{pgfplots}
\usepackage[eulergreek]{sansmath}
\usepackage{caption,subcaption}
\usepackage{bbm}
\usepackage{comment}
\usepackage{acro}
\usepackage{listings}
\usepackage{multirow}
\usepackage{tabularx}
\usepackage{booktabs}
\usepackage{wrapfig}
\usepackage[vlined]{algorithm2e}
\SetAlCapNameFnt{\small}
\SetAlCapFnt{\small}
\usepackage{float}
\usepackage{overpic}

\usepackage{cuted}
\usepackage{pbox}

\usepackage[pagebackref=true,breaklinks=true,letterpaper=true,colorlinks,bookmarks=false]{hyperref}
\usepackage[capitalise]{cleveref}
\crefname{section}{Sec.}{Secs.}
\Crefname{section}{Section}{Sections}
\Crefname{table}{Table}{Tables}
\crefname{table}{Tab.}{Tabs.}
\crefname{figure}{Fig.}{Figs.}

\newtheorem{theorem}{Theorem}

\newtheorem{definition}[theorem]{Definition}

\renewcommand{\paragraph}[1]{\textbf{#1}}

\usepackage{enumitem}
\setitemize{noitemsep,topsep=0pt,parsep=0pt,partopsep=0pt,leftmargin=1em}
\setenumerate{noitemsep,topsep=0pt,parsep=0pt,partopsep=0pt,leftmargin=1em}

\usepackage{graphicx}
\usepackage{amsmath}
\usepackage{amssymb}
\usepackage{booktabs}
\usepackage{tikz-qtree}

 \renewcommand{\subsubsection}[1]{\paragraph{#1.}}

\def\cvprPaperID{5477} %
\def\confName{CVPR}
\def\confYear{2023}

\newcommand{\verts}{\mathcal{V}}
\newcommand{\edges}{\mathcal{E}}
\newcommand{\arbitraryShape}{\mathcal{X}}
\newcommand{\contour}{\mathcal{M}}
\newcommand{\contourVerts}{\verts_\contour}
\newcommand{\contourVert}{i}
\newcommand{\contourEdgs}{\edges_\contour}
\newcommand{\mesh}{\mathcal{N}}
\newcommand{\meshVerts}{\verts_\mesh}
\newcommand{\meshVert}{j}
\newcommand{\meshEdgs}{\edges_\mesh}
\newcommand{\extEdgs}{\edges^+}
\newcommand{\prodGraph}{\mathcal{P}}
\newcommand{\prodVerts}{\mathcal{V}}
\newcommand{\prodEdgs}{\mathcal{E}}

\newcommand{\conjProdGraph}{\mathcal{P}^*}
\newcommand{\conjProdVerts}{\mathcal{V}^*}
\newcommand{\conjProdEdgs}{\mathcal{E}^*}

\newcommand{\graph}{\mathcal{G}}
\newcommand{\graphVerts}{\mathcal{V}_\graph}
\newcommand{\graphEdgs}{\mathcal{E}_\graph}
\newcommand{\conjGraph}{\mathcal{G}^*}
\newcommand{\conjGraphVerts}{\mathcal{V}_\conjGraph}
\newcommand{\conjGraphEdgs}{\mathcal{E}_\conjGraph}
\newcommand{\branchSet}{\mathcal{B}}
\newcommand{\laehner}{L{\"a}hner}
\newcommand{\laehneretal}{\laehner~\etal}

\definecolor{mycolor1}{RGB}{90,130,213}%
\definecolor{mycolor2}{RGB}{230,130,46}%
\definecolor{mycolor3}{RGB}{231,92,46}%
\definecolor{mycolor11}{RGB}{134,168,235}%
\definecolor{mycolor33}{RGB}{231,156,130}%
\definecolor{cBLUE}{RGB}{90,130,213}%
\definecolor{cBLUE1}{RGB}{90,183,214}%
\definecolor{cBLUE2}{RGB}{132,217,226}%
\definecolor{cRED}{RGB}{231,92,46}%
\definecolor{cPINK}{RGB}{200,57,170}%
\definecolor{cPINKLIGHT}{RGB}{255,209,245}%
\definecolor{cGREEN}{RGB}{80,150,80}%
\definecolor{cYELLOW}{RGB}{247,179,43}%
\definecolor{cORANGE}{RGB}{242,105,0}%
\definecolor{cGRAY}{RGB}{129,141,146}%

\begin{document}

\title{Conjugate Product Graphs for Globally Optimal 2D-3D Shape Matching}

\author{Paul Roetzer$^{1}$ $\qquad$ Zorah L{\"a}hner$^2$ $\qquad$ Florian Bernard$^1$\\
	University of Bonn$^1$ $\qquad$ University of Siegen$^2$ 
}
\maketitle

\newcommand{\draftFigs}{false}
\newcommand{\teaserheight}[0]{3.05cm}
\newcommand{\heighthuman}[0]{3cm}
\newcommand{\heighthumanstack}[0]{1.5cm}
\newcommand{\heighthumanquery}[0]{2cm}
\newcommand{\heightwolf}[0]{1.6cm}
\newcommand{\heightwolfsketch}[0]{1.5cm}
\maketitle%
\begin{strip}%
  \centerline{%
  \footnotesize%
  \begin{tabular}{cccccccc}%
  \setlength{\tabcolsep}{0pt}
        \hspace{-0.3cm}
        \begin{tabular}{c}
            \includegraphics[height=\heightwolfsketch]{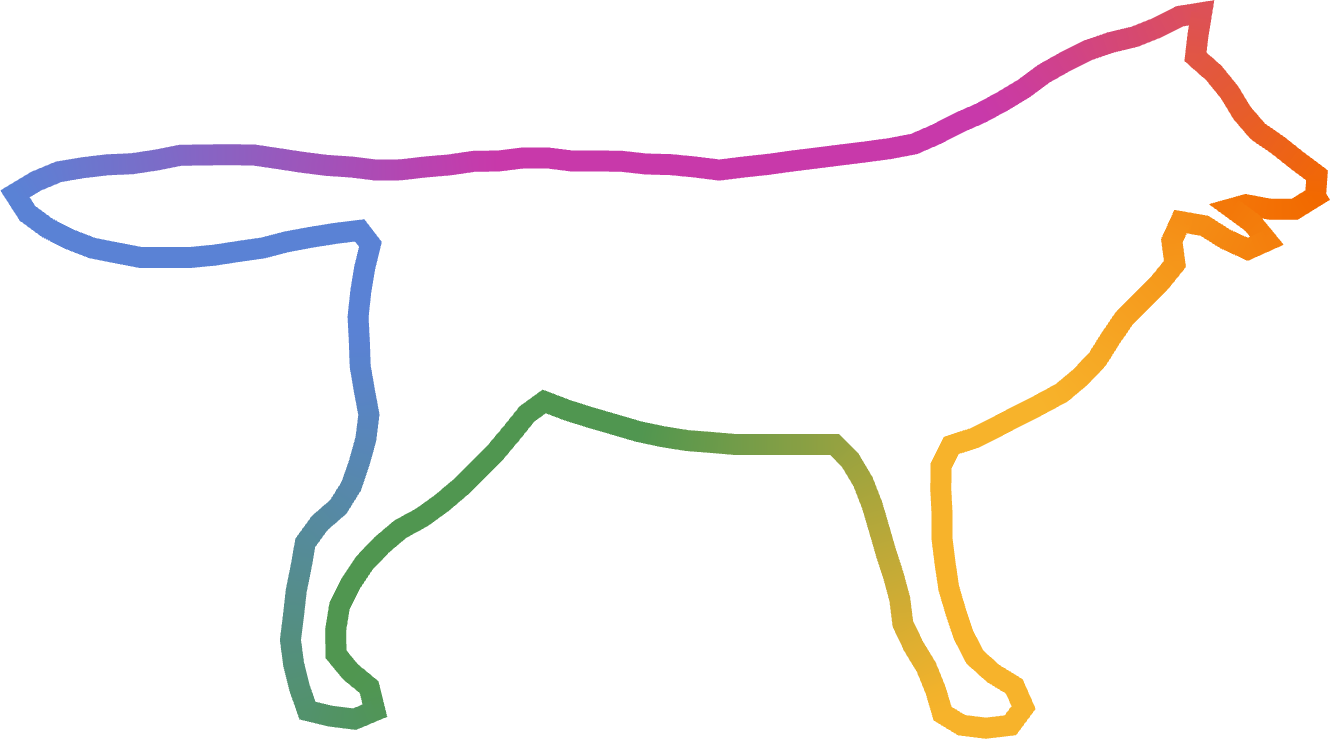}
        \end{tabular} &
        \hspace{-0.4cm}
        \begin{tabular}{cccc}
            \rotatebox{90}{\laehner~\cite{lahner2016}}&
            \includegraphics[height=\heightwolf]{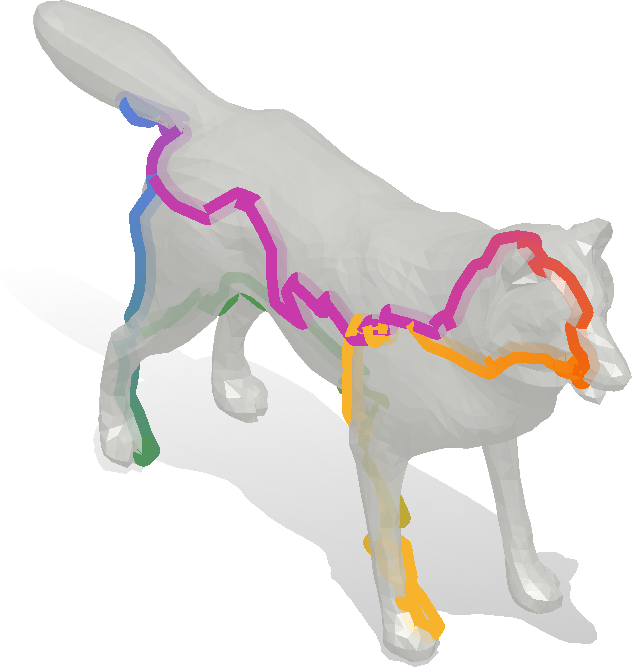} &
            \includegraphics[height=\heightwolf]{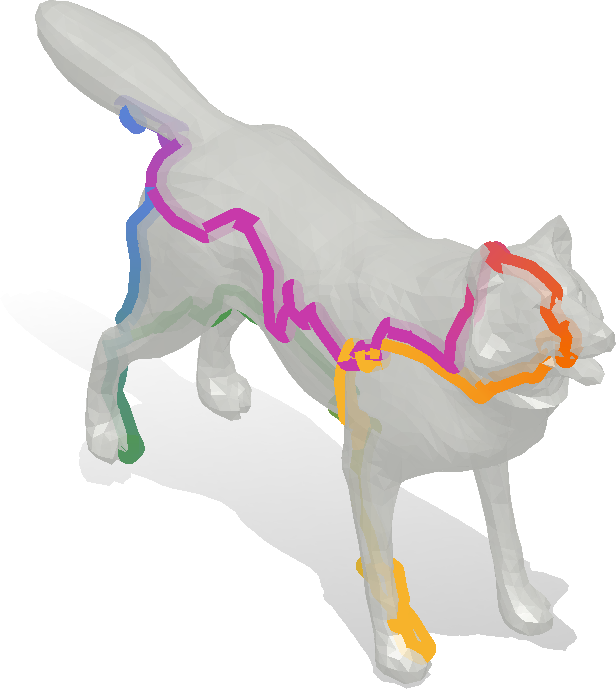} &
            \includegraphics[height=\heightwolf]{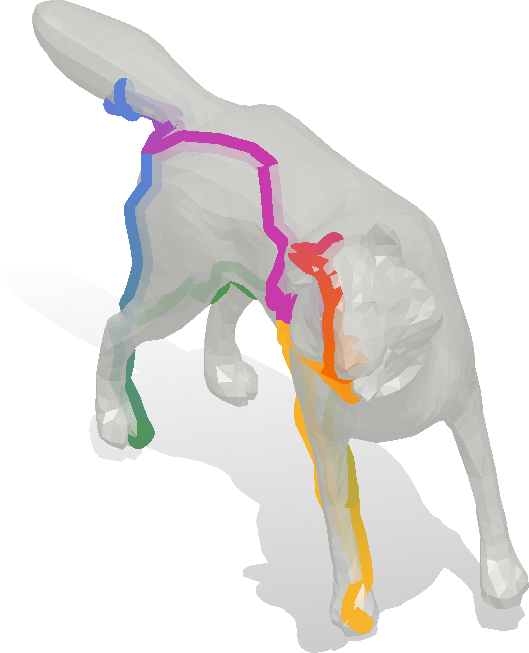}\\
            \rotatebox{90}{$\qquad$Ours} &
            \includegraphics[height=\heightwolf]{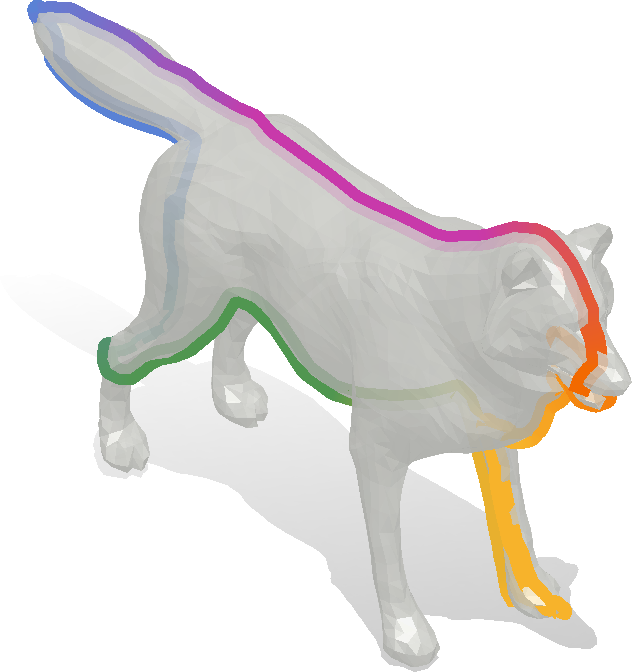} &
            \includegraphics[height=\heightwolf]{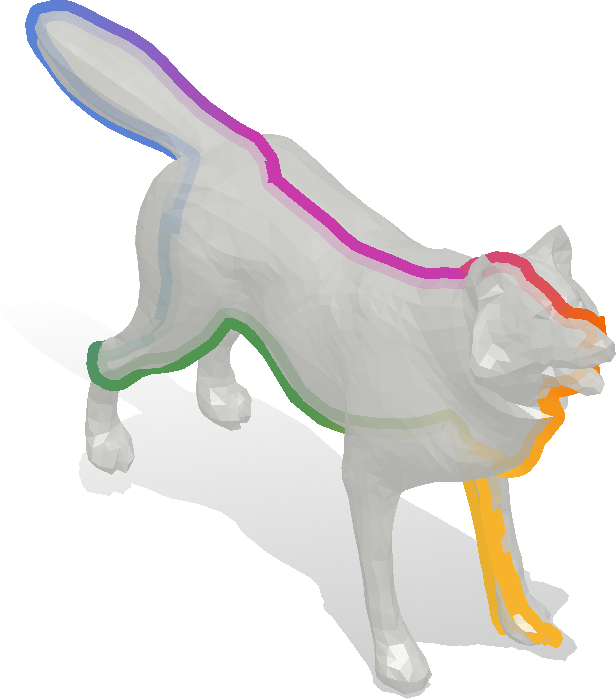} &
            \includegraphics[height=\heightwolf]{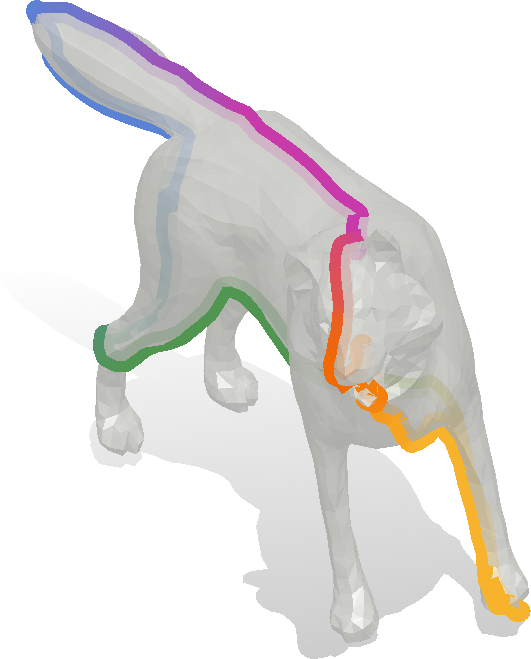}
        \end{tabular} &
        \hspace{-0.6cm}
        \begin{tabular}{c}\rotatebox{90}{\textcolor{gray}{\rule{3cm}{0.5pt}}} \end{tabular} &
        \hspace{-0.6cm}
        \begin{tabular}{cc}
            \includegraphics[height=\heighthuman]{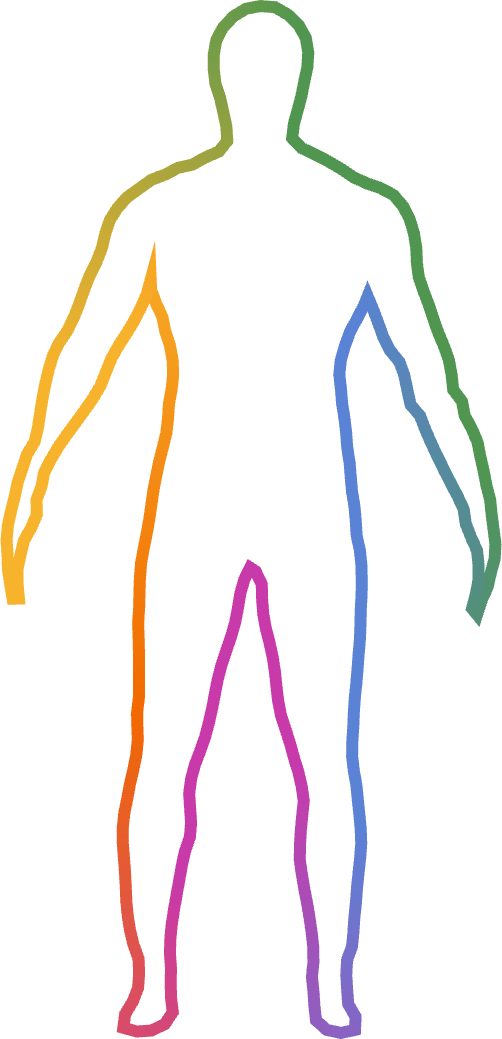}&
            \hspace{-0.3cm}
            \includegraphics[height=\heighthuman]{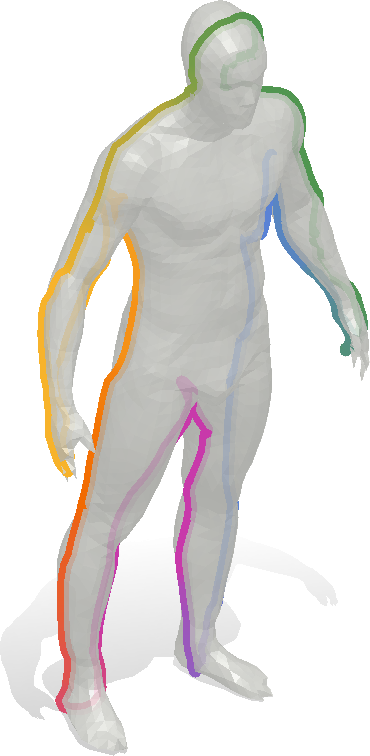}
        \end{tabular} &
        \hspace{-0.6cm}
        $\rightarrow$&
        \hspace{-0.6cm}
        \begin{tabular}{c}
            \includegraphics[height=\heighthumanstack]{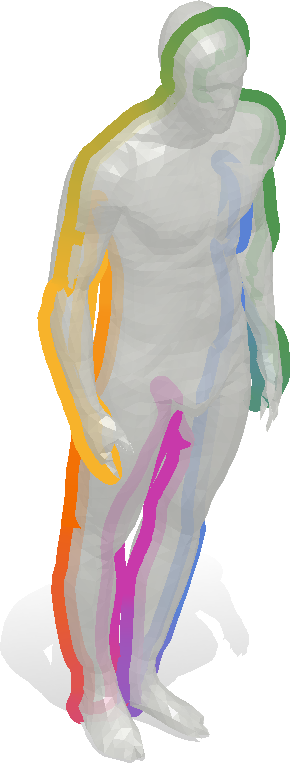}\\
            \includegraphics[height=\heighthumanstack]{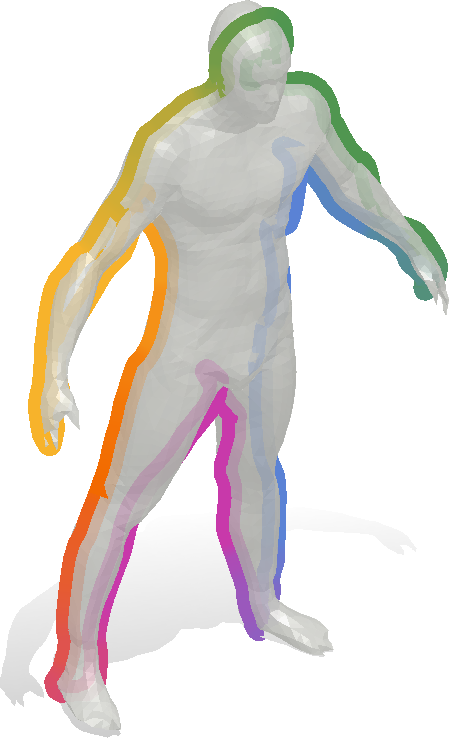}\\
        \end{tabular}&
        \hspace{-0.8cm}
        \begin{tabular}{c}
            \includegraphics[height=\heighthumanquery]{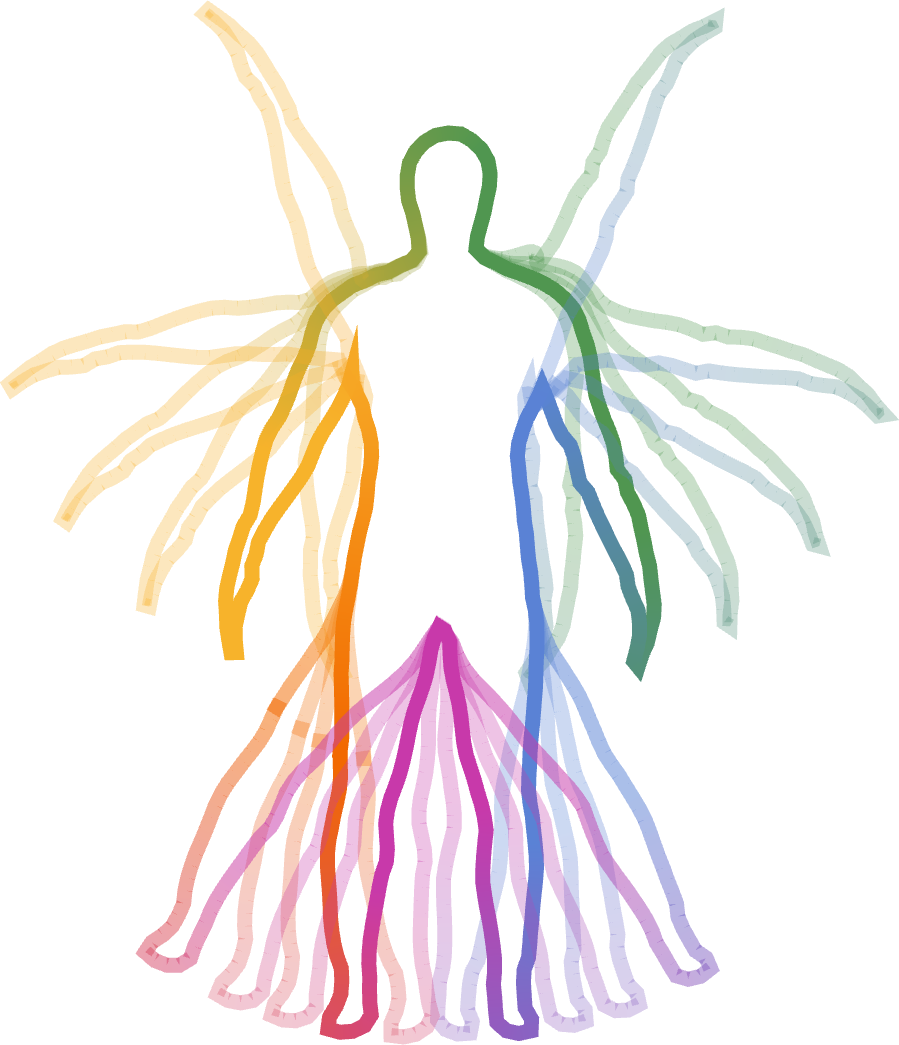}
        \end{tabular}&
        \hspace{-0.8cm}
        \begin{tabular}{c}
            \includegraphics[height=\heighthumanstack]{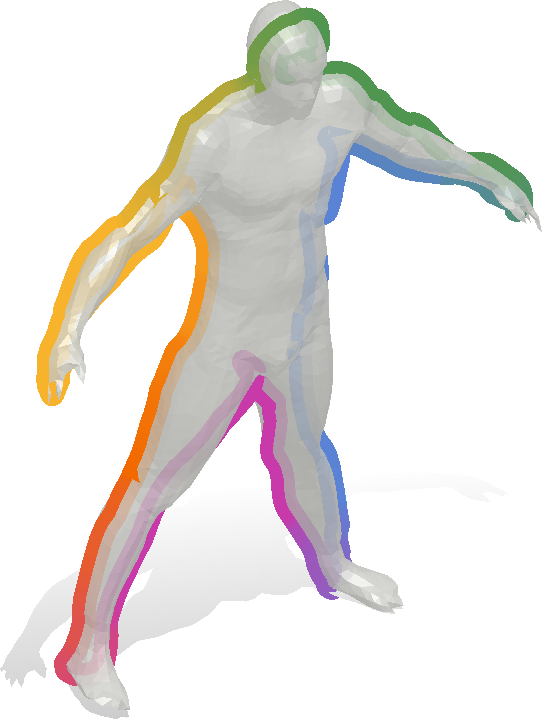} \\
            \includegraphics[ height=\heighthumanstack]{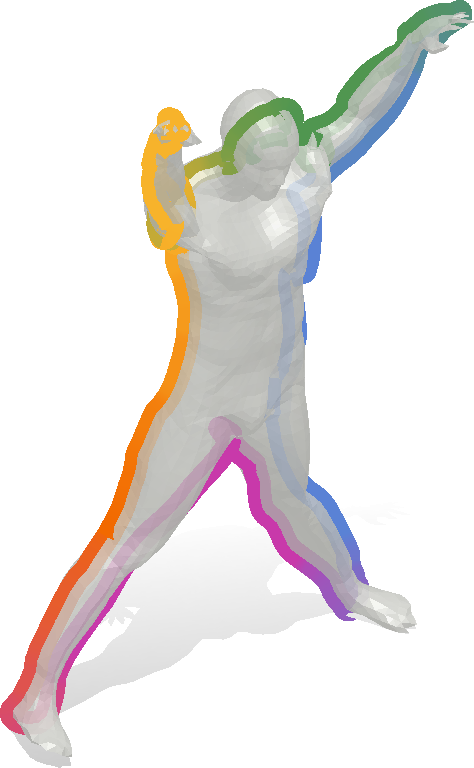}\\
        \end{tabular}\\
        \multicolumn{2}{c}{Results of \laehneretal~\cite{lahner2016} (top) and ours (bottom) on the TOSCA dataset.} & &
        \hspace{-0.6cm}
        (i) Matching with our approach & & 
        \multicolumn{3}{c}{\hspace{-0.4cm}(ii) 2D to 3D deformation transfer}
    \end{tabular}
  }
  \vspace{-2mm}
\captionof{figure}{We propose a novel formalism for \textbf{globally optimal 2D contour to 3D shape matching} based on shortest paths in the \emph{conjugate product graph}. 
For the first time we make it possible to incorporate higher-order costs within a shortest path-based matching formalism, which in turn enables to integrate powerful priors, \eg favouring locally rigid deformations. \textbf{Left:} Our method produces compelling 2D-3D matchings that significantly outperform the previous state of the art~\cite{lahner2016}. %
\textbf{Right:}
Sketch-based 2D to 3D deformation transfer by (i) computing a 2D-3D matching using our approach, (ii) manipulating the 2D sketch, and then  transferring 2D deformations to the 3D shape.}
\label{fig:teaser}
\end{strip}

\begin{abstract}
We consider the problem of finding a continuous and non-rigid matching between a 2D contour and a 3D mesh. While such problems can be solved to global optimality by finding a shortest path in the product graph between both shapes, existing solutions heavily rely on unrealistic prior assumptions to avoid degenerate solutions (e.g.~knowledge to which region of the 3D shape each point of the 2D contour is matched). To address this, we propose a novel 2D-3D shape matching formalism based on the conjugate product graph between the 2D contour and the 3D shape. Doing so allows us for the first time to consider higher-order costs, i.e.~defined for edge chains, as opposed to costs defined for single edges. This offers substantially more flexibility, which we utilise to incorporate a local rigidity prior. By doing so, we effectively circumvent degenerate solutions and thereby obtain smoother and more realistic matchings, even when using only a one-dimensional feature descriptor. Overall, our method finds globally optimal and continuous 2D-3D matchings, has the same asymptotic complexity as previous solutions, produces state-of-the-art results for shape matching and is even capable of matching partial shapes. Our code is publicly available.\footnote{\url{https://github.com/paul0noah/sm-2D3D}}
\end{abstract}
\vspace{-3mm}
\section{Introduction} \label{sec:introduction}
In recent years the computer vision community has put great effort into the matching of either two 2D or two 3D shapes. %
However, the task of matching a 2D shape to a 3D shape is a problem that has received less attention, even though it has a high practical relevance due to its wide variety of applications.
For example, 2D-3D shape matching has the potential to bridge the gap between the 2D and 3D domain by making  interaction with 3D objects more accessible to non-experts, who typically find the manipulation of 2D shapes more intuitive than operating 3D modelling tools.
In addition to  the  modelling and manipulation of 3D shapes using 2D sketches (see \cref{fig:teaser}), 2D-3D shape matching is relevant for 3D shape retrieval from 2D queries (\eg~searching a hand-drawn sketch in a 3D shape database), for augmented reality interactions (\eg~using hand-gestures to select 3D objects), for 3D image analysis based on 2D images (\eg~matching 2D X-ray image segmentations to 3D CT image segmentations), or for multimodal 2D-3D shape analysis. %

2D-3D shape matching can be framed as finding a continuous mapping of a 2D contour (\eg~a sketch of an animal outline) to a 3D shape (\eg~a 3D model of this animal), see \cref{fig:teaser} (left). Here, the matched 2D contour that is deformed to the 3D shape should resemble the original 2D shape as much as possible, \ie spatial shape deformations should be small.
While humans have an intuitive and implicit understanding of \textit{good} 2D-3D matchings, %
unfortunately, it is non-trivial to transfer this understanding into a rigorous mathematical framework: left-right flips are not distinguishable; the 2D shape does not contain all parts of the 3D shape (\eg~2D shape of the wolf in \cref{fig:teaser} (left) contains only two legs); usually there is more than one \textit{good} solution; and even slight deviations from a \textit{good} matching can either be another good matching or can be a bad matching (\eg~zig-zagging on the 3D shape).
In addition, phrasing 2D-3D shape matching as an optimisation problem requires to compute features on both shapes that allow to distinguish corresponding points from non-corresponding points -- this is particularly difficult as many of the widely-used features for 2D or 3D shapes
do not have a natural counterpart in the other domain, and are thus not directly comparable.

Nevertheless, previous work shows that matching a 2D contour to a 3D shape can be efficiently and  optimally solved based on  shortest paths in product graphs~\cite{lahner2016}. However, existing solutions require strong, unnatural assumptions (\eg~a coarse pre-matching, see \cref{sub:bg:matching}) in order to resolve (some of) the above-mentioned difficulties.
In this paper we present a novel graph-based formalism that relaxes previous unnatural assumptions, which in turn allows to solve a substantially more difficult setting of 2D-3D shape matching.
Our main contributions are:
\begin{itemize}
    \item We present a novel {matching} formalism based on conjugate product graphs that allow to encode more expressive higher-order information.
    \item For the first time this makes it possible to 
    impose a local rigidity prior that penalises deformations, which in turn leads to previously unseen matching quality. 
    \item Opposed to involved high-dimensional feature descriptors that were previously used (\eg~spectral features), our method requires only a simple one-dimensional feature that encodes the local object thickness -- a feature that can naturally be defined for 2D and 3D shapes. %
    \item Overall, our technical contributions allow us to solve 2D-3D matching for the first time without the requirement of a coarse pre-matching. %
\end{itemize}
\section{Related Work} \label{sec:relatedwork}
In the following we summarise existing works that we consider most relevant in the context of this paper.

\subsubsection{Geometric Feature Descriptors}
Most matching approaches rely on point-wise features to decide what are good potential matches. 
For 2D contours, popular features are cumulative angles \cite{veltkamp2001state}, curvature \cite{veltkamp2001shape, kun2019shape} and various distance metrics  \cite{veltkamp2001shape,veltkamp2001state, schmidt2007fast,ling2007inner,michel2011scale}.
One (for our work) notable example from the class of distance-based metrics is to consider
the distance from each point to other parts of the contour along several fixed rays \cite{michel2011scale}.

On 3D shapes, other feature types are predominant because the geometry is more complicated and 2D features often do not have a direct equivalent in 3D. 
While curvature does exist in 2D and in 3D, in 3D there are multiple notions of curvature, like mean, Gaussian and directional curvature. 
More popular are higher-dimensional features like the heat kernel signature \cite{sun09hks} or wave kernel signature \cite{aubry2011}, which are based on spectral properties of the 3D surface, or the SHOT descriptor \cite{tombari10shot} based on the distribution of normals in the neighbourhood of a vertex. 
Recent approaches aim to learn suitable features for a specific matching pipeline \cite{litany17fmnet,marin2020embedding}.
Overall, there is a discrepancy between 2D and 3D features, and even for features that can conceptually be calculated for both domains, 
they are typically not directly comparable. 
While \cite{lahner2016} successfully addresses 2D-3D \emph{shape retrieval} based on spectral 2D and 3D features, our experiments confirm that these features are insufficient to achieve precise \emph{correspondences}.  
Similarly, many approaches learn multi-modal or multi-dimensional descriptors for entire shapes \cite{herzog15lesss,wang2015sketchbased,qin2022shrecsketch}, but these are only useful in retrieval settings and not capable of point-to-point comparisons needed for finding reliable correspondences. 
In this paper we instead shift  focus to incorporating a powerful deformation prior, so that in turn substantially simpler feature descriptors are sufficient. 
We demonstrate that this allows to consider simple distance-based features which can be consistently computed both in 2D and 3D.

\subsubsection{2D-3D Matching} \label{sub:rw:2D3Dmatching}
Matching pairs of 2D objects
is well-researched and it is widely known that respective solutions can be represented as paths in a graph. With that, shortest path algorithms can be used to efficiently find globally optimal solutions. This has for example been done for open contours, known as dynamic time warping~\cite{sakoe1978dynamic}, and closed contours \cite{schmidt2007fast}, including invariance to scale and partiality \cite{michel2011scale}. Similarly, it was shown that matching 2D contours to 2D images (e.g.~for template-based image segmentation) can be addressed using a similar framework~\cite{schoenemann2008elasticimages}.
Matching two 3D shapes is considerably harder as the solution is not a shortest path (in the product manifold) anymore but rather a minimal surface embedded in four-dimensional space.
Thus, imposing constraints on the continuity of the solution is not possible in an efficient way~\cite{windheuser2011,windheuser2011a,roetzer2022scalable}.
We note that there are works which consider (geodesic) shortest paths in the space of 3D shapes, but they are not solvable to global optimality \cite{glaunes2008large, jermyn2012elastic}.
While from an algorithmic perspective finding a 2D-3D matching is easier than the 3D-3D case  (as the former also amounts to a (cyclic) shortest path problem~\cite{lahner2016}), quantifying matching costs is significantly more difficult for the 2D-3D case (cf.~previous paragraph on feature descriptors).
In this work we build upon the path-based 2D-3D matching formalism of~\cite{lahner2016} and propose a novel formalism that enables the use of higher-order costs (defined for chains of edges, opposed to costs of single edges). In turn, our formalism allows for the first time to incorporate a spatial deformation prior, so that
our framework requires substantially less descriptive features -- in fact, we demonstrate that a one-dimensional distance-based feature descriptor 
is sufficient to successfully solve 2D-3D shape matching.

\subsubsection{Extensions of Graph Representations} \label{sub:rw:productgraphs}
Graphs are not only relevant for diverse subfields of visual computing, such as e.g.~image analysis~\cite{geman1984stochastic,mortensen1995intelligent,shi2000normalized,rother2004grabcut,roth2005fields,boykov2006graph,kappes2011globally,keuper2015efficient,bernard2018ds},
recognition~\cite{conte2004thirty,felzenszwalb2005pictorial}, tracking~\cite{zhang2008global,iqbal2017posetrack}, or mesh processing~\cite{mitchell1987discrete,kalogerakis2010learning,lahner2016,bernard2017combinatorial}, but also for a wide variety of other application domains, for example in DNA research \cite{eghdami2012application}, language processing \cite{wilson2005recognizing}, or social sciences \cite{jin2013understanding}.
In graph theory, many graph extensions have been proposed, including
multilayer networks~\cite{kivela2014multilayer},
dual graphs \cite{harary1969graphtheory}, hypergraphs~\cite{berge1984hypergraphs}, 
and product graphs~\cite{hammack2011handbook}, to name just a few. 
The product graph extends the concept of the Cartesian product (and other types of products) to graphs by additionally encoding neighbourhood information.
This has been used in the context of different matching problems, including 2D-2D \cite{schmidt2007fast}, 3D-3D \cite{windheuser2011}, and 2D-3D \cite{lahner2016} settings.
Another graph extension relevant for this paper is the \emph{conjugate graph} (also known as \emph{line graph}), which encodes connectivity information into the vertices instead of edges~\cite{harary1969graphtheory}. 
This has for example been used for route planning \cite{winter2002modeling} or graph link prediction \cite{cai2021line}.
In this work, we propose to combine product graphs with conjugate graphs (in fact we consider the conjugate graph of a product graph between two shapes)  and showcase that this substantially increases modelling expressiveness and flexibility, and therefore allows for globally optimal 
2D-3D shape matching.

\section{Background \& Notation}
In this section we introduce our notation (also see \cref{table:notation}), conjugate graphs, and the formalism for the matching of shapes as shortest path problem on a product graph.
\begin{definition}[Directed Graph]
A directed graph $\graph$ is defined as a tuple $(\graphVerts, \graphEdgs)$ of vertices $\graphVerts$ and oriented edges $\graphEdgs \subset \graphVerts \times \graphVerts$ (\ie~oriented edge $(v_1, v_2) \in \graphEdgs$ does not imply $(v_2, v_1) \in \graphEdgs$).
\end{definition}
We directly work with discrete graph-based representations for shapes, \ie 2D shapes are represented as contours sampled at $m$ many points, and 3D shapes are represented as (manifold) triangular surface meshes:
\begin{definition}[2D Shape]
We define a 2D shape (or contour) $\contour$ as a tuple $(\contourVerts, \contourEdgs)$ of $m$ vertices $\contourVerts$ and $m$ oriented
edges $\contourEdgs \subset \contourVerts \times \contourVerts$ s.t. $\contour$ is a directed cycle graph.%
\end{definition}
\begin{definition}[3D Shape]
We define a 3D shape $\mesh$ as a tuple $(\meshVerts, \meshEdgs)$ of vertices $\meshVerts$ and oriented
edges $\meshEdgs \subset \meshVerts \times \meshVerts$ such that $\mesh$ forms a 2D manifold in 3D space (triangular surface mesh, possibly with multiple boundaries). 
\end{definition}
We also consider an extended edge set, which contains all vertices represented as self-edges:
\begin{definition}[Extended Set of Edges]
For a (2D or 3D) shape $\arbitraryShape = (\verts_\arbitraryShape, \edges_\arbitraryShape)$, we define the extended set of edges  $\extEdgs_\arbitraryShape = \edges_\arbitraryShape \cup \{(a,a) |\;a \in \verts_\arbitraryShape\} \subset \verts_\arbitraryShape \times \verts_\arbitraryShape$. We call the additional edges \emph{degenerate edges}.
\end{definition}

\begin{table}[h!]
\small\centering
	\begin{tabularx}{\columnwidth}{lp{5.6cm}}
 \toprule
        \textbf{Symbol} & \textbf{Description} \\
		\toprule
		$\contour = (\contourVerts, \contourEdgs)$ &2D shape (closed contour)\\
		$\mesh = (\meshVerts, \meshEdgs)$ &3D shape (manifold triangular surface mesh)\\
		$e^\contour,\; e^\mesh $ &edge $e^\contour$ of contour, edge $e^\mesh$ of mesh\\
		$\extEdgs_\arbitraryShape$ &extended set of edges of shape $\arbitraryShape {\in} \{\contour, \mesh \}$ \\
		$\prodGraph = (\prodVerts, \prodEdgs)$ &product graph of $\contour\times \mesh$ with product vertices $\prodVerts$ and product edges $\prodEdgs$\\
		$e,\; v$ &edge $e$, vertex $v$ of $\prodGraph$\\
		$\conjProdGraph = (\conjProdVerts, \conjProdEdgs)$ &conjugate product graph of $\prodGraph$ with vertices $\conjProdVerts$ and edges $\conjProdEdgs$\\
		$e^*,\; v^*$ &edge $e^*$, vertex $v^*$ of $\conjProdGraph$\\
		\bottomrule
	\end{tabularx}
	\caption{Summary of the notation used in this paper.
	}
	\label{table:notation}
\end{table}
\subsection{Conjugate Graphs}
\begin{figure}[h]
    \centering
    \includegraphics[trim={1cm 0 0 0},clip,width=.9\columnwidth]{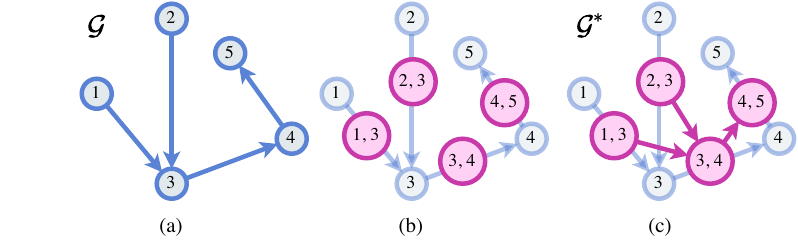}
    \vspace{-0.3cm}
    \caption{Illustration  of the \textbf{conjugate graph}. (a) Input graph $\mathcal{G}$. (b) Each edge of $\mathcal{G}$ becomes a conjugate vertex (\textcolor{cPINK}{\large$\bullet$}). (c) The conjugate graph $\mathcal{G}^*$ is now formed by connecting the newly introduced vertices (\textcolor{cPINK}{\large$\bullet$}) by edges ({\color{cPINK} \rule[0.6mm]{0.3cm}{0.9mm}}) according to \cref{def:conjgraph} (\eg conjugate vertex $(1,3)$ and $(3,4)$ are connected by a directed edge since they are both adjacent to vertex $3$ in $\mathcal{G}$).
    }
    \label{fig:line-graph}
\end{figure}

\begin{definition}[Conjugate Graph~\cite{harary1969graphtheory}]\label{def:conjgraph}
The conjugate graph $\conjGraph$ of a directed graph $\graph$ is defined as a tuple $(\conjGraphVerts, \conjGraphEdgs)$ with
$$
    \conjGraphVerts = \graphEdgs,~~
    \conjGraphEdgs = \{\big((v_1, v_2), (v_2, v_3)\big) \in \conjGraphVerts {\times} \conjGraphVerts \}. \label{eq:conjEdge}
$$
\end{definition}
Intuitively, the edges of $\graph$ become the vertices of $\conjGraph$ and the conjugate edges connect pairs of adjacent edges from $\graph$.
\cref{fig:line-graph} illustrates the construction of the conjugate graph.

\subsection{Matching Formalism} \label{sub:bg:matching}
Next, we introduce the product graph $\mathcal{G}$ between a 2D contour and a 3D shape, and we summarise L\"{a}hner \etal's~\cite{lahner2016} representation of a 2D-3D shape matching as shortest (cyclic) path in the product graph.
\begin{definition}[Product Graph]\label{def:prodgraph}
The product graph $\prodGraph$ of the 2D contour $\contour$ and the 3D shape $\mesh$
is a tuple $(\prodVerts_\prodGraph, \prodEdgs_\prodGraph)$ of product vertices $\prodVerts_\prodGraph$ and product edges $\prodEdgs_\prodGraph$, where%
\begin{align}
    \prodVerts_\prodGraph &= \contourVerts\ \times \meshVerts, \nonumber\\
    \prodEdgs_\prodGraph &= \{\left( e_1^\contour, e_2^\mesh\right) \in \prodVerts_\prodGraph\times \prodVerts_\prodGraph\; |\; e_1^\contour \in \extEdgs_\contour,e_2^\mesh \in \extEdgs_\mesh, \nonumber\\
                &\qquad \; e_1^\contour \text{ or } e_2^\mesh \text{ non-deg.}\}. \nonumber
\end{align}
\end{definition}
The product graph $\prodGraph$ is visualised in \cref{fig:product-graph} (left).
To simplify the notation, we will refer to the vertices and edges of the product graph  as $\prodVerts$ and $\prodEdgs$ for the remainder of the paper. 
A matching between the 2D contour and the 3D shape can be represented as the subset $\mathcal{C} \subset \contourVerts \times \meshVerts$ %
where tuples in $\mathcal{C}$ indicate which vertices
of the 2D contour and 3D shape are in correspondence. 
Desirable properties of such matchings are that a) each vertex on $\contour$ is matched to at least one vertex on $\mesh$, and b) the matching is continuous, \ie if two vertices on $\contour$ are adjacent, their matches on $\mesh$ should also be adjacent. 
These properties can be guaranteed if the solution is a (cyclic) path that goes through all layers of the product graph (cf.~\cref{fig:product-graph}).
A path that minimises costs defined on the (product graph) edges can efficiently be computed based on Dijkstra's algorithm \cite{dijkstra59}.
To ensure the path is cyclic, Dijkstra's algorithm needs to be run multiple times (once for each vertex of the 3D mesh); however, the number of Dijkstra runs can be drastically reduced based on a simple branch and bound strategy (see~\cite{lahner2016} and Appendix).

Despite the theoretical elegance of \laehneretal's formalism, 
a major limitation is that shortest paths only take into account costs of individual edges. With that, the approach is not capable of penalising local deformations induced by a matching (which is only possible with \emph{pairs} of product graph edges).
Instead, the authors use high-dimensional features in combination with knowledge about pre-matched segmentations between 2D and 3D shapes. While such a pre-matching drastically reduces the search space and avoids many degenerate solutions, the knowledge of such a pre-matching is typically not available in practice.

\newcommand{\threeedgepathProd}[0]{%
{\color{cBLUE} \rule[0.6mm]{0.3cm}{0.9mm}}%
{\large\color{cBLUE} $\bullet$}%
{\color{cGRAY} \rule[0.6mm]{0.3cm}{0.9mm}}%
{\large\color{cBLUE1} $\bullet$}%
{\color{cBLUE1}\rule[0.6mm]{0.3cm}{0.9mm}} 
}
\newcommand{\layerOne}{{\large\color{cBLUE} $\bullet$}}
\newcommand{\layerTwo}{{\large\color{cBLUE1} $\bullet$}}
\newcommand{\layerThree}{{\large\color{cBLUE2} $\bullet$}}
\newcommand{\threeedgepathConjProd}[0]{%
{\large\color{cORANGE} $\bullet$}%
{\color{cPINK} \rule[0.6mm]{0.3cm}{0.9mm}}%
{\large\color{cORANGE} $\bullet$}%
{\color{cPINK} \rule[0.6mm]{0.3cm}{0.9mm}}%
{\large\color{cORANGE} $\bullet$}%
}
\begin{figure}[h!]
    \centering
     \includegraphics[trim={1cm 0 0 0.2cm},clip,width=.9\columnwidth]{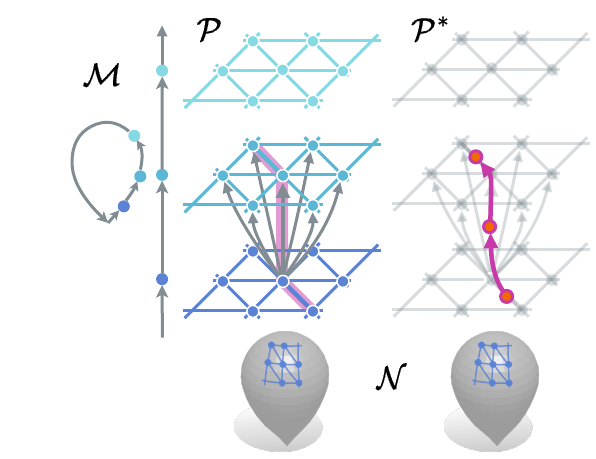}
    \vspace{-0.2cm}
    \caption{Illustration of the product graph $\prodGraph$ (left) and the conjugate product graph $\conjProdGraph$ (right) for a water drop shape.
    \textbf{Left:} the product graph $\prodGraph$ is structured into layers (\layerOne,\layerTwo,\layerThree), where each layer represents a single vertex on $\contour$ and the entire 3D shape $\mesh$.
    \textbf{Right:} we illustrate (part of) the conjugate product graph  $\conjProdGraph$ for the three-edge path \threeedgepathProd{}   in $\prodGraph$ (highlighted in pink), which becomes \threeedgepathConjProd{} in $\conjProdGraph$. Conjugate product vertices {\large\color{cORANGE} $\bullet$} are shown in {\textcolor{cORANGE}{orange}} and conjugate product edges {\color{cPINK} \rule[0.6mm]{0.3cm}{0.9mm}} are shown in {\textcolor{cPINK}{pink}}.
    }
    \label{fig:product-graph}
\end{figure}

\section{Our 2D-3D Shape Matching Approach}\label{se c:method}
In the following, we present our solution for 2D-3D shape matching which allows to incorporate higher-order edge costs which in turn is enabled through conjugate product graphs. 
With this, spatial deformations can be naturally penalised within a shortest path matching framework while still finding a globally optimal matching in polynomial time (with the same asymptotic complexity as~\cite{lahner2016}), see Section~\ref{sub:bg:matching}. 
We emphasise that we address a significantly more difficult
problem setting than~\cite{lahner2016} since we do not rely on the unrealistic assumption that a coarse 2D-3D pre-matching is available.

\begin{figure}[h]
    \small
    \centering
    \begin{tikzpicture}[level distance=0.7cm, sibling distance=0.8cm,
    edge from parent/.style={draw,%
    thick}]
    \Tree 
    [.\node (level0-right){$e^*$};
        [.\node (level1-right){$e_1 ~(= v_1^*)$};
            [.\node (level2-right){$e^\contour_1$}; ]
            [.{$e^\mesh_1$} ]
        ] 
        [.{$e_2 ~(= v_2^*)$}
            [.{$e^\contour_2$} ]
            [.{$e^\mesh_2$} ] 
        ] 
    ]
    \foreach \Value/\Text in {0/{$\conjProdGraph = (\conjProdVerts,\conjProdEdgs$)},1/{$\prodGraph = (\prodVerts, \prodEdgs)$},2/{$\contour,\mesh$}}
    {  
      \node[anchor=west] 
        at ([xshift=-3cm]{level2-right}|-{level\Value-right}) 
        {\Text};
    }
    \end{tikzpicture}
    \vspace{-0.1cm}
    \caption{Hierachical relationship between edges of the involved graphs. An edge $e^*\in\conjProdEdgs$ of the conjugate product graph $\conjProdGraph$ is formed by two edges $e_1, e_2 \in\prodEdgs$ of the product graph $\prodGraph$ (which correspond to vertices $v_1^*, v_2^*\in \conjProdVerts$ in the conjugate product graph $\conjProdGraph$, respectively). Each  edge $e_\bullet$ of $\prodGraph$ is formed by one edge $e_\bullet^\contour\in \contourEdgs^+$ and one edge $e_\bullet^\mesh\in \meshEdgs^+$ of the
    shapes $\contour$ and $\mesh$, respectively.}
    \label{fig:notation-tree}
\end{figure}
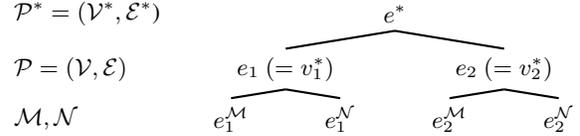

\subsection{Conjugate Product Graphs}
Our formalism builds upon conjugate product graphs, i.e.~the conjugate graph (\cref{def:conjgraph}) of a product graph (\cref{def:prodgraph}).
We refer to the conjugate product graph as $\conjProdGraph = (\conjProdVerts, \conjProdEdgs)$. 
Here, edges in the product graph become vertices in the conjugate product graph and  are connected based on the  adjacency of vertices in the product graph, see \cref{def:conjgraph}. %
Thus, an edge $e^*\in\conjProdEdgs$ in $\conjProdGraph$ has the scope of two edges in the product graph $\prodGraph$, i.e.~$e^* = (e_1,e_2),\; e_1, e_2 \in \prodEdgs$, see \cref{fig:notation-tree}. In turn, this enables the definition of cost functions that consider two product graph edges simultaneously which allows to integrate powerful priors, \ie~regularise the problem so that locally rigid deformations are favoured an thus resulting matchings are smoother and more realistic (cf. \cref{fig:teaser} and \cref{fig:against-sota-qualitative}). %
We note that higher-order costs can also be defined by repeating the conjugation process, \eg~an edge in the conjugate of the conjugate product graph is formed by three edges of the product graph so that costs can be defined for triplets of product graph edges (and so on). For brevity and a simpler exposure, in the following we restrict ourselves w.l.o.g.~to second-order costs.

\subsection{Cost Function}
We define our cost function $d: \conjProdEdgs \rightarrow \mathbb{R}$  for every  edge $e^* = (v_1^*, v_2^*) = (e_1, e_2) \in \conjProdEdgs$ in the conjugate product graph as
\begin{equation}
    d(e^*) = d_\text{data}(e^*) + d_\text{reg}(e^*).
\end{equation}

$d_\text{data}$ is the data term which measures the similarity between the product edges based on feature descriptors.
$d_\text{reg}$ is a local rigidity regulariser which ensures that adjacent edges on the 2D contour are deformed similarly %
to adjacent elements on the 3D shape.
We first describe the data term followed by the local rigidity regulariser.

\subsubsection{Data Term}
A major difficulty when comparing 2D and 3D shapes is that many of the existing geometric feature descriptors cannot be consistently defined for 2D and 3D shapes (\eg~although the notion of curvature exists for both shapes, in 3D the curvature is direction-dependent, which makes it difficult to compare 2D and 3D curvature).
We build a simple one-dimensional descriptor based on the observation that corresponding points
$\contourVert \in \contour$ and $\meshVert \in \mesh$ of the same shape class should have a
similar distance to the other side going through the interior of the respective shape, see \cref{fig:local-thickness}. %
As such, we consider \emph{local thickness} as feature descriptor. It is computed as follows:
\begin{itemize}
    \item For a vertex $\contourVert \in \contour$, we compute the \emph{2D local thickness} $\ell_\contourVert^{\text{2D}}$ by following the inverted vertex-normal (through the interior of the shape) until the (first) intersection with the contour \cite{michel2011scale}.
    \item  For a vertex $\meshVert \in \mesh$, we compute the \emph{3D local thickness} $\ell_\meshVert^{\text{3D}}$ by following the inverted vertex-normal (through the interior of the shape) until the (first) intersection with the mesh. We employ a triangle-ray-intersection algorithm for this \cite{moller2005fast}.
\end{itemize}
    
\noindent
With that, we define the local thickness difference for each conjugate product edge $e^* = (e_1, e_2)\in \conjProdEdgs$, so that our data term $d_\text{data}(e^*)$ reads 
\begin{equation}
    d_\text{data}(e^*) = \psi_1\big(|\ell_{\contourVert}^{\text{2D}} - \ell_{\meshVert}^{\text{3D}}|)\,,
\end{equation}
where $\ell_{\contourVert}^{\text{2D}}$ and $\ell_{\meshVert}^{\text{3D}}$ are local thickness values at vertices $\contourVert \in \contourVerts$ and $\meshVert \in \meshVerts$ on the 2D and 3D shape, respectively. Large deformations of shapes potentially lead to large, local outliers of respective local thickness. To reduce the influence of such outliers, 
we additionally apply the function $\psi_1(\cdot)$ to the absolute value of the local thickness difference. For example, $\psi_1(\cdot)$ can be chosen to be a robust loss function. We avoid taking into account the same local thickness value multiple times by computing the local thickness difference $d_\text{data}(e^*)$ at a conjugate product vertex $e^*$ solely with the local thickness at vertex $i$ shared by $e_1^\contour$ and $e_2^\contour$, and respectively the local thickness at vertex $j$ shared by $e_1^\mesh$ and $e_2^\mesh$.
We have found that despite its simplicity, local thickness is an effective one-dimensional feature descriptor that, in combination with our local rigidity regulariser, enables faithful 2D-3D shape matchings, see \cref{sec:experiments}.

\subsubsection{Regularisation}
Inspired by \cite{bernard2017combinatorial, sorkine2007rigid}, we employ a regularisation term which favours deformations that are locally rigid.
To compute the regularisation of the conjugate product edge $e^* = (e_1,e_2) = \left((e_1^\contour,e_1^\mesh),(e_2^\contour,e_2^\mesh)\right)$ (cf.~\cref{fig:notation-tree}), we define a local 3D coordinate frame for each of the four (shape) edges $e_1^\contour,e_1^\mesh,e_2^\contour,e_2^\mesh$. %
To this end, we embed the 2D contour into 3D space by adding a third constant coordinate. With that, for both 2D contour and 3D shape, we can define a local 3D coordinate frame based on the normalised edge direction, outward-pointing unit normal, and their cross product. Subsequently, we solve the orthogonal Procrustes problem~\cite{ten1977orthogonal} in order to compute the rotation $R_{e_1}$ that aligns the 3D coordinate frame of $e_1^\contour$ to the 3D coordinate frame of $e_1^\mesh$, and the rotation $R_{e_2}$ that aligns the 3D coordinate frame of $e_2^\contour$ to the coordinate of $e_2^\mesh$, see \cref{fig:local-rigidity}. In presence of degenerate edges we simply use the previous edge, see also \cref{sub:method:theory}.%

By computing the geodesic distance between  $R_{e_1}$ and $R_{e_2}$ on the Lie group $\text{SO}(3)$, we can quantify the amount of non-rigidity of the matching that is induced by $e^*$.
\newcommand{\featureFigureHeight}{2.2cm}
\begin{figure}[t]
    \centering
    \begin{subfigure}{0.3\columnwidth}
        \centering
        \includegraphics[trim={0 0.5cm 1.3cm 0},clip, height=\featureFigureHeight]{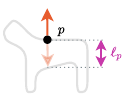}
        \vspace{-0.9cm}
        \caption{}%
        \label{fig:local-thickness}
    \end{subfigure}
    \hfill
    \begin{subfigure}{0.6\columnwidth}
        \centering
        \includegraphics[trim={0.5cm 0 0.7cm 0}, clip, height=\featureFigureHeight]{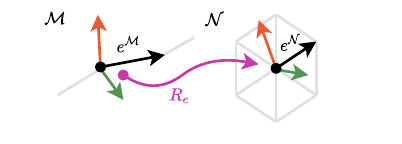}
        \vspace{-0.9cm}
        \caption{}%
        \label{fig:local-rigidity}
    \end{subfigure}
    \vspace{-0.1cm}
    \caption{(a) The \textbf{local thickness $\ell_p$} for the point $p$  is found by intersecting the ray from $p$ in the opposite normal direction (light red) with the shape. %
    (b) Illustration of finding the \textbf{rotation $R_{e}$} that aligns the 3D coordinate frame defined for a 2D contour edge $e^\contour$ and the 3D coordinate frame defined for a 3D shape edge $e^\mesh$. The black vector shows the shape edge, the red vector the normal and the green vector their cross product.
    }
\end{figure}
For computational efficiency, we consider
unit quaternion representations $q_{e_\bullet}$ of $R_{e_\bullet}$, so that
our local rigidity regularisation term $d_{\text{reg}}$ for the conjugate product edge $e^*$ reads
\begin{equation}
    d_{\text{reg}}(e^*) = \psi_2\big(\arccos \left( \langle q_{e_1}, q_{e_2} \rangle \right)\big)\,,
\end{equation}
where $\langle\cdot,\cdot\rangle$ is the inner product for quaternions \cite{huynh2009metrics}.
$\psi_2(\cdot)$ is again a robust loss function, see \cref{sub:method:theory}.

\subsection{Theoretical Analysis and Implementation} \label{sub:method:theory}
In the following we provide a theoretical analysis and additional implementation details.

\subsubsection{Cyclic Shortest Paths} \label{sub:method:extension}
To find a \emph{cyclic} shortest path, we can run an ordinary (non-cyclic) shortest path algorithm (\eg~Dijkstra's algorithm~\cite{dijkstra59}) $|\meshVerts|$ many times. 
To this end, we duplicate the last layers in the conjugate product graph (see \cref{fig:product-graph}), disconnect the duplicate layers from each other, and for each vertex from the `upper duplicate' find the shortest path to the corresponding vertex in the `lower duplicate'. 
The globally optimal shortest \emph{cyclic} path is now formed by the minimum among the $|\meshVerts|$ individual paths. 
To reduce the number of shortest paths that need to be computed, we can instead resort to a more efficient branch-and-bound strategy, we refer to the Appendix for details.%

\paragraph{Degenerate Cases.} %
Conjugate product vertices containing degenerate edges of the 3D shape do not contain directional information on the 3D shape which we need to compute $d_{\text{reg}}$.
We inject the relevant directional information into the conjugate product graph by introducing new conjugate product vertices that reflect (non-degenerate) edges on the 3D shape adjacent to respective degenerate 3D edge.

\subsubsection{Pruning}\label{sub:method:pruning} %
To decrease the size of 
the conjugate product graph $\conjProdGraph$, we apply a pruning strategy.
To this end, we prune conjugate product edges that reflect local turning points on the 3D shape since such paths represent undesirable matchings. 
In addition, we prune edges that
first represent a degenerate edge of $\contour$, followed by a degenerate edge of $\mesh$ (or vice-versa).
Such combinations are equivalent to a matching with two non-degenerate edges. Overall, our pruning reduces the graph size (and thus runtime) and excludes obvious non-desirable solutions.

\subsubsection{Runtime Analysis}\label{sub:method:runtime}
The runtime of our algorithm depends on the size of the conjugate product graph and the number of shortest path runs. 
The number of vertices in $\conjProdGraph$ corresponds to the number of edges in $\prodGraph$. 
The number of edges in $\conjProdGraph$ can be approximated by $c \cdot |\contourVerts| \cdot \big(|\meshEdgs| + |\meshVerts|\big)$ where $c$ is a constant that is related to the  maximum number of neighbours of the vertices in $\mesh$.
\cref{tab:graphsizes} sums up the sizes of the product graph $\prodGraph$ and the conjugate product graph $\conjProdGraph$.

\begin{table}[H]
    \centering
    \small
    \begin{tabular}{lcc}
      \toprule
      &\textbf{\# vertices} & \textbf{\# edges}\\
      \midrule
      $\prodGraph$ & $|\contourVerts|\cdot|\meshVerts|$& $|\contourVerts| \cdot\big(2|\meshEdgs| + |\meshVerts|$\big)\\
      $\conjProdGraph$ & $|\contourVerts| \cdot \big(2|\meshEdgs| + |\meshVerts|\big)$ & 
      $c \cdot |\contourVerts| \cdot \big(2|\meshEdgs| + |\meshVerts|\big)$ 
      \\
      \bottomrule
    \end{tabular}
    \vspace{-0.2cm}
    \caption{Comparison of sizes of the product graph $\prodGraph$ and the conjugate product graph $\conjProdGraph$.
    } 
    \label{tab:graphsizes}
    \vspace{-0.1cm}
\end{table}

Using $|\meshEdgs| \approx 3|\meshVerts|$~\cite{botsch2010} %
shows that the %
conjugate product graph $\conjProdGraph$ has $7$ times more vertices and $c \approx 11$ (see Appendix) %
times more edges than the product graph $\prodGraph$, which shows that asymptotically both graphs have the same size.
In the worst case $\mathcal{O}(\vert\meshVerts\vert)$ shortest path problems -- one for each vertex in $\mesh$ -- have to be solved.
Together with the complexity of each Dijkstra run 
the final time complexity can be estimated as  $\mathcal{O}\left( \vert \contourVerts \vert \cdot \vert \meshVerts \vert^2 \cdot \log( \vert \meshVerts \vert ) \right)$, which is the same as in \cite{lahner2016}.
We provide more details as well as runtime plots in the Appendix.

\subsubsection{Implementation Details}
We implement the shortest path algorithm in C++ wrapped in a MATLAB \cite{matlab} mex-function.
Computation of quantities on meshes, mesh simplification as well as local thickness computations are done using \cite{matlab, gptoolbox}.
For all experiments we choose $\psi_1(x)$ to be the robust loss function of \cite{barron2019general}, for which we choose $\alpha_1 = -2$ and $c_1 = 0.15$.
For $\psi_2(x)$ we also choose the same loss function with $\alpha_2 = 0.7$ and $c_2 = 0.6$, but with a cubic bowl instead of a quadratic bowl as we want to ensure that small errors due to discretisation artefacts are not penalised.
The choice of different $\psi_1(x)$ and $\psi_2(x)$ is required since $d_\text{data}$ and $d_\text{reg}$ have different ranges.

\section{Experiments} \label{sec:experiments}
In this section we compare our method on two datasets, conduct an ablation study, %
showcase results on partial shapes and for sketch-based shape manipulation.
We emphasise that the matching of contours to 3D meshes is ill-posed: the same contour can arise from different configurations, \ie the ground-truth is not necessarily unique, the space of solutions that seem natural is even bigger, and evaluation criteria that capture this non-uniqueness do not exist.

\noindent\paragraph{Datasets.}
We evaluate on the following two datasets:
\begin{itemize}%
    \item \textbf{TOSCA 2D-3D}~\cite{lahner2016}: 80 shapes of 9 different classes (humans, animals, etc.) in different poses. For each class exists at least one 2D shape.
    \item \textbf{FAUST 2D-3D}~\cite{lahner2016}:  100 human shapes in different poses subdivided into 10 classes. Each class has one 2D shape. Ground-truth correspondences between 2D and 3D are available for all instances.
\end{itemize}
Both datasets contain segmentation information across all shapes which form consistent 2D part to 3D part mappings.
\begin{figure*}[t]
    \hspace{-0.9cm}
    \begin{tabular}{cc|cc}
    \setlength{\tabcolsep}{0pt}
        \newcommand{\pckLineWidth}{4pt}
\newcommand{\plotWidth}{\columnwidth}
\newcommand{\plotHeight}{0.75\columnwidth}
\newcommand{\pckTitle}{TOSCA}

\pgfplotsset{%
    every axis/.style={line width=0.01pt},
    label style = {font=\sffamily\Large},
    tick label style = {font=\sffamily\large},
    title style =  {font=\Large\sffamily},
    legend style={  fill= gray!10,
                    fill opacity=0.6, 
                    font=\sffamily\Large,
                    draw=gray!20, %
                    text opacity=1}
}
\begin{tikzpicture}[scale=0.5, transform shape]
	\begin{axis}[
		width=\plotWidth,
		height=\plotHeight,
		grid=major,
		title=\pckTitle,
		legend style={
			at={(0.97,0.03)},
			anchor=south east,
			legend columns=1},
		legend cell align={left},
		ylabel={{\sffamily\Large$\%$ Correct Segment}},
        xlabel={Geodesic Error Threshold},
		xmin=0,
        xmax=1,
        ylabel near ticks,
        xtick={0, 0.25, 0.5, 0.75, 1},
		ymin=60,
        ymax=103,
        ytick={0, 10, 20, 30, 40, 50, 60, 70, 80, 90, 100},
	]
	
	\addplot [color=mycolor1, smooth, line width=\pckLineWidth]
    table[row sep=crcr]{%
0 61.2096\\
0.01 62.5748\\
0.02 64.6432\\
0.03 66.1468\\
0.04 67.295\\
0.05 68.189\\
0.06 68.7949\\
0.07 69.4569\\
0.08 70.1077\\
0.09 70.6987\\
0.1 71.2074\\
0.11 71.6712\\
0.12 72.2247\\
0.13 72.6848\\
0.14 73.0925\\
0.15 73.5189\\
0.16 74.0013\\
0.17 74.5399\\
0.18 75.0411\\
0.19 75.561\\
0.2 75.9725\\
0.21 76.5335\\
0.22 77.1507\\
0.23 77.7304\\
0.24 78.3027\\
0.25 78.8188\\
0.26 79.3387\\
0.27 79.7838\\
0.28 80.2551\\
0.29 80.8498\\
0.3 81.4782\\
0.31 82.1626\\
0.32 82.8247\\
0.33 83.3857\\
0.34 83.9729\\
0.35 84.5227\\
0.36 85.095\\
0.37 85.6598\\
0.38 86.1909\\
0.39 86.6472\\
0.4 87.212\\
0.41 87.8591\\
0.42 88.637\\
0.43 89.3178\\
0.44 90.0845\\
0.45 90.9224\\
0.46 91.6629\\
0.47 92.3175\\
0.48 92.9645\\
0.49 93.4433\\
0.5 93.9969\\
0.51 94.4158\\
0.52 94.8197\\
0.53 95.3134\\
0.54 95.7361\\
0.55 96.1924\\
0.56 96.6525\\
0.57 97.1798\\
0.58 97.6661\\
0.59 98.0551\\
0.6 98.4328\\
0.61 98.747\\
0.62 98.9602\\
0.63 99.1472\\
0.64 99.3193\\
0.65 99.4165\\
0.66 99.4951\\
0.67 99.6185\\
0.68 99.697\\
0.69 99.7905\\
0.7 99.8728\\
0.71 99.9289\\
0.72 99.9813\\
0.73 100\\
0.99 100\\
    };
    \addlegendentry{\textcolor{black}{L{\"a}hner et al.: 0.89}}
    
    \addplot [color=mycolor3, smooth, line width=\pckLineWidth]
          table[row sep=crcr]{%
0 68.3598\\
0.01 70.054\\
0.02 72.4929\\
0.03 74.5532\\
0.04 75.9154\\
0.05 77.1596\\
0.06 78.3977\\
0.07 79.4682\\
0.08 80.1353\\
0.09 80.6969\\
0.1 81.2368\\
0.11 81.876\\
0.12 82.4625\\
0.13 82.8565\\
0.14 83.3406\\
0.15 83.865\\
0.16 84.2032\\
0.17 84.4638\\
0.18 84.6934\\
0.19 84.8175\\
0.2 84.9975\\
0.21 85.224\\
0.22 85.5157\\
0.23 86.0804\\
0.24 86.5428\\
0.25 86.9213\\
0.26 87.2223\\
0.27 87.576\\
0.28 88.0135\\
0.29 88.3176\\
0.3 88.7241\\
0.31 88.9971\\
0.32 89.4191\\
0.33 89.8039\\
0.34 90.1297\\
0.35 90.6013\\
0.36 90.9985\\
0.37 91.5198\\
0.38 91.9697\\
0.39 92.5406\\
0.4 93.003\\
0.41 93.4095\\
0.42 93.8501\\
0.43 94.2224\\
0.44 94.6103\\
0.45 94.9857\\
0.46 95.4232\\
0.47 95.777\\
0.48 96.1772\\
0.49 96.5341\\
0.5 96.8599\\
0.51 97.2012\\
0.52 97.5487\\
0.53 97.7349\\
0.54 97.9087\\
0.55 98.0514\\
0.56 98.2407\\
0.57 98.5541\\
0.58 98.7619\\
0.59 99.0412\\
0.6 99.2119\\
0.61 99.3298\\
0.62 99.4477\\
0.63 99.5439\\
0.64 99.6339\\
0.65 99.699\\
0.66 99.7735\\
0.67 99.8262\\
0.68 99.8976\\
0.69 99.9504\\
0.7 99.9783\\
0.71 100\\
0.99 100\\
        };
        \addlegendentry{\textcolor{black}{Ours: 0.93}}

	\addplot [color=mycolor11, smooth, dashed, line width=\pckLineWidth]
    table[row sep=crcr]{%
0 99.901\\
0.01 99.9116\\
0.02 99.9682\\
0.03 99.9965\\
0.04 100\\
0.99 100\\
    };
    \addlegendentry{\textcolor{black}{(L{\"a}hner-Seg): 1}}

    \addplot [color=mycolor33, smooth, dotted, line width=\pckLineWidth]
    table[row sep=crcr]{%
0 99.8718\\
0.01 99.9078\\
0.02 99.9978\\
0.03 100\\
0.99 100\\
    };
    \addlegendentry{\textcolor{black}{(Ours-Seg): 1}}
        
	\end{axis}
\end{tikzpicture} &
        \hspace{-1cm}
        \newcommand{\pckLineWidth}{4pt}
\newcommand{\plotWidth}{\columnwidth}
\newcommand{\plotHeight}{0.75\columnwidth}
\newcommand{\pckTitle}{FAUST}

\pgfplotsset{%
    every axis/.style={line width=0.01pt},
    label style = {font=\sffamily\Large},
    tick label style = {font=\sffamily\large},
    title style =  {font=\Large\sffamily},
    legend style={  fill= gray!10,
                    fill opacity=0.6, 
                    font=\sffamily\Large,
                    draw=gray!20, %
                    text opacity=1}
}
\begin{tikzpicture}[scale=0.5, transform shape]
	\begin{axis}[
		width=\plotWidth,
		height=\plotHeight,
		grid=major,
		title=\pckTitle,
		legend style={
			at={(0.97,0.03)},
			anchor=south east,
			legend columns=1},
		legend cell align={left},
		ylabel={{\sffamily\Large$\%$ Correct Segment}},
        xlabel={Geodesic Error Threshold},
		xmin=0,
        xmax=1,
        ylabel near ticks,
        xtick={0, 0.25, 0.5, 0.75, 1},
		ymin=60,
        ymax=103,
        ytick={0, 10, 20, 30, 40, 50, 60, 70, 80, 90, 100},
	]
	
	\addplot [color=mycolor1, smooth, line width=\pckLineWidth]
    table[row sep=crcr]{%
0 69.5491\\
0.01 70.6157\\
0.02 71.9603\\
0.03 73.0467\\
0.04 73.8668\\
0.05 74.4504\\
0.06 74.8585\\
0.07 75.2371\\
0.08 75.5269\\
0.09 75.795\\
0.1 76.0375\\
0.11 76.2998\\
0.12 76.3589\\
0.13 76.3806\\
0.14 76.489\\
0.15 76.493\\
0.16 76.6073\\
0.17 76.6132\\
0.18 76.6606\\
0.19 76.6901\\
0.2 77.0056\\
0.21 77.2067\\
0.22 77.4196\\
0.23 78.019\\
0.24 78.3581\\
0.25 78.9791\\
0.26 79.6416\\
0.27 80.4716\\
0.28 81.2859\\
0.29 81.9917\\
0.3 82.8533\\
0.31 83.6084\\
0.32 84.4818\\
0.33 85.5918\\
0.34 86.3252\\
0.35 86.8693\\
0.36 87.4056\\
0.37 87.9853\\
0.38 88.6221\\
0.39 89.2668\\
0.4 89.9272\\
0.41 90.635\\
0.42 91.3428\\
0.43 91.9875\\
0.44 92.715\\
0.45 93.6259\\
0.46 94.2509\\
0.47 95.1618\\
0.48 96.0095\\
0.49 96.7489\\
0.5 97.5789\\
0.51 98.2414\\
0.52 98.9531\\
0.53 99.5327\\
0.54 99.7831\\
0.55 99.933\\
0.56 99.9763\\
0.57 100\\
0.99 100\\
    };
    \addlegendentry{\textcolor{black}{L{\"a}hner et al.: 0.91}}
    
    \addplot [color=mycolor3, smooth, line width=\pckLineWidth]
          table[row sep=crcr]{%
0 78.2217\\
0.01 79.3683\\
0.02 81.924\\
0.03 84.1805\\
0.04 86.0533\\
0.05 87.5812\\
0.06 88.9173\\
0.07 90.2306\\
0.08 91.5255\\
0.09 92.768\\
0.1 93.8985\\
0.11 94.8829\\
0.12 95.5863\\
0.13 96.0637\\
0.14 96.3583\\
0.15 96.5182\\
0.16 96.6141\\
0.17 96.7055\\
0.18 96.7922\\
0.19 96.8767\\
0.2 96.9795\\
0.21 97.0663\\
0.22 97.144\\
0.23 97.2125\\
0.24 97.2719\\
0.25 97.3244\\
0.26 97.4112\\
0.27 97.5117\\
0.28 97.5996\\
0.29 97.6521\\
0.3 97.7321\\
0.31 97.8074\\
0.32 97.9057\\
0.33 98.0381\\
0.34 98.1272\\
0.35 98.2163\\
0.36 98.319\\
0.37 98.4241\\
0.38 98.5977\\
0.39 98.6936\\
0.4 98.8717\\
0.41 98.9905\\
0.42 99.1344\\
0.43 99.2349\\
0.44 99.3354\\
0.45 99.477\\
0.46 99.5957\\
0.47 99.7145\\
0.48 99.8219\\
0.49 99.9064\\
0.5 99.9543\\
0.51 99.9817\\
0.52 99.9931\\
0.53 100\\
0.99 100\\
        };
        \addlegendentry{\textcolor{black}{Ours: 0.98}}

	\addplot [color=mycolor11, smooth, dashed, line width=\pckLineWidth]
    table[row sep=crcr]{%
0 100\\
0.99 100\\
    };
    \addlegendentry{\textcolor{black}{(L{\"a}hner-Seg): 1}}

    \addplot [color=mycolor33, smooth, dotted, line width=\pckLineWidth]
    table[row sep=crcr]{%
0 99.8718\\
0.01 99.9078\\
0.02 99.9978\\
0.03 100\\
0.99 100\\
    };
    \addlegendentry{\textcolor{black}{(Ours-Seg): 1}}
        
	\end{axis}
\end{tikzpicture}&
        \hspace{-0.6cm}
        \newcommand{\pckLineWidth}{4pt}
\newcommand{\plotWidth}{\columnwidth}
\newcommand{\plotHeight}{0.75\columnwidth}
\newcommand{\pckTitle}{FAUST}

\pgfplotsset{%
    every axis/.style={line width=0.01pt},
    label style = {font=\sffamily\Large},
    tick label style = {font=\sffamily\large},
    title style =  {font=\Large\sffamily},
    legend style={  fill= gray!10,
                    fill opacity=0.6, 
                    font=\sffamily\Large,
                    draw=gray!20, %
                    text opacity=1}
}
\begin{tikzpicture}[scale=0.5, transform shape]
	\begin{axis}[
		width=\plotWidth,
		height=\plotHeight,
		grid=major,
		title=\pckTitle,
		legend style={
			at={(0.97,0.03)},
			anchor=south east,
			legend columns=1},
		legend cell align={left},
		ylabel={{\sffamily\Large$\%$ Correct Matchings}},
        xlabel={Geodesic Error Threshold},
		xmin=0,
        xmax=1,
        ylabel near ticks,
        xtick={0, 0.25, 0.5, 0.75, 1},
		ymin=0,
        ymax=103,
        ytick={0, 20, 40, 60, 80, 100},
	]
	
	\addplot [color=mycolor1, smooth, line width=\pckLineWidth]
    table[row sep=crcr]{%
0 0.97083\\
0.01 5.9208\\
0.02 14.72\\
0.03 23.317\\
0.04 30.9559\\
0.05 37.096\\
0.06 41.5244\\
0.07 44.9308\\
0.08 47.4878\\
0.09 49.5678\\
0.1 51.0177\\
0.11 52.1886\\
0.12 53.251\\
0.13 54.1197\\
0.14 54.8031\\
0.15 55.4844\\
0.16 56.4041\\
0.17 57.4601\\
0.18 58.5693\\
0.19 59.7679\\
0.2 60.9836\\
0.21 61.9885\\
0.22 63.1978\\
0.23 64.1814\\
0.24 65.1778\\
0.25 66.072\\
0.26 66.8022\\
0.27 67.4516\\
0.28 68.1137\\
0.29 68.6523\\
0.3 69.1846\\
0.31 69.753\\
0.32 70.3683\\
0.33 70.9857\\
0.34 71.6692\\
0.35 72.3611\\
0.36 73.102\\
0.37 73.8407\\
0.38 74.5156\\
0.39 75.1756\\
0.4 75.7739\\
0.41 76.3572\\
0.42 76.947\\
0.43 77.5197\\
0.44 78.0711\\
0.45 78.5714\\
0.46 79.1016\\
0.47 79.6253\\
0.48 80.1618\\
0.49 80.6898\\
0.5 81.2008\\
0.51 81.7352\\
0.52 82.3504\\
0.53 83.0764\\
0.54 83.7343\\
0.55 84.3943\\
0.56 84.9713\\
0.57 85.5653\\
0.58 86.1635\\
0.59 86.7213\\
0.6 87.3132\\
0.61 87.9072\\
0.62 88.5438\\
0.63 89.1505\\
0.64 89.819\\
0.65 90.4152\\
0.66 91.0879\\
0.67 91.7692\\
0.68 92.4888\\
0.69 93.2106\\
0.7 93.9046\\
0.71 94.5816\\
0.72 95.3204\\
0.73 95.9953\\
0.74 96.7192\\
0.75 97.3621\\
0.76 98.0072\\
0.77 98.6247\\
0.78 99.1527\\
0.79 99.6146\\
0.8 99.8552\\
0.81 99.951\\
0.82 99.9894\\
0.83 100\\
0.99 100\\
    };
    \addlegendentry{\textcolor{black}{L{\"a}hner et al.: 0.77}}
    
    \addplot [color=mycolor3, smooth, line width=\pckLineWidth]
          table[row sep=crcr]{%
0 6.0414\\
0.01 10.9102\\
0.02 27.3681\\
0.03 37.9835\\
0.04 46.6101\\
0.05 52.567\\
0.06 56.7066\\
0.07 60.0805\\
0.08 62.636\\
0.09 64.803\\
0.1 66.435\\
0.11 67.6419\\
0.12 68.3757\\
0.13 68.858\\
0.14 69.1254\\
0.15 69.4112\\
0.16 69.7289\\
0.17 70.2821\\
0.18 70.9221\\
0.19 71.6696\\
0.2 72.561\\
0.21 73.6285\\
0.22 74.7897\\
0.23 75.8343\\
0.24 76.6915\\
0.25 77.4435\\
0.26 78.1636\\
0.27 78.8905\\
0.28 79.4505\\
0.29 79.9534\\
0.3 80.4905\\
0.31 80.9797\\
0.32 81.5123\\
0.33 82.054\\
0.34 82.6255\\
0.35 83.1855\\
0.36 83.7341\\
0.37 84.2393\\
0.38 84.717\\
0.39 85.2405\\
0.4 85.7411\\
0.41 86.2417\\
0.42 86.7285\\
0.43 87.1857\\
0.44 87.652\\
0.45 88.0977\\
0.46 88.5732\\
0.47 89.0418\\
0.48 89.5149\\
0.49 89.995\\
0.5 90.4498\\
0.51 90.9002\\
0.52 91.3116\\
0.53 91.7505\\
0.54 92.2008\\
0.55 92.5985\\
0.56 93.042\\
0.57 93.4557\\
0.58 93.8603\\
0.59 94.2832\\
0.6 94.706\\
0.61 95.1038\\
0.62 95.5129\\
0.63 95.9404\\
0.64 96.345\\
0.65 96.6787\\
0.66 97.0719\\
0.67 97.4536\\
0.68 97.8353\\
0.69 98.1553\\
0.7 98.4479\\
0.71 98.7634\\
0.72 99.0194\\
0.73 99.2548\\
0.74 99.4491\\
0.75 99.6114\\
0.76 99.7783\\
0.77 99.856\\
0.78 99.9291\\
0.79 99.9657\\
0.8 99.9954\\
0.81 100\\
0.99 100\\
        };
        \addlegendentry{\textcolor{black}{Ours: 0.85}}

	\addplot [color=mycolor11, smooth, dashed, line width=\pckLineWidth]
    table[row sep=crcr]{%
0 1.4431\\
0.01 9.7602\\
0.02 23.1411\\
0.03 37.3515\\
0.04 52.2057\\
0.05 64.0959\\
0.06 72.884\\
0.07 79.1704\\
0.08 83.7459\\
0.09 87.1765\\
0.1 89.7516\\
0.11 91.7131\\
0.12 93.3549\\
0.13 94.5949\\
0.14 95.5196\\
0.15 96.2843\\
0.16 96.8373\\
0.17 97.3212\\
0.18 97.7231\\
0.19 98.1033\\
0.2 98.3193\\
0.21 98.4619\\
0.22 98.639\\
0.23 98.8594\\
0.24 99.1186\\
0.25 99.218\\
0.26 99.2957\\
0.27 99.3649\\
0.28 99.421\\
0.29 99.4513\\
0.3 99.4859\\
0.31 99.5161\\
0.32 99.542\\
0.33 99.5679\\
0.34 99.5852\\
0.35 99.5982\\
0.36 99.6111\\
0.37 99.6284\\
0.38 99.6414\\
0.39 99.6587\\
0.4 99.676\\
0.41 99.6976\\
0.42 99.7105\\
0.43 99.7364\\
0.44 99.758\\
0.45 99.784\\
0.46 99.8229\\
0.47 99.8877\\
0.48 99.9482\\
0.49 100\\
0.99 100\\
    };
    \addlegendentry{\textcolor{black}{(L{\"a}hner-Seg): 0.94}}

    \addplot [color=mycolor33, smooth, dotted, line width=\pckLineWidth]
    table[row sep=crcr]{%
0 4.4937\\
0.01 10.0262\\
0.02 26.0281\\
0.03 39.8904\\
0.04 53.4954\\
0.05 63.6693\\
0.06 71.1175\\
0.07 77.1742\\
0.08 81.7775\\
0.09 84.9178\\
0.1 87.9152\\
0.11 90.7458\\
0.12 93.5621\\
0.13 95.735\\
0.14 97.4839\\
0.15 98.5228\\
0.16 99.0374\\
0.17 99.228\\
0.18 99.3043\\
0.19 99.3567\\
0.2 99.3853\\
0.21 99.4425\\
0.22 99.4615\\
0.23 99.4901\\
0.24 99.5187\\
0.25 99.5378\\
0.26 99.5711\\
0.27 99.5997\\
0.28 99.6283\\
0.29 99.6283\\
0.3 99.6521\\
0.31 99.6617\\
0.32 99.6903\\
0.33 99.7093\\
0.34 99.7236\\
0.35 99.757\\
0.36 99.776\\
0.37 99.8046\\
0.38 99.8332\\
0.39 99.8523\\
0.4 99.8666\\
0.41 99.8809\\
0.42 99.8999\\
0.43 99.9142\\
0.44 99.919\\
0.45 99.9285\\
0.46 99.9428\\
0.47 99.9476\\
0.48 99.9571\\
0.49 99.9666\\
0.5 99.9714\\
0.51 99.9762\\
0.52 99.9762\\
0.53 99.9762\\
0.54 99.9762\\
0.55 99.9762\\
0.56 99.9762\\
0.57 99.9762\\
0.58 99.9762\\
0.59 99.9762\\
0.6 99.9809\\
0.61 99.9809\\
0.62 99.9857\\
0.63 100\\
0.99 100\\
    };
    \addlegendentry{\textcolor{black}{(Ours-Seg): 0.95}}
        
	\end{axis}
\end{tikzpicture} &
        \hspace{-1cm}
        \newcommand{\pckLineWidth}{4pt}
\newcommand{\plotWidth}{\columnwidth}
\newcommand{\plotHeight}{0.75\columnwidth}
\newcommand{\pckTitle}{FAUST w/o  Flips}

\pgfplotsset{%
    every axis/.style={line width=0.01pt},
    label style = {font=\sffamily\Large},
    tick label style = {font=\sffamily\large},
    title style =  {font=\Large\sffamily},
    legend style={  fill= gray!10,
                    fill opacity=0.6, 
                    font=\sffamily\Large,
                    draw=gray!20, %
                    text opacity=1}
}
\begin{tikzpicture}[scale=0.5, transform shape]
	\begin{axis}[
		width=\plotWidth,
		height=\plotHeight,
		grid=major,
		title=\pckTitle,
		legend style={
			at={(0.97,0.03)},
			anchor=south east,
			legend columns=1},
		legend cell align={left},
		ylabel={{\sffamily\Large$\%$ Correct Matchings}},
        xlabel={Geodesic Error Threshold},
		xmin=0,
        xmax=1,
        ylabel near ticks,
        xtick={0, 0.25, 0.5, 0.75, 1},
		ymin=0,
        ymax=103,
        ytick={0, 20, 40, 60, 80, 100},
	]
	
	\addplot [color=mycolor1, smooth, line width=\pckLineWidth]
    table[row sep=crcr]{%
0 1.0186\\
0.01 6.3802\\
0.02 15.8765\\
0.03 25.1412\\
0.04 33.3966\\
0.05 39.9157\\
0.06 44.5551\\
0.07 48.1202\\
0.08 50.7524\\
0.09 52.8591\\
0.1 54.3013\\
0.11 55.4102\\
0.12 56.4126\\
0.13 57.209\\
0.14 57.8341\\
0.15 58.4637\\
0.16 59.2972\\
0.17 60.3042\\
0.18 61.383\\
0.19 62.5313\\
0.2 63.6934\\
0.21 64.6009\\
0.22 65.7399\\
0.23 66.6705\\
0.24 67.6035\\
0.25 68.4161\\
0.26 69.0944\\
0.27 69.7217\\
0.28 70.3422\\
0.29 70.8538\\
0.3 71.3492\\
0.31 71.8955\\
0.32 72.4697\\
0.33 73.0507\\
0.34 73.699\\
0.35 74.3495\\
0.36 75.0787\\
0.37 75.7871\\
0.38 76.4122\\
0.39 77.0395\\
0.4 77.5836\\
0.41 78.1299\\
0.42 78.6809\\
0.43 79.2319\\
0.44 79.7319\\
0.45 80.2042\\
0.46 80.7135\\
0.47 81.1927\\
0.48 81.7228\\
0.49 82.2368\\
0.5 82.6813\\
0.51 83.1929\\
0.52 83.7716\\
0.53 84.4453\\
0.54 85.0634\\
0.55 85.6931\\
0.56 86.2372\\
0.57 86.7997\\
0.58 87.3275\\
0.59 87.853\\
0.6 88.4133\\
0.61 88.9712\\
0.62 89.5685\\
0.63 90.131\\
0.64 90.7445\\
0.65 91.2932\\
0.66 91.9205\\
0.67 92.5595\\
0.68 93.2285\\
0.69 93.8791\\
0.7 94.5296\\
0.71 95.1431\\
0.72 95.8538\\
0.73 96.4858\\
0.74 97.1548\\
0.75 97.7475\\
0.76 98.3031\\
0.77 98.8448\\
0.78 99.2754\\
0.79 99.6481\\
0.8 99.868\\
0.81 99.9491\\
0.82 99.9884\\
0.83 100\\
0.99 100\\
    };
    \addlegendentry{\textcolor{black}{L{\"a}hner et al.: 0.79}}
    
    \addplot [color=mycolor3, smooth, line width=\pckLineWidth]
          table[row sep=crcr]{%
0 8.6824\\
0.01 15.6203\\
0.02 39.2465\\
0.03 54.3263\\
0.04 66.408\\
0.05 74.5597\\
0.06 80.0418\\
0.07 84.3067\\
0.08 87.2848\\
0.09 89.666\\
0.1 91.3441\\
0.11 92.4452\\
0.12 93.0985\\
0.13 93.4434\\
0.14 93.5761\\
0.15 93.6988\\
0.16 93.785\\
0.17 93.921\\
0.18 94.0371\\
0.19 94.1731\\
0.2 94.3521\\
0.21 94.5677\\
0.22 94.7766\\
0.23 94.9789\\
0.24 95.2244\\
0.25 95.3968\\
0.26 95.5361\\
0.27 95.6588\\
0.28 95.7815\\
0.29 95.8677\\
0.3 95.9772\\
0.31 96.0767\\
0.32 96.2027\\
0.33 96.3155\\
0.34 96.4382\\
0.35 96.5509\\
0.36 96.6869\\
0.37 96.793\\
0.38 96.8991\\
0.39 97.0351\\
0.4 97.1479\\
0.41 97.2474\\
0.42 97.3568\\
0.43 97.4397\\
0.44 97.526\\
0.45 97.6221\\
0.46 97.7283\\
0.47 97.8112\\
0.48 97.8941\\
0.49 97.9969\\
0.5 98.0931\\
0.51 98.1727\\
0.52 98.2556\\
0.53 98.3385\\
0.54 98.4346\\
0.55 98.5176\\
0.56 98.6204\\
0.57 98.7066\\
0.58 98.7796\\
0.59 98.8658\\
0.6 98.9354\\
0.61 99.0416\\
0.62 99.1211\\
0.63 99.2173\\
0.64 99.3135\\
0.65 99.3931\\
0.66 99.4826\\
0.67 99.5556\\
0.68 99.6319\\
0.69 99.6982\\
0.7 99.7446\\
0.71 99.8242\\
0.72 99.8673\\
0.73 99.9138\\
0.74 99.9536\\
0.75 99.9735\\
0.76 99.9934\\
0.77 99.9934\\
0.78 100\\
0.99 100\\
        };
        \addlegendentry{\textcolor{black}{Ours: 0.94}}

	\addplot [color=mycolor11, smooth, dashed, line width=\pckLineWidth]
    table[row sep=crcr]{%
0 1.4431\\
0.01 9.7602\\
0.02 23.1411\\
0.03 37.3515\\
0.04 52.2057\\
0.05 64.0959\\
0.06 72.884\\
0.07 79.1704\\
0.08 83.7459\\
0.09 87.1765\\
0.1 89.7516\\
0.11 91.7131\\
0.12 93.3549\\
0.13 94.5949\\
0.14 95.5196\\
0.15 96.2843\\
0.16 96.8373\\
0.17 97.3212\\
0.18 97.7231\\
0.19 98.1033\\
0.2 98.3193\\
0.21 98.4619\\
0.22 98.639\\
0.23 98.8594\\
0.24 99.1186\\
0.25 99.218\\
0.26 99.2957\\
0.27 99.3649\\
0.28 99.421\\
0.29 99.4513\\
0.3 99.4859\\
0.31 99.5161\\
0.32 99.542\\
0.33 99.5679\\
0.34 99.5852\\
0.35 99.5982\\
0.36 99.6111\\
0.37 99.6284\\
0.38 99.6414\\
0.39 99.6587\\
0.4 99.676\\
0.41 99.6976\\
0.42 99.7105\\
0.43 99.7364\\
0.44 99.758\\
0.45 99.784\\
0.46 99.8229\\
0.47 99.8877\\
0.48 99.9482\\
0.49 100\\
0.99 100\\
    };
    \addlegendentry{\textcolor{black}{(L{\"a}hner-Seg): 0.94}}

    \addplot [color=mycolor33, smooth, dotted, line width=\pckLineWidth]
    table[row sep=crcr]{%
0 4.4937\\
0.01 10.0262\\
0.02 26.0281\\
0.03 39.8904\\
0.04 53.4954\\
0.05 63.6693\\
0.06 71.1175\\
0.07 77.1742\\
0.08 81.7775\\
0.09 84.9178\\
0.1 87.9152\\
0.11 90.7458\\
0.12 93.5621\\
0.13 95.735\\
0.14 97.4839\\
0.15 98.5228\\
0.16 99.0374\\
0.17 99.228\\
0.18 99.3043\\
0.19 99.3567\\
0.2 99.3853\\
0.21 99.4425\\
0.22 99.4615\\
0.23 99.4901\\
0.24 99.5187\\
0.25 99.5378\\
0.26 99.5711\\
0.27 99.5997\\
0.28 99.6283\\
0.29 99.6283\\
0.3 99.6521\\
0.31 99.6617\\
0.32 99.6903\\
0.33 99.7093\\
0.34 99.7236\\
0.35 99.757\\
0.36 99.776\\
0.37 99.8046\\
0.38 99.8332\\
0.39 99.8523\\
0.4 99.8666\\
0.41 99.8809\\
0.42 99.8999\\
0.43 99.9142\\
0.44 99.919\\
0.45 99.9285\\
0.46 99.9428\\
0.47 99.9476\\
0.48 99.9571\\
0.49 99.9666\\
0.5 99.9714\\
0.51 99.9762\\
0.52 99.9762\\
0.53 99.9762\\
0.54 99.9762\\
0.55 99.9762\\
0.56 99.9762\\
0.57 99.9762\\
0.58 99.9762\\
0.59 99.9762\\
0.6 99.9809\\
0.61 99.9809\\
0.62 99.9857\\
0.63 100\\
0.99 100\\
    };
    \addlegendentry{\textcolor{black}{(Ours-Seg): 0.95}}
        
	\end{axis}
\end{tikzpicture}
    \end{tabular}
    \vspace{-0.3cm}
    \caption{\textbf{Quantitative comparison} for the FAUST and TOSCA datasets. \textbf{Left:} Cumulative segmentation errors. The y-axis shows the percentage of points in the correct segment, and the x-axis the geodesic error threshold. Vacuously, when integrating the segmentation information into the optimisation (methods with suffix `-Seg', dashed lines), the results are `perfect' for both methods. \textbf{Right:} Cumulative geodesic errors on FAUST with and without left-right flips (manually removed for all approaches), which confirms that in many cases our method finds plausible solutions but does not resolve the intrinsic symmetry ambiguity. The y-axis shows the percentage of points below the x-axis threshold. We can see that our method consistently outperforms \laehneretal~\cite{lahner2016}. Scores shown in the legends are respective areas under the curves.}
    \label{fig:quantitive-seg-pck}
\end{figure*}
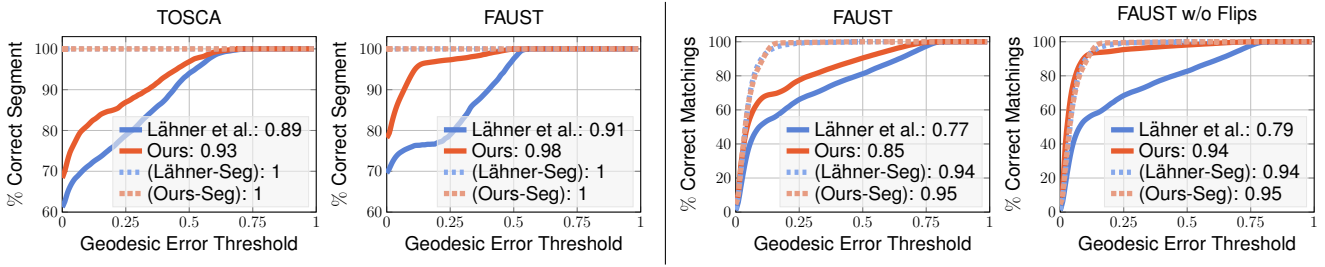

\paragraph{Competing Approach.}
The only other method able to produce continuous matchings between 2D contours and 3D shapes is \cite{lahner2016}.
Due to their weaker model expressiveness that prevents the incorporation of a deformation prior,
they use global spectral features and a pre-matched segmentation as additional feature in order to prevent degenerate solutions (\eg collapsing).
To enable a fair comparison, for both methods we provide results with and without this pre-matching.
However, we consider the pre-matching as unrealistic prior knowledge, and therefore regard the cases without  pre-matching as main results. As we show in \cref{fig:teaser}, our results are superior without the segmentation term even in comparison to~\cite{lahner2016} using the segmentation term.

\subsection{Matching}
Next we evaluate our approach on the task of 2D-3D shape matching.
First, we introduce a new error metric designed for the ambiguous setting of matching a contour onto a mesh.
Subsequently, we compare quantitatively and qualitatively to the approach by \laehneretal~\cite{lahner2016}.

\subsubsection{Error Metric}
We use two different error metrics: a) geodesic error and b) segmentation error. 
We only evaluate the geodesic error on FAUST due to the lack of 2D-3D ground truth correspondences in the TOSCA dataset. 
Additionally, there exist many valid matchings that may not correspond to the ground truth because the problem is ill-posed as explained above. %
Hence, we aim to derive a more robust quantitative evaluation for 2D-3D matchings. 
For that, we utilise part-based shape segmentations, which are available for all classes in the FAUST and TOSCA datasets and are generally consistent between 2D and 3D shapes.
We argue that a good solution must have the same segmentation in the target domain, \ie on the 3D shape, as in the source domain, \ie on the 2D shape.
For both we plot the cumulative curves measuring for each geodesic error value the percentage of matches with an error lower than this.

\textbf{Geodesic Error.} Let $(x,y) \in \mathcal{C} \subset \contourVerts\ \times \meshVerts$  %
be a computed match and $\hat y$ be the ground-truth match of $x$. 
The normalised geodesic error of this matching is defined as
\begin{equation}
    \varepsilon_\text{geo}(x,y) = \frac{\text{dist}_\mesh (y, \hat y)}{\text{diam}(\mesh)}.
\end{equation}
Here $\text{dist}_\mesh : \mesh \times \mesh \rightarrow \mathbb{R}^+_0$ is the geodesic distance on $\mesh$ and $\text{diam}(\mesh) = \underset{x, y\in\mesh}{\max}\text{dist}_\mesh(x,y)$.

\textbf{Segmentation Error.}
Let $\sigma_\contour(x)$ be the source segment of its matched point $y \in \mesh$ and let $\sigma_\mesh(y)$ be its target segment.
We define the segmentation error as
\begin{equation}
    \varepsilon_{\text{seg}}(x,y) = \underset{
    y'\in \mesh
    }
    {\min} \frac{\text{dist}_\mesh (y, y')}{\text{diam}(\mesh)}\; \text{s.t.}\; \sigma_\mesh(y') = \sigma_\contour(x). 
\end{equation}
For shapes with symmetries %
or other ambiguities, %
we choose the best of all plausible segmentation combinations.
\begin{figure}[t]
    \hspace{-0.6cm}
	\newcommand{\widthQualF}{2.3cm}
\newcommand{\heightQualF}{2.3cm}
\newcommand{\lrflip}{{\color{cPINK} $\nwarrow$}}
\begin{tabular}{c}
    \begin{tabular}{ccccc}%
        \includegraphics[height=\heightQualF,width=\widthQualF,keepaspectratio]{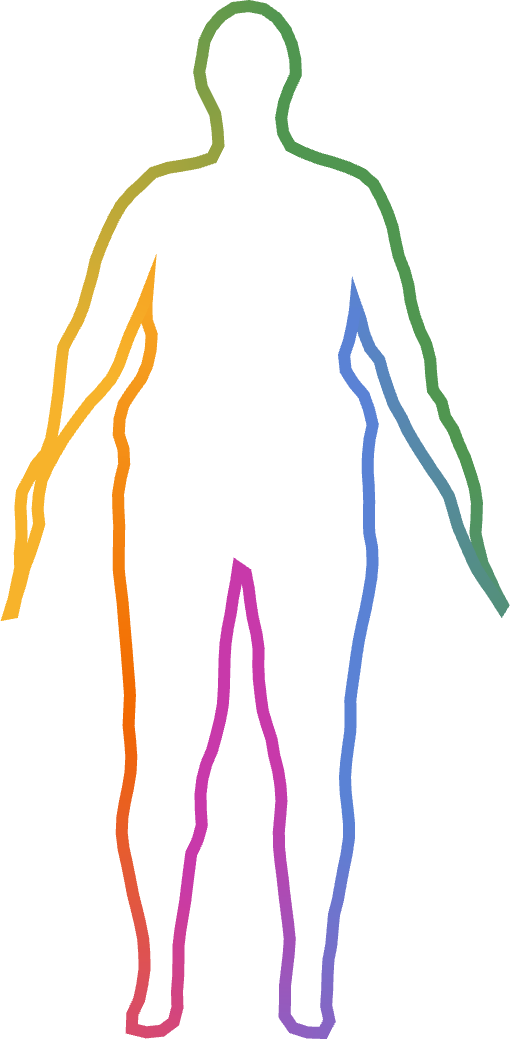} &
        \includegraphics[height=\heightQualF,width=\widthQualF,keepaspectratio]{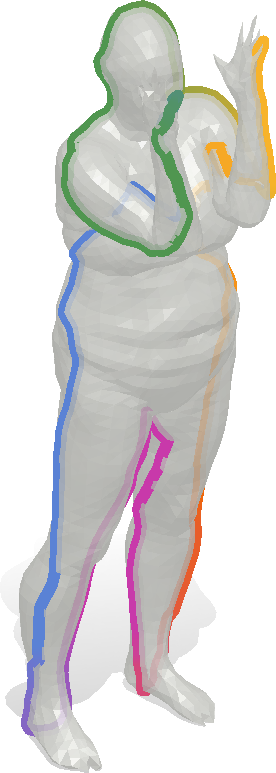}\lrflip &
        \hspace{-0.4cm}
        \includegraphics[trim={0 0 1.3cm 0}, height=\heightQualF,width=\widthQualF,keepaspectratio]{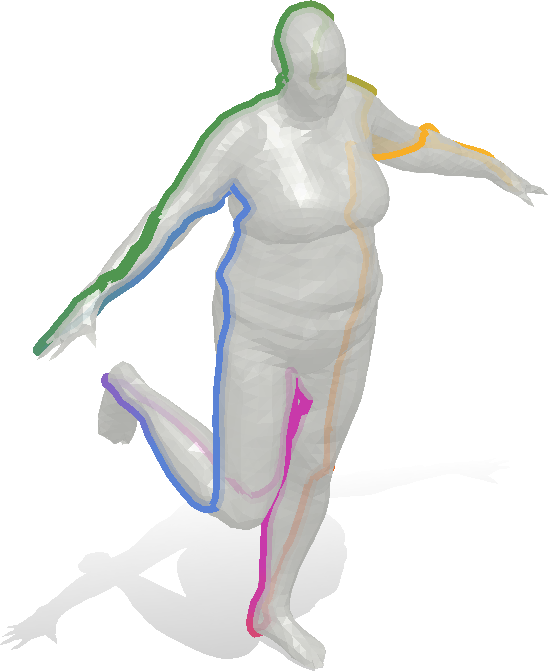}\lrflip \hspace{0.3cm}&
        \includegraphics[height=\heightQualF,width=\widthQualF,keepaspectratio]{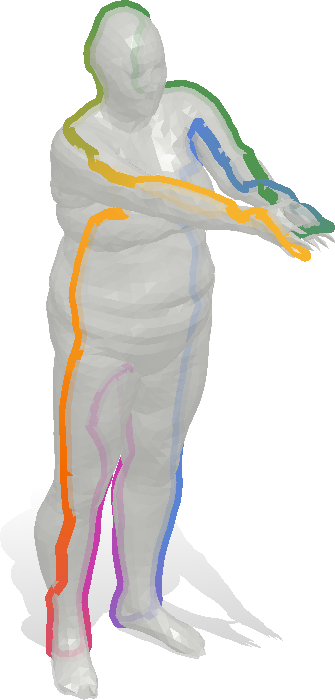}&
        \hspace{-0.35cm}
        \includegraphics[height=\heightQualF,width=\widthQualF,keepaspectratio]{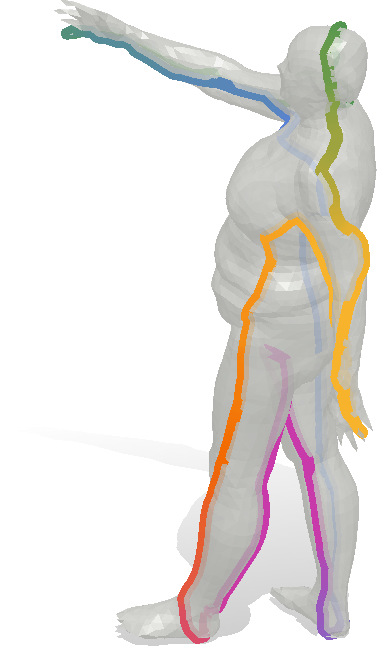}\lrflip
        \\
        \includegraphics[height=\heightQualF,width=\widthQualF,keepaspectratio]{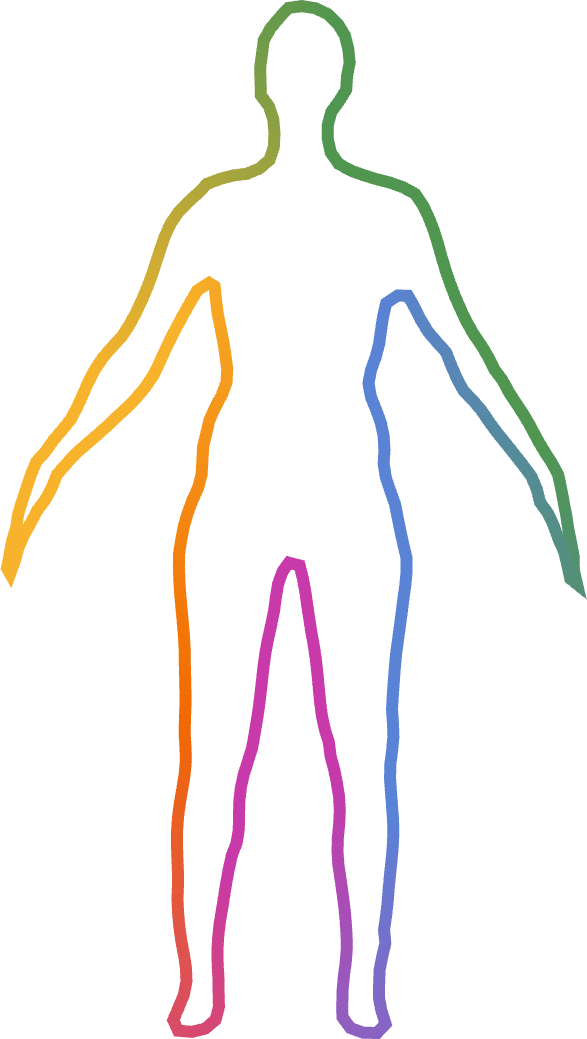} &
        \includegraphics[height=\heightQualF,width=\widthQualF,keepaspectratio]{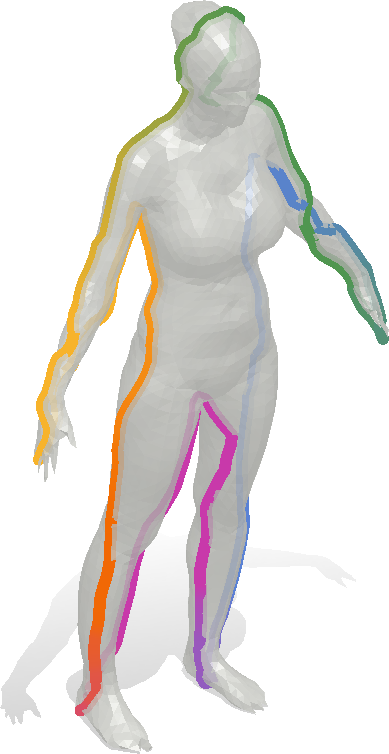} &
        \includegraphics[height=\heightQualF,width=\widthQualF,keepaspectratio]{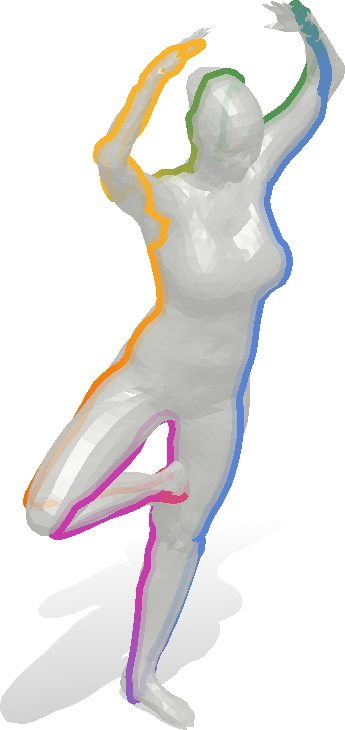}&
        \includegraphics[trim={0 0 0 0.5cm}, height=\heightQualF,width=\widthQualF,keepaspectratio]{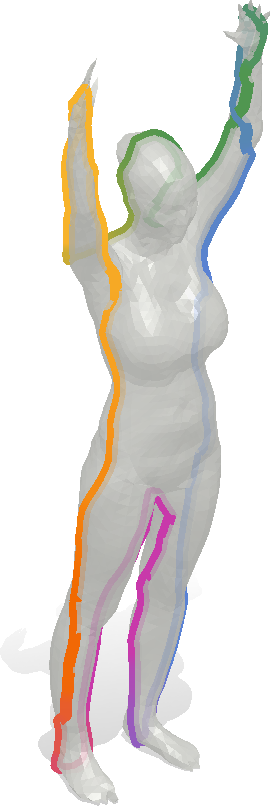} &
        \includegraphics[height=\heightQualF,width=\widthQualF,keepaspectratio]{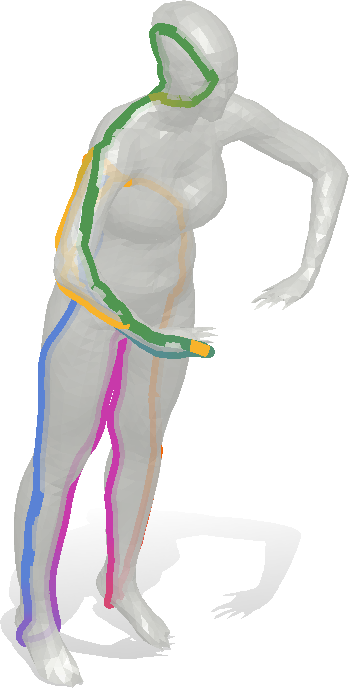}
    \end{tabular}%
\end{tabular}
	\vspace{-0.38cm}
	\caption{\textbf{Qualitative results} of our method on FAUST. We can see the occurrence of left-right-flips (indicated by \lrflip) which nevertheless can be considered as plausible matchings. %
	}
	\label{fig:qualitative-faust}
	\vspace{-0.25cm}
\end{figure}

\subsubsection{Quantitative Matching Results}
In \cref{fig:quantitive-seg-pck} (left) we show that our method outperforms the competing method by \laehneretal~\cite{lahner2016} by a great margin in terms of the segmentation error%
, both on FAUST on TOSCA.
Since for FAUST ground truth is available, in \cref{fig:quantitive-seg-pck} (right) we show the percentage of correct matchings, for which our method is superior.
In addition, when left-right-flips (which form plausible solutions that stem from shape symmmetries) are removed, our method (without pre-matching) is on par with the approach by \laehneretal that uses pre-matching.
\begin{figure*}[t]
    \hspace{-0.8cm}
	\newcommand{\widthQual}{2cm}
\newcommand{\widthCat}{3cm}
\newcommand{\widthDog}{3cm}
\newcommand{\heightQual}{1.8cm}
\newcommand{\widthHuman}{1cm}
\newcommand{\heightHuman}{1.6cm}
\newcommand{\heightCentaur}{1.6cm}
\begin{tabular}{ccccccccc}
    \rotatebox{90}{\small$\quad$2D Shape}&
    \includegraphics[trim={0.25cm 0 0.25cm 0}, width=\widthQual,height=\heightQual,keepaspectratio]{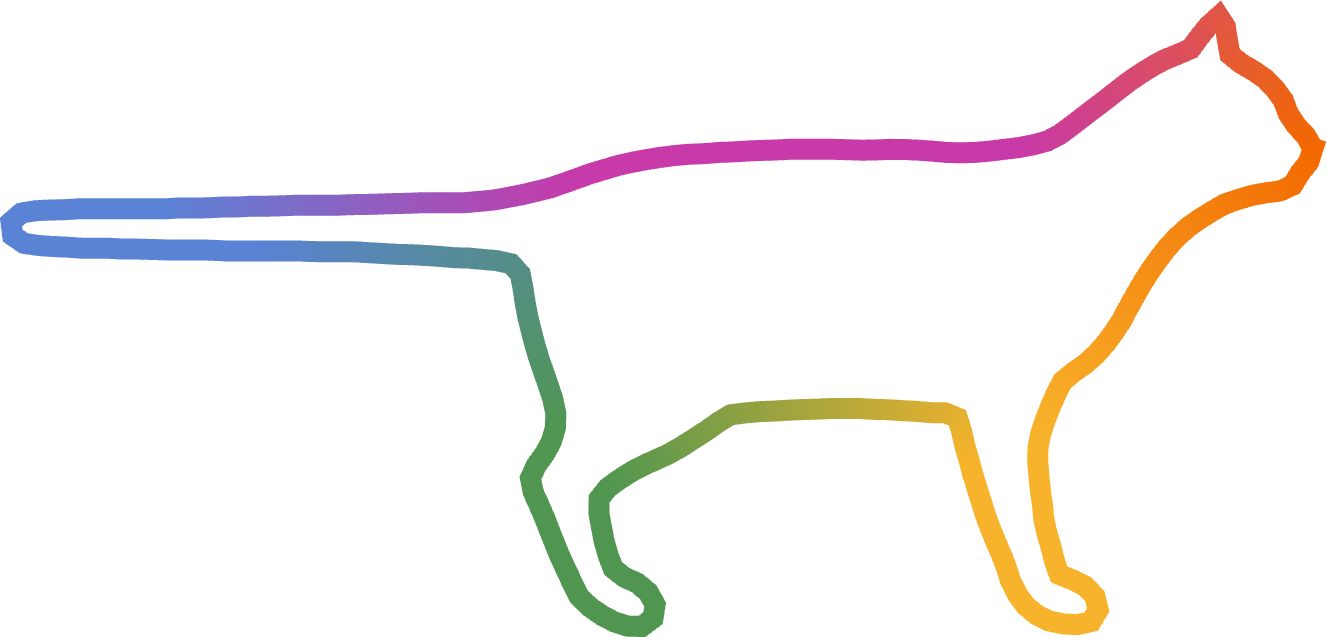} &
    \hspace{-0.1cm}
    \includegraphics[trim={0.25cm 0 0.25cm 0}, width=\widthQual,height=\heightQual,keepaspectratio]{figs/qualComparison/ours/tosca/tosca_cat_sketch_cat1_closed_N.png} &
    \includegraphics[width=\widthQual,height=\heightCentaur,keepaspectratio]{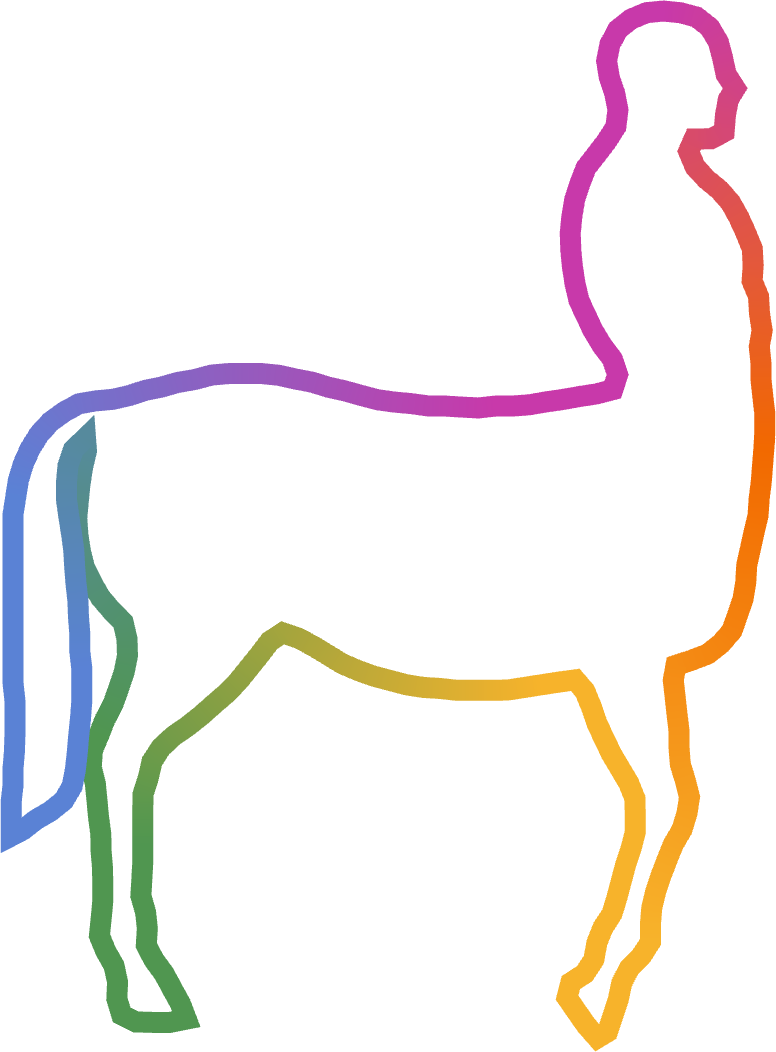} &
    \includegraphics[trim={0 0 0.25cm 0}, width=\widthQual,height=\heightQual,keepaspectratio]{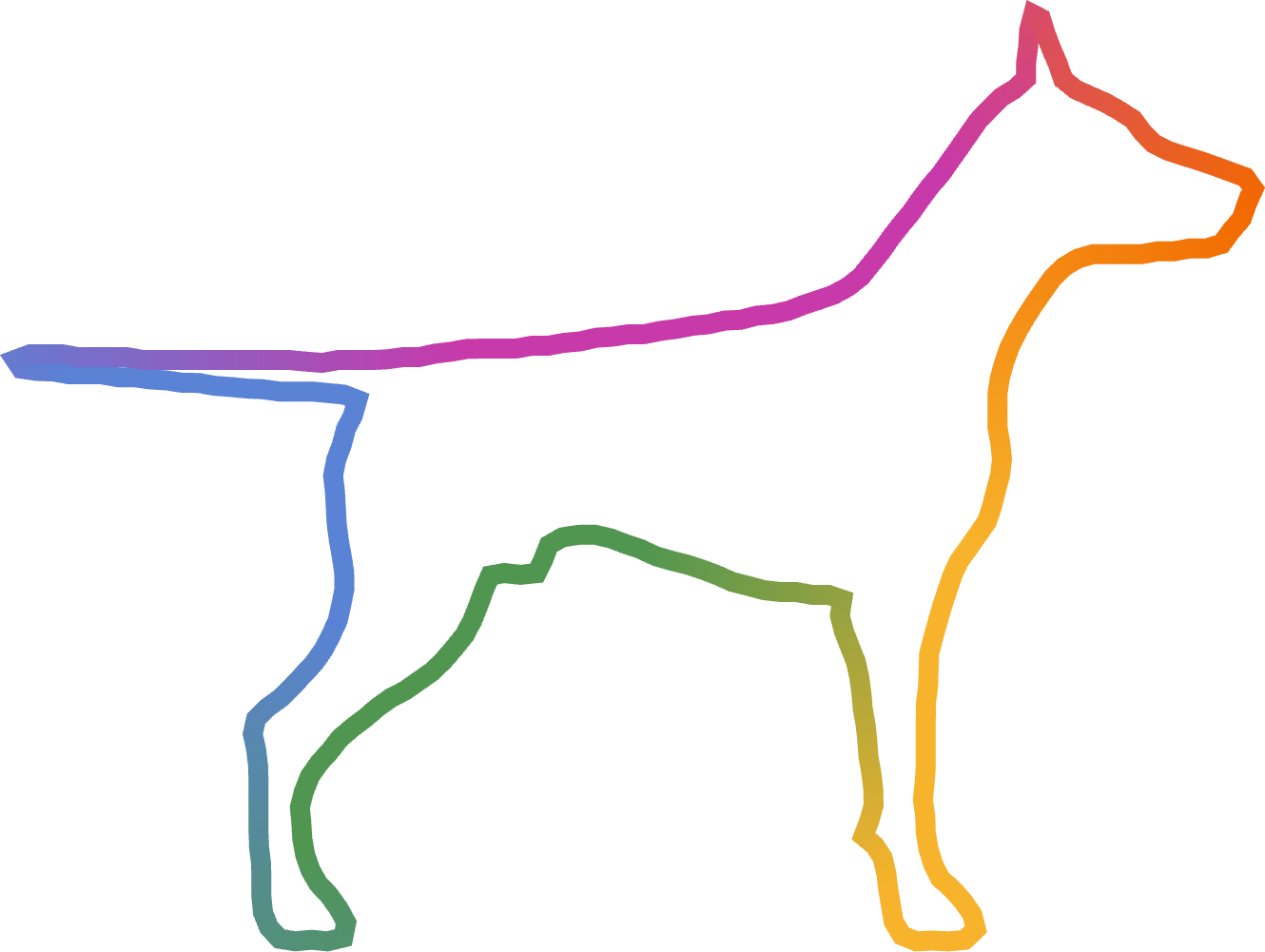} &
    \includegraphics[trim={0.25cm 0 0 0}, width=\widthQual,height=\heightQual,keepaspectratio]{figs/qualComparison/ours/tosca/tosca_dog_sketch_dog6_closed_N.png} &
    \includegraphics[width=\widthQual,height=\heightQual,keepaspectratio]{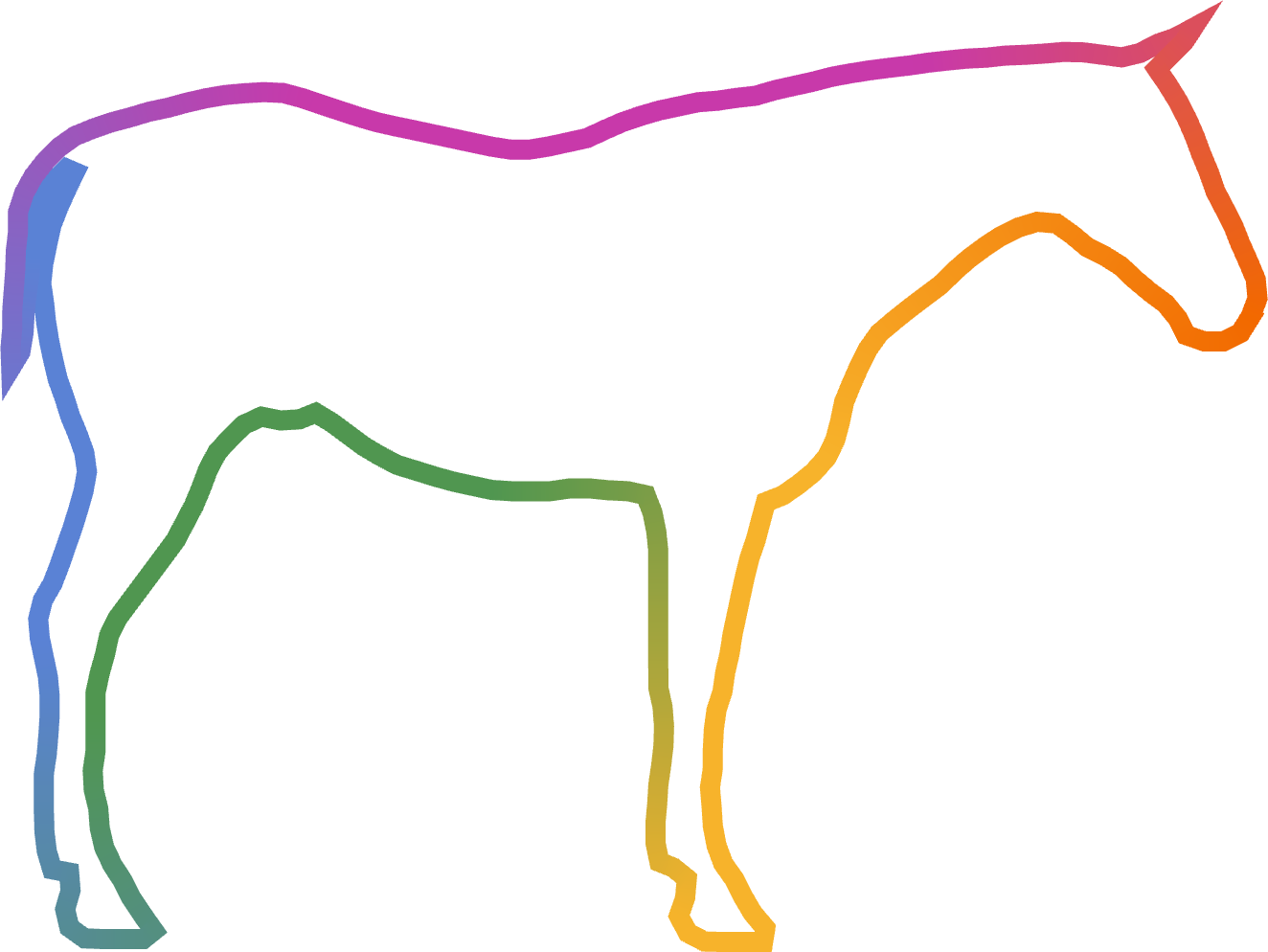} &
    \multicolumn{2}{c}{
        \includegraphics[width=\widthQual,height=\heightHuman,keepaspectratio]{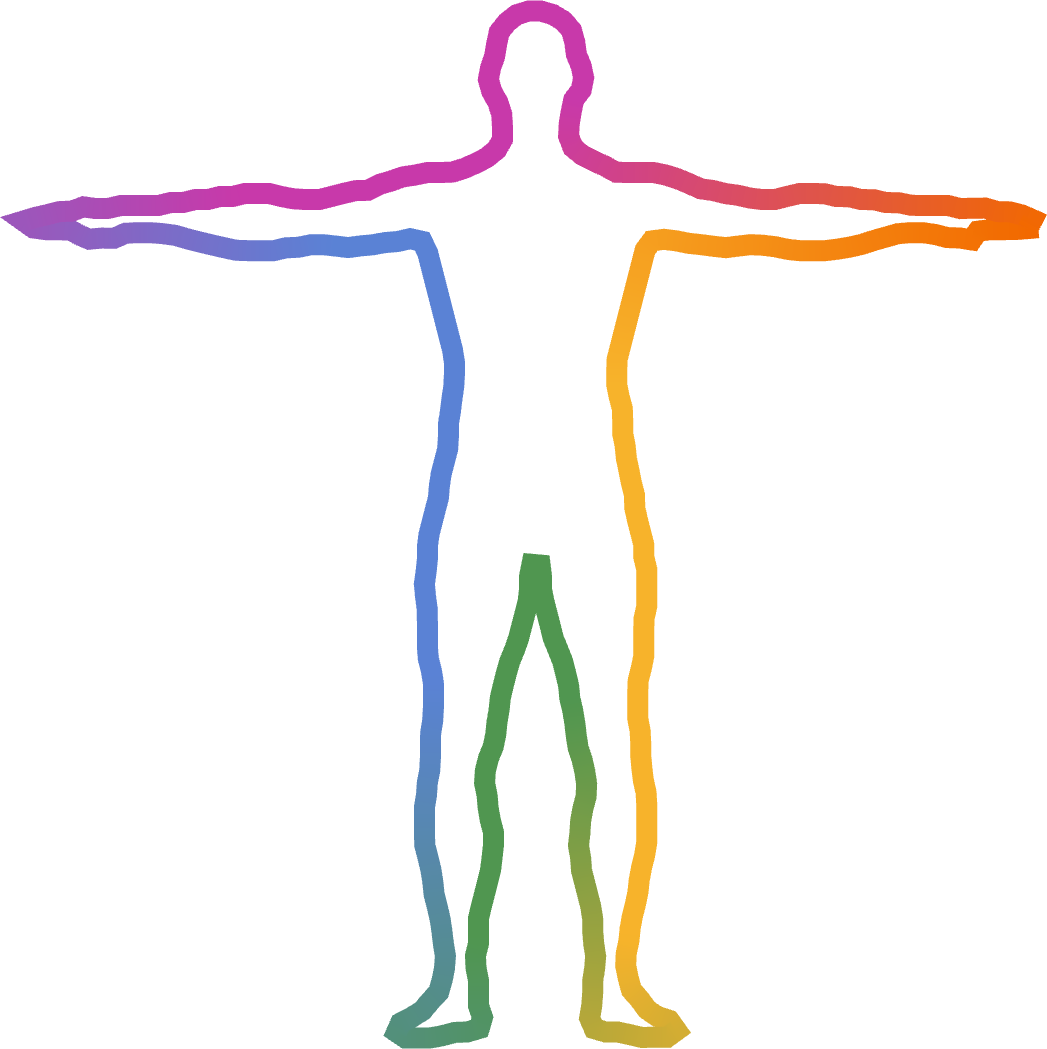}
    }\\
    \rotatebox{90}{\small\laehner~\etal}&
    \includegraphics[width=\widthQual,height=\heightQual,keepaspectratio]{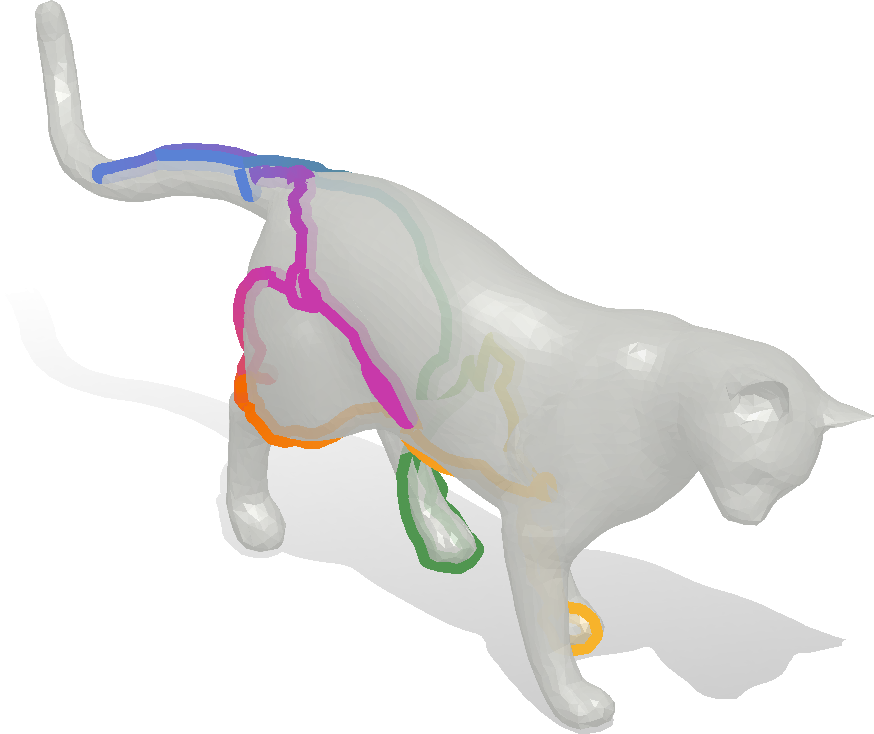} &
    \hspace{-0.1cm}
    \includegraphics[width=\widthQual,height=\heightQual,keepaspectratio]{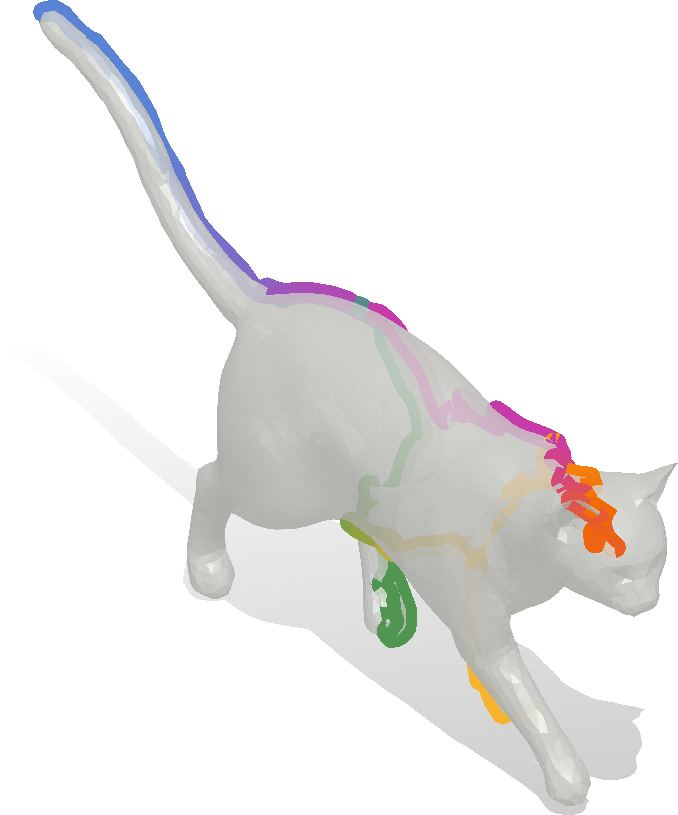} &
    \includegraphics[width=\widthQual,height=\heightQual,keepaspectratio]{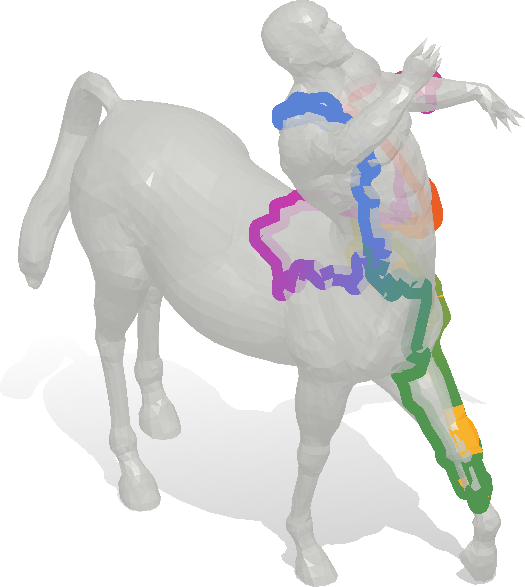} &
    \includegraphics[width=\widthQual,height=\heightQual,keepaspectratio]{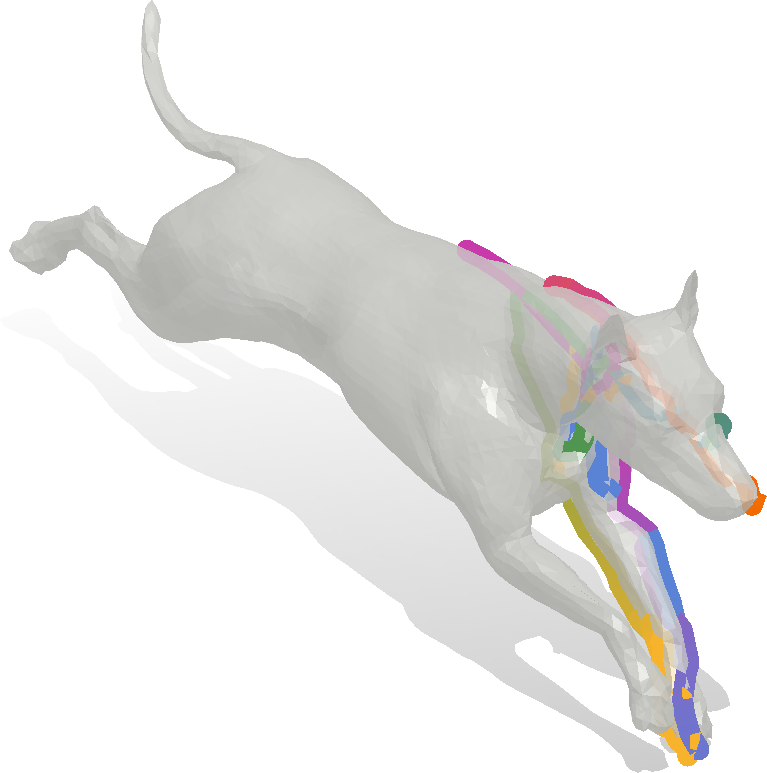} &
    \includegraphics[width=\widthQual,height=\heightQual,keepaspectratio]{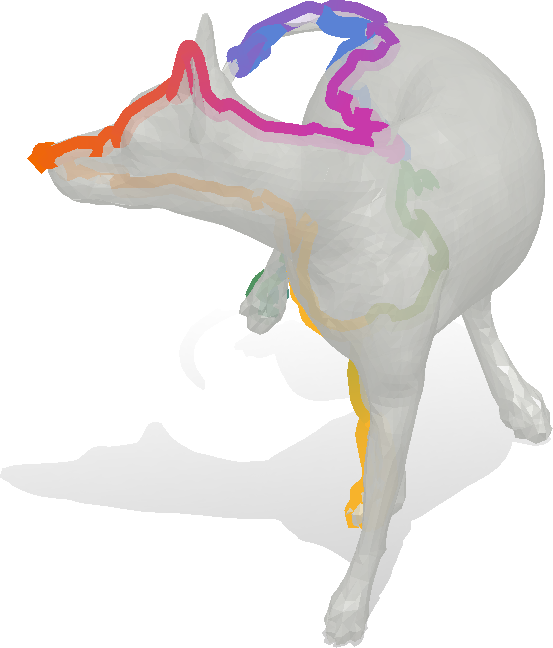} &
    \includegraphics[width=\widthQual,height=\heightQual,keepaspectratio]{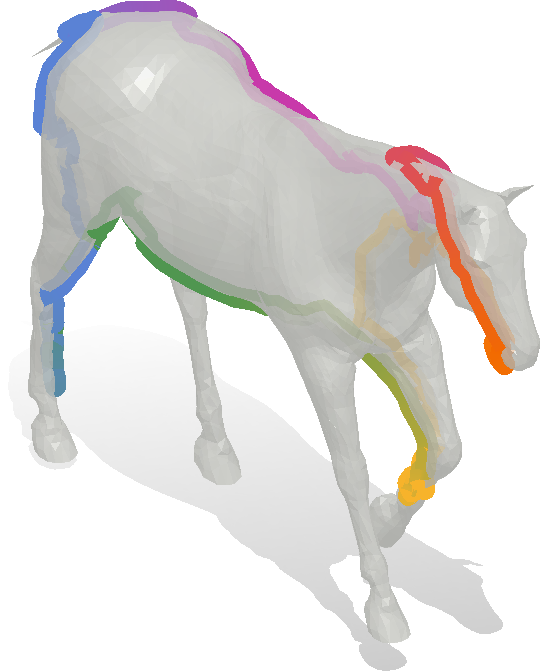} &
    \includegraphics[width=\widthQual,height=\heightQual,keepaspectratio]{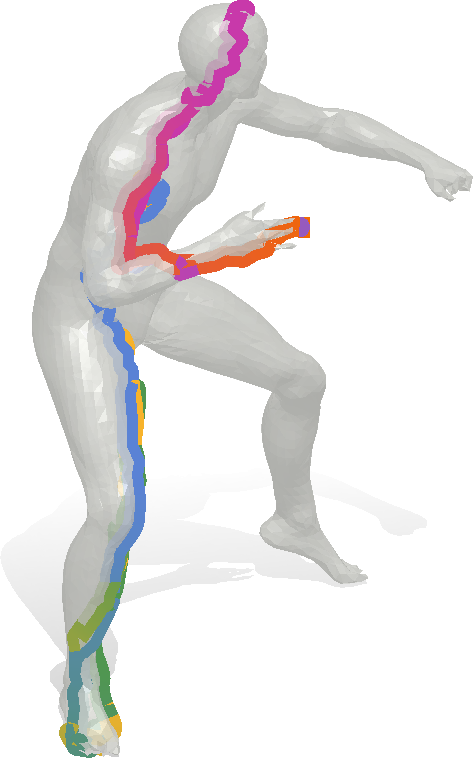} &
    \includegraphics[width=\widthQual,height=\heightQual,keepaspectratio]{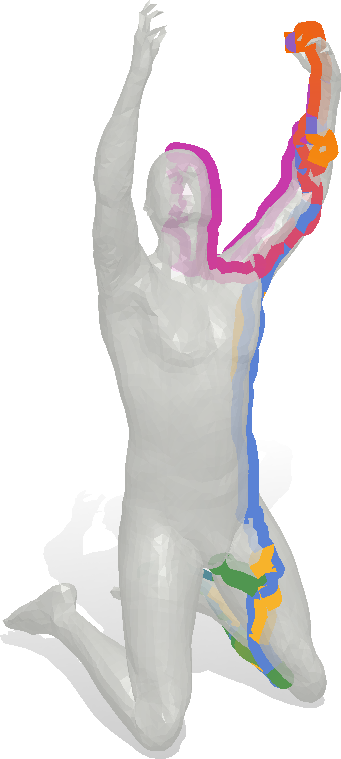}\\
    \rotatebox{90}{\small$\quad$Ours} &
    \includegraphics[width=\widthQual,height=\heightQual,keepaspectratio]{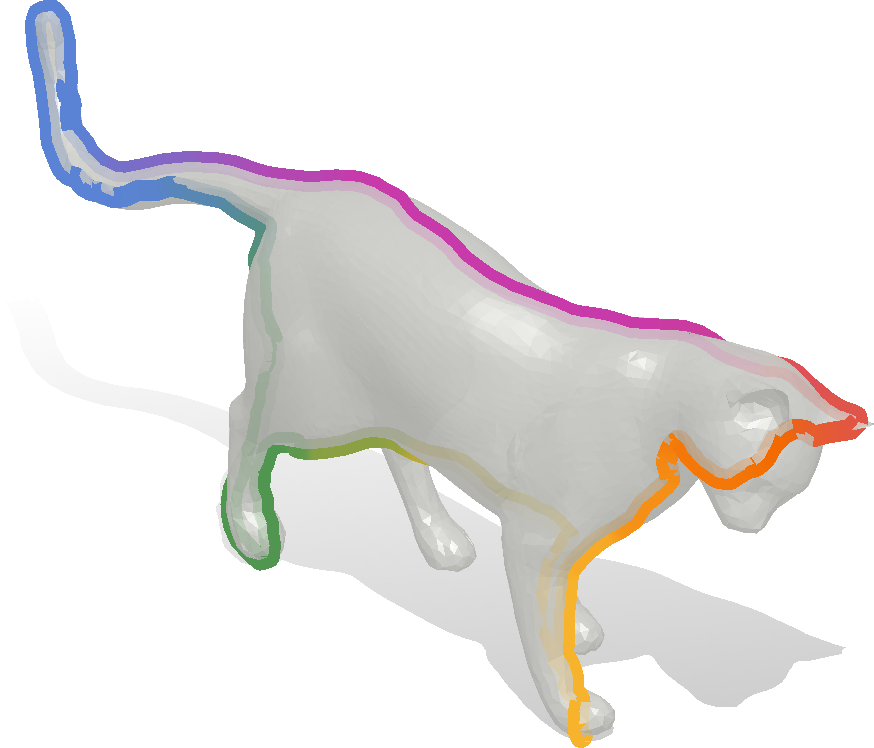} &
    \hspace{-0.1cm}
    \includegraphics[width=\widthQual,height=\heightQual,keepaspectratio]{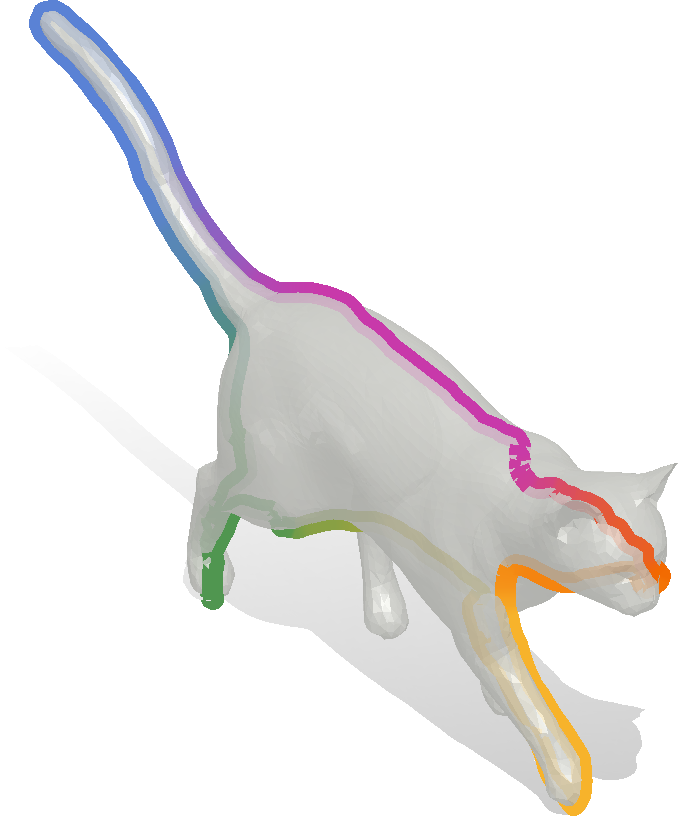} &
    \includegraphics[width=\widthQual,height=\heightQual,keepaspectratio]{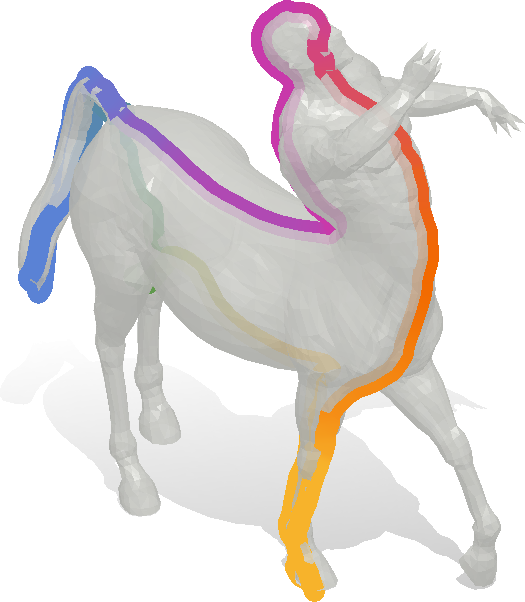} &
    \includegraphics[width=\widthQual,height=\heightQual,keepaspectratio]{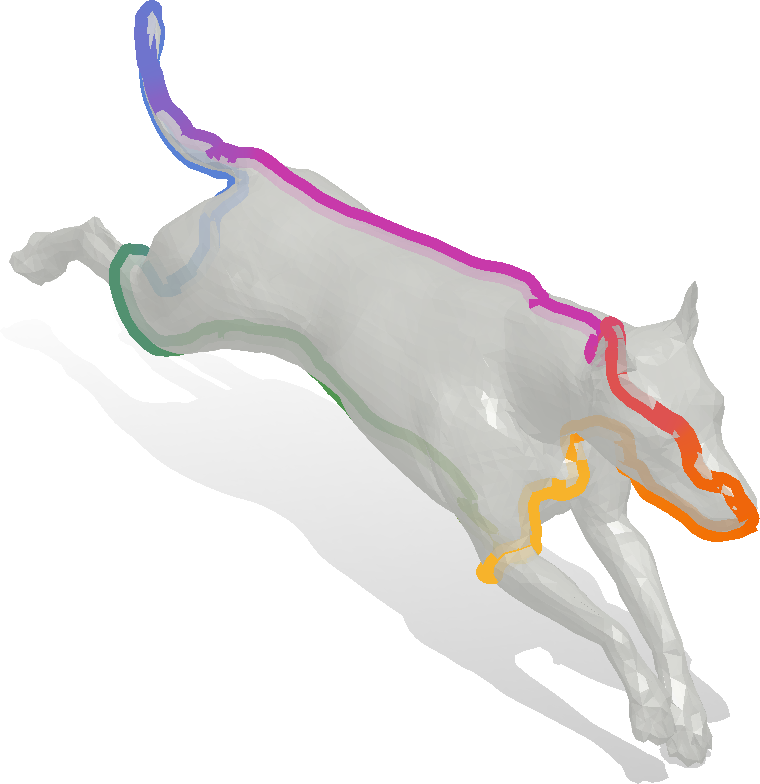} &
    \includegraphics[width=\widthQual,height=\heightQual,keepaspectratio]{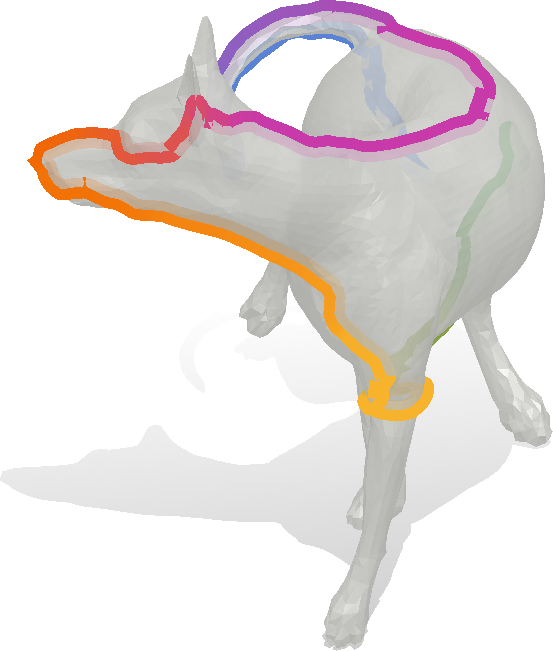} &
    \includegraphics[width=\widthQual,height=\heightQual,keepaspectratio]{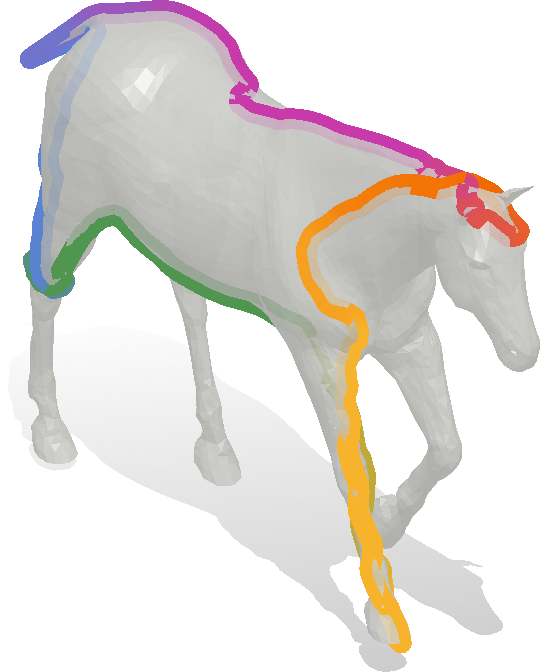} &
    \includegraphics[width=\widthQual,height=\heightQual,keepaspectratio]{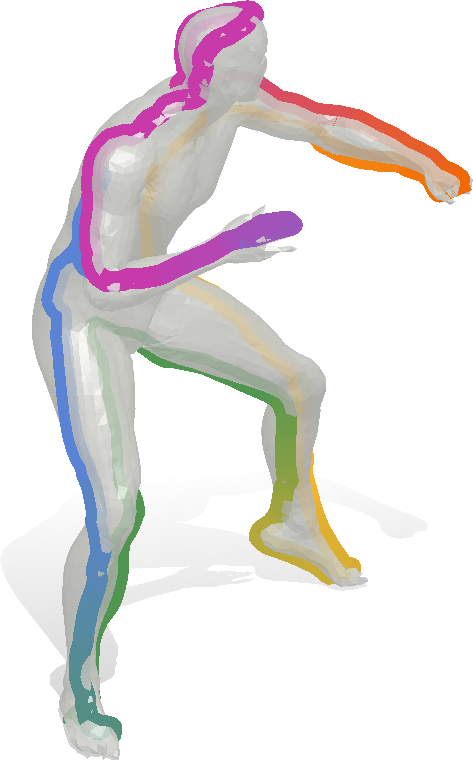} &
    \includegraphics[width=\widthQual,height=\heightQual,keepaspectratio]{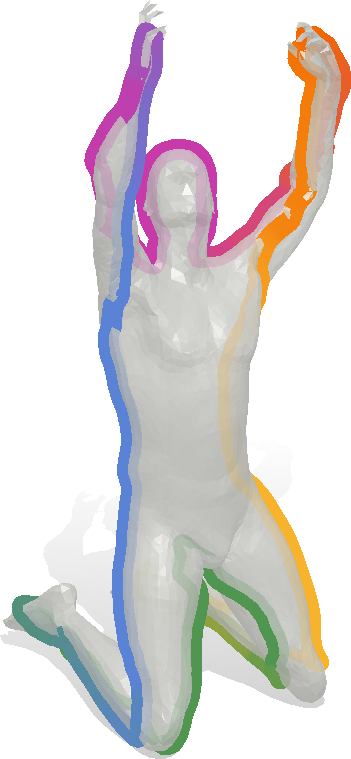}
\end{tabular}
	\vspace{-0.43cm}
	\caption{\textbf{Qualitative comparison} of the method by L\"{a}hner~\etal \cite{lahner2016} (second row) and our approach (third row) on TOSCA. Our approach results in more plausible matchings despite that \laehneretal use a coarse segmentation-based pre-matching. Our local rigidity regulariser, which is enabled by our novel conjugate product graph formalism, ensures that resulting paths on 3D target shapes are much smoother.
    }
	\label{fig:against-sota-qualitative}
\end{figure*}

\subsubsection{Qualitative Matching Results}
We also compare our method qualitatively to \laehneretal~\cite{lahner2016}.
Even though our method is not using segmentation information, matchings computed with our approach are consistently of better quality and reflect a more plausible path on the 3D shape, \ie are locally straight, see \cref{fig:teaser}, \cref{fig:qualitative-faust} and \cref{fig:against-sota-qualitative}.
\subsection{Ablation Studies}
In the Appendix we provide ablation studies considering different parts of our cost function, discretisation, shape discrepancies and noise.

\subsection{Partial Shapes}
\begin{figure}
    \vspace{-0.2cm}
    \hspace{-0.5cm}
    \newcommand{\widthPartial}{2cm}
\newcommand{\widthFirstPartial}{1.1cm}
\newcommand{\heightPartial}{2cm}
\newcommand{\heightPartialQuery}{1.65cm}
\begin{tabular}{ccccccc}
    \rotatebox{90}{\small$\quad$2D Shape}&
    \hspace{-0.3cm}
    \includegraphics[width=\widthFirstPartial,height=\heightPartialQuery,keepaspectratio]{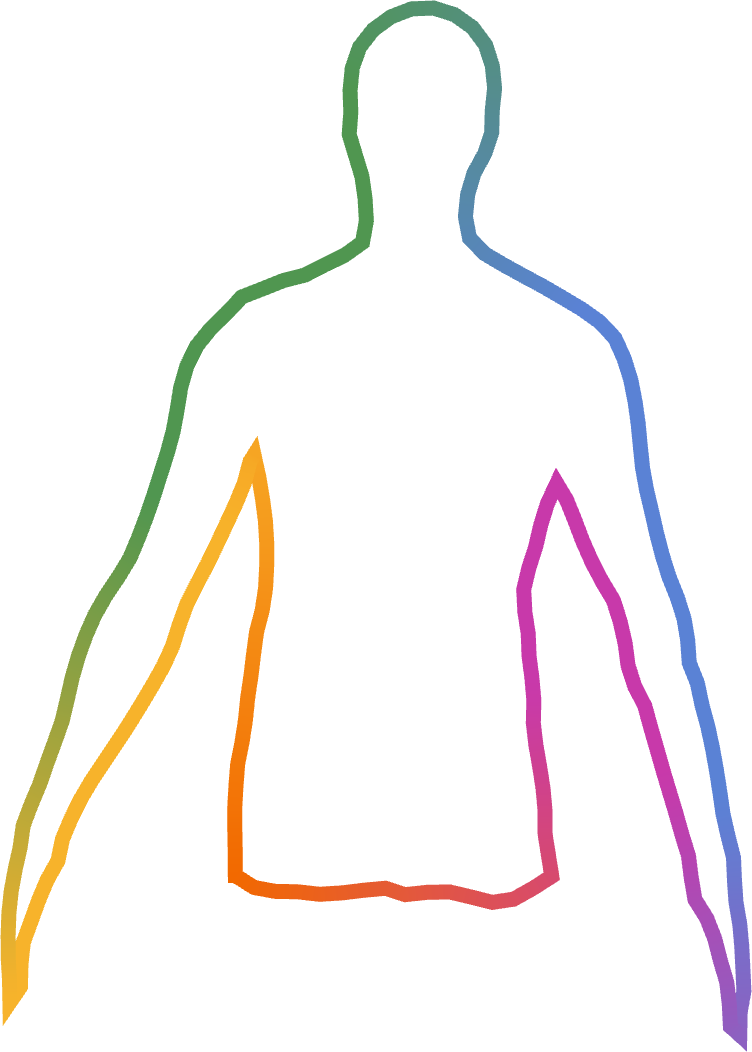} &
    \includegraphics[width=\widthPartial,height=\heightPartialQuery,keepaspectratio]{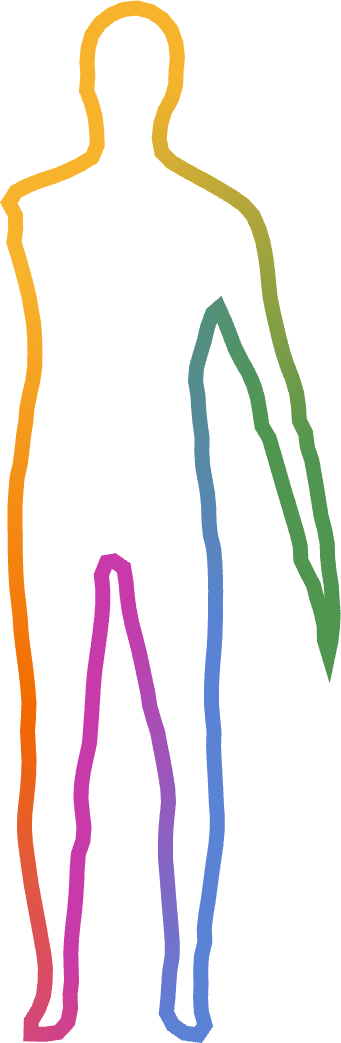} &
    \includegraphics[width=\widthPartial,height=\heightPartialQuery,keepaspectratio]{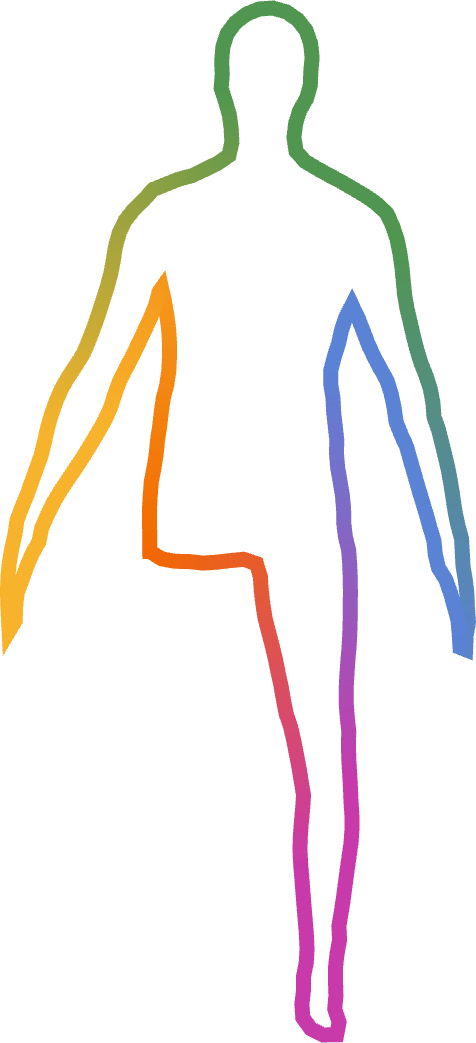} &
    \includegraphics[width=\widthPartial,height=\heightPartialQuery,keepaspectratio]{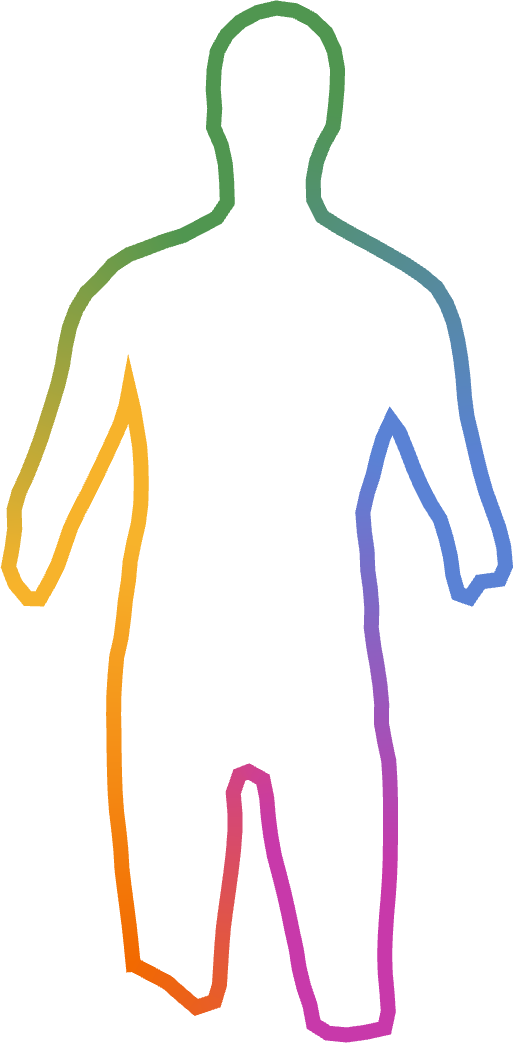} &
    \includegraphics[width=\widthPartial,height=\heightPartialQuery,keepaspectratio]{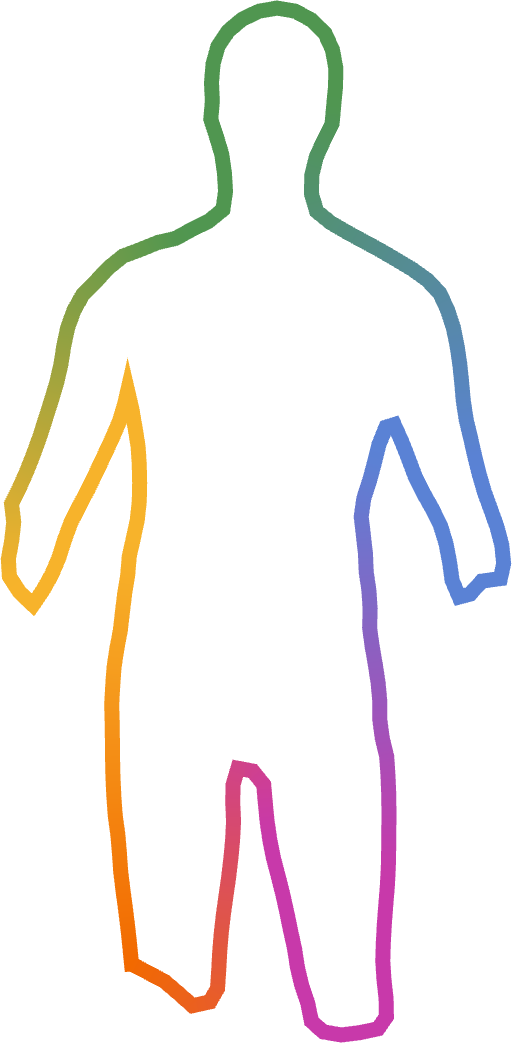} & 
    \includegraphics[width=\widthPartial,height=\heightPartialQuery,keepaspectratio]{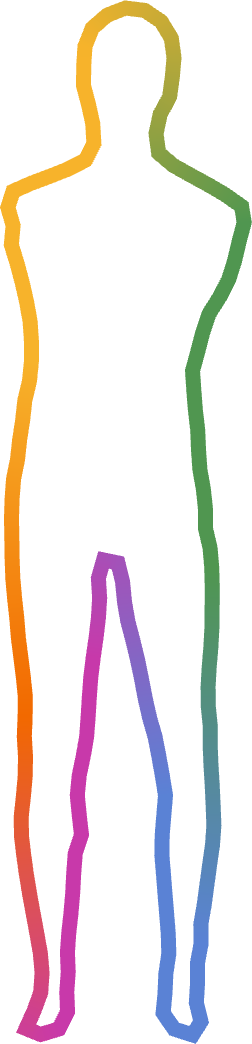} \\
    \rotatebox{90}{\small\laehneretal}&
    \hspace{-0.3cm}
    \includegraphics[width=\widthPartial,height=\heightPartial,keepaspectratio]{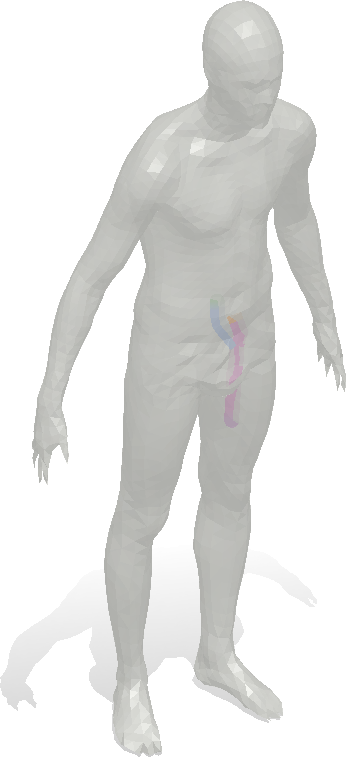} &
    \includegraphics[width=\widthPartial,height=\heightPartial,keepaspectratio]{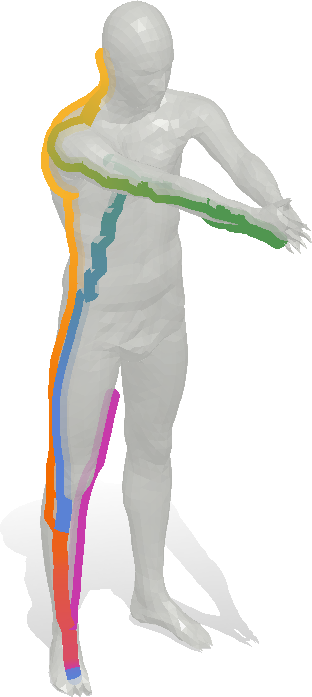} &
    \includegraphics[width=\widthPartial,height=\heightPartial,keepaspectratio]{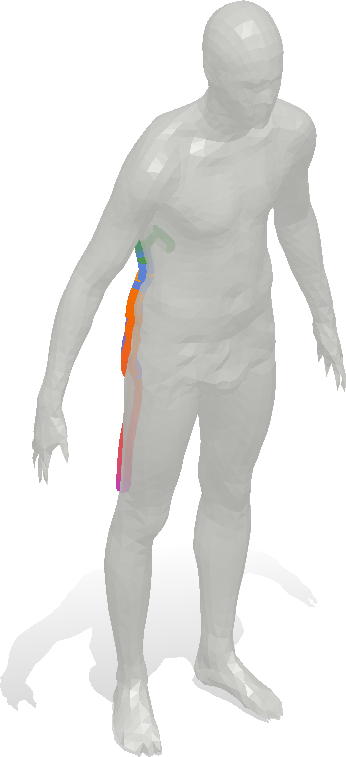} &
    \includegraphics[width=\widthPartial,height=\heightPartial,keepaspectratio]{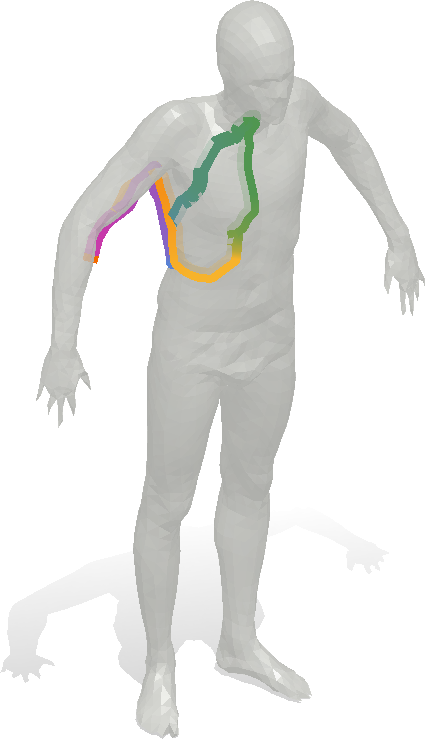} &
    \includegraphics[width=\widthPartial,height=\heightPartial,keepaspectratio]{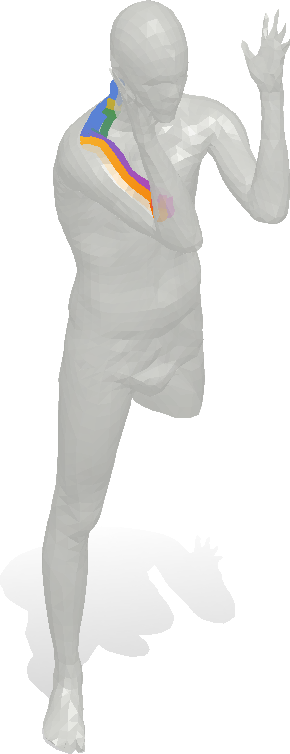} & 
    \includegraphics[width=\widthPartial,height=\heightPartial,keepaspectratio]{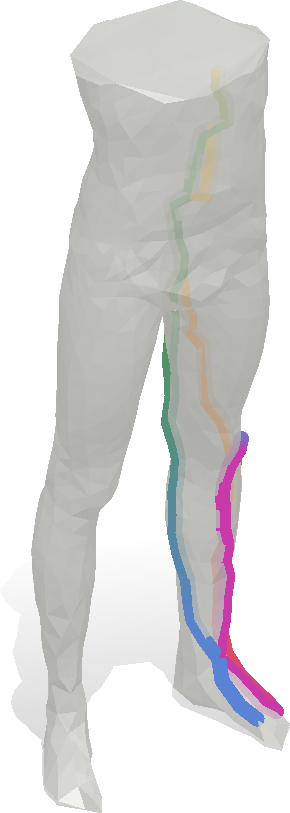} \\
    \rotatebox{90}{\small $\qquad$Ours}&
    \hspace{-0.3cm}
    \includegraphics[width=\widthPartial,height=\heightPartial,keepaspectratio]{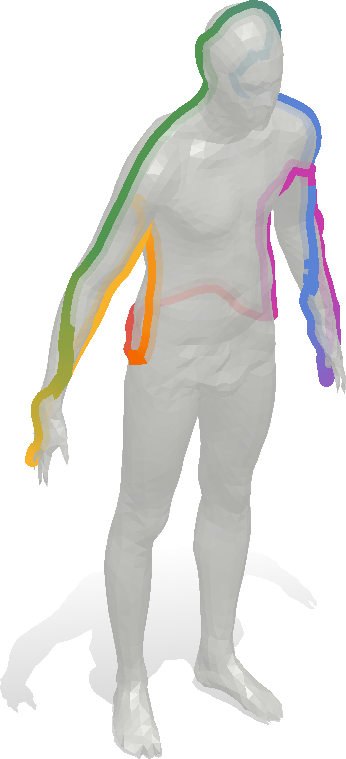} &
    \includegraphics[width=\widthPartial,height=\heightPartial,keepaspectratio]{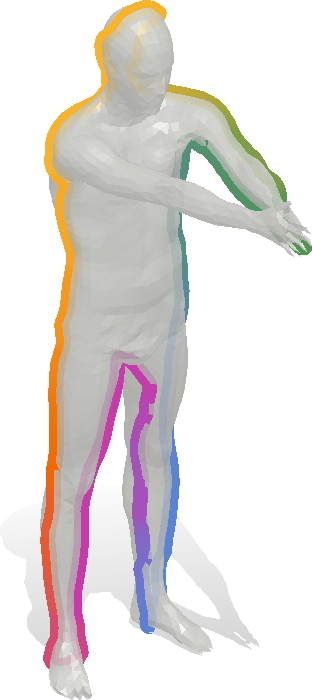} &
    \includegraphics[width=\widthPartial,height=\heightPartial,keepaspectratio]{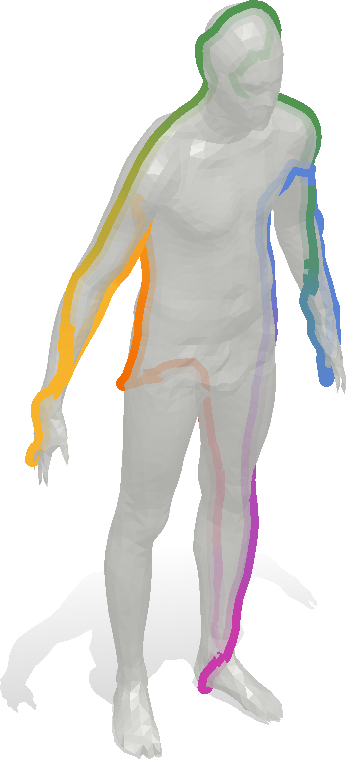} &
    \includegraphics[width=\widthPartial,height=\heightPartial,keepaspectratio]{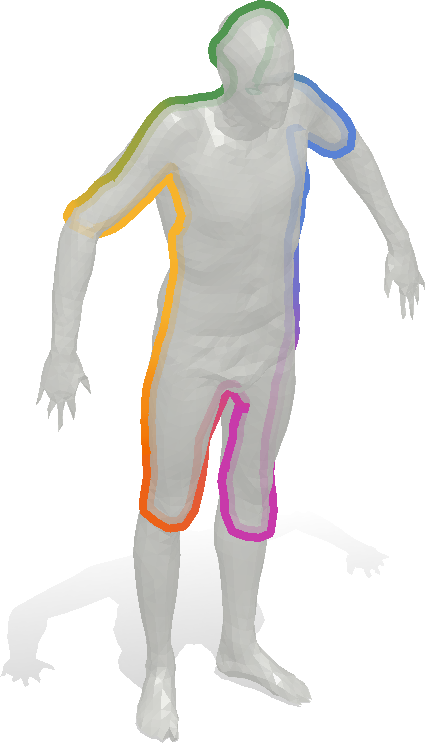} &
    \includegraphics[width=\widthPartial,height=\heightPartial,keepaspectratio]{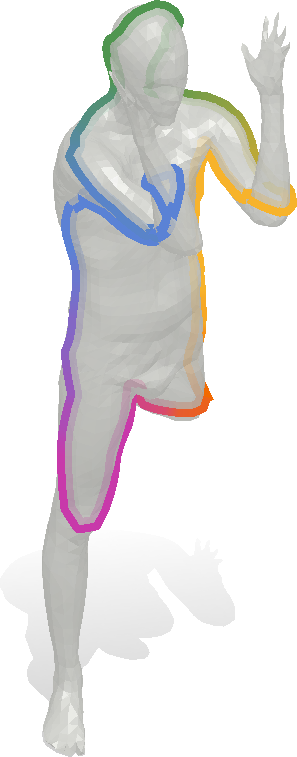} & 
    \includegraphics[width=\widthPartial,height=\heightPartial,keepaspectratio]{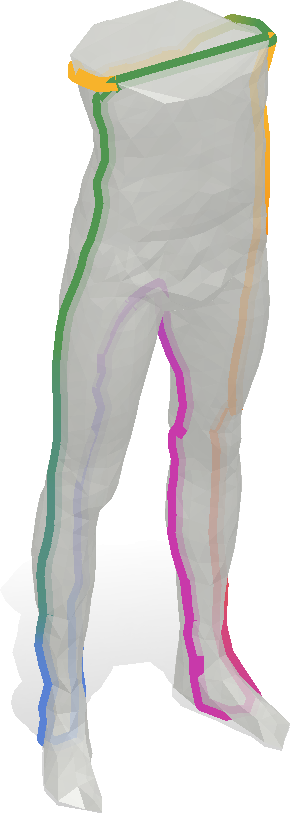} %
\end{tabular}
    \vspace{-0.43cm}
    \caption{\textbf{Qualitative comparison} of \laehneretal \cite{lahner2016} and ours on partial FAUST shapes. The global features of \cite{lahner2016} result in poor matchings in scenarios without full shape, whereas we use local features and thus obtain valid partial matchings.}
    \label{fig:partial-matchings}
    \vspace{-0.25cm}
\end{figure}

We show experiments on partial shapes, for which we removed parts of either the 2D or 3D shape in FAUST, see \cref{fig:partial-matchings}. 
Our approach is substantially more robust in the partial setting compared to \laehneretal~ \cite{lahner2016}, likely due to the locality of our features and strong spatial regularisation, in contrast to the global spectral features of \cite{lahner2016}.

\subsection{Sketch-Based 3D Shape Manipulation}
We show the high quality of our matchings by performimg 2D sketch-based 3D shape manipulation. 
After deforming the contour, the 3D shape is brought into a corresponding pose through as-rigid-as-possible shape deformation \cite{sorkine2007rigid}, see \cref{fig:teaser} (right). 
Details can be found in the Appendix.

\section{Discussion \& Limitations}
Our experimental results confirm that conjugate product graphs enable 2D-3D shape matching without the need of a coarse pre-matching. 
Even though we compute results to global optimality, scenarios like  symmetries (\eg for human shapes) lead to ambiguities that are challenging to reflect in the cost function, which may result in matchings that collapse to one side of the 3D shape, see \cref{fig:qualitative-faust} (bottom-right).
Although our method has the same asymptotic complexity as \cite{lahner2016}, in practice the computation is slower due to the  conjugate product graph being larger (by a constant factor) than the product graph (cf. \cref{tab:graphsizes}, also see Appendix).

\section{Conclusion}
We presented conjugate product graphs for 2D-3D shape matching,
which for the first time allows for the incorporation of higher-order costs within path-based matching formalisms. 
Our novel concept significantly increases model expressiveness and flexibility, allowing to inject desirable properties, like local rigidity regularisation, into respective optimisation problems.
Our results show significant improvements in challenging matching settings, even allowing for 
2D sketch-based 3D shape manipulation.
Since our powerful higher-order regularisation allows to get rid of the need for global features, our method is the first that solves \emph{partial} 2D-3D shape matching. 
We believe that our work is of high relevance to the field of shape analysis, 
and hope to inspire more work on inter-dimensional applications. %

\vspace{0.2cm}
\noindent\textbf{Acknowledgements.}\ PR is funded by the TRA Modellling (University of Bonn) as part of the Excellence Strategy of the federal and state governments.
ZL is funded by a KI-Starter grant of the Ministry of Culture and Science NRW.

{\small
\bibliographystyle{ieee_fullname}
\bibliography{egbib}
}

\clearpage
\appendix
\begin{strip}
	\begin{center}
		{\Large\textbf{Appendix}}
	\end{center}
\end{strip}
\section{Segmentation Pre-Matching}

\begin{figure}[h!]
    \centering
    \begin{tabular}{cc}
        \includegraphics[height=2.5cm, width=4cm, keepaspectratio]{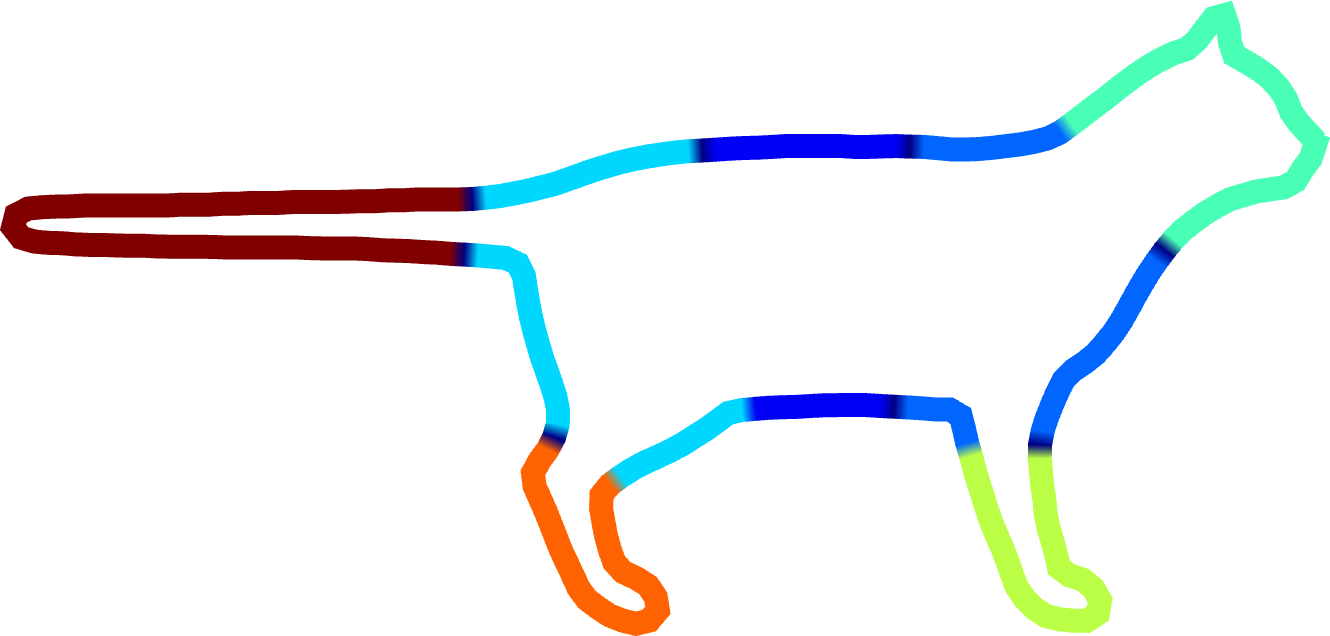} &
        \includegraphics[height = 2.5cm,  width=4cm, keepaspectratio]{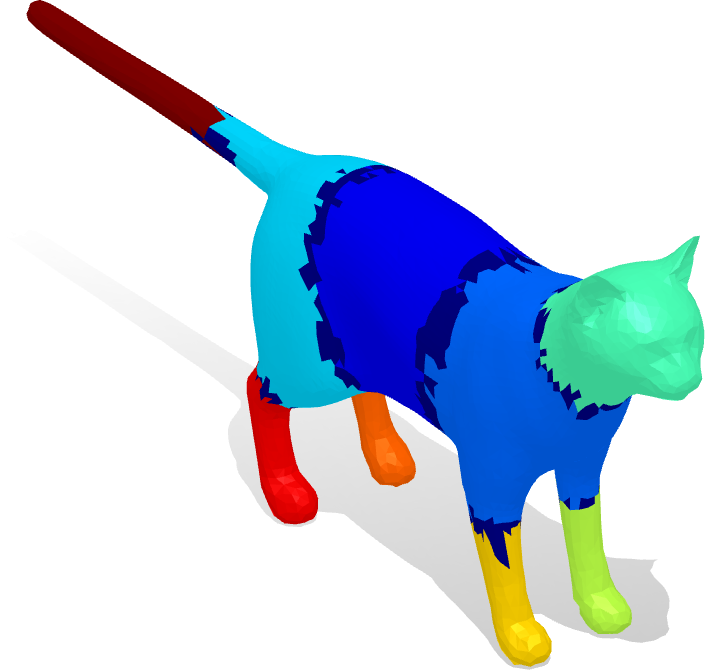}
    \end{tabular}
    \caption{Visualisation of pre-matched segmentation information on cat from the TOSCA dataset. Different colours encode different segments, while darkest blue encodes the transition between different segments.}
    \label{fig:segmentation-visualisation}
\end{figure}

In \cref{fig:segmentation-visualisation}, we visualise the pre-matching which is used by the approach in \cite{lahner2016}.
It is obvious that the injection of such information into the objective function makes finding valid solutions substantially easier.

For a fair comparison, in addition to our method that does \emph{not} use this information, we also evaluate
"Ours-Seg.", in which we incorporate the above segmentation information as an additional feature descriptor, see \cite{lahner2016} for details. 

\section{Branch and Bound Algorithm}\label{sub:bnb}
\cref{alg:b-n-b} describes our optimisation strategy.
We adapt the branch and bound algorithm introduced in \cite{lahner2016} to conjugate product graphs and implement runtime improvements by increasing chances of finding tighter upper bounds earlier.

The final goal of the optimisation is to find a \emph{cyclic} path with minimal cost. However, Dijkstra's algorithm only finds shortest (but not necessarily cyclic) paths. %
To that end, 
we represent  the (conjugate) product graph as sequential graph, in which the first and last layers are duplicates, such that a path going from the same vertex in the first and last layer corresponds to a cyclic path.

Thus, the cyclic path with minimal cost can be found by computing the shortest path for every vertex on the first layer to every respective vertex on the last layer, and subsequently choosing among the computed paths the one with minimal cost.
In general, this requires to solve a total of $2|\meshEdgs| + |\meshVerts|$ (ordinary) shortest path problems, and is computationally more expensive than the  branch-and-bound strategy that we pursue.

\begin{figure}[h!]
    \includegraphics[width=\columnwidth]{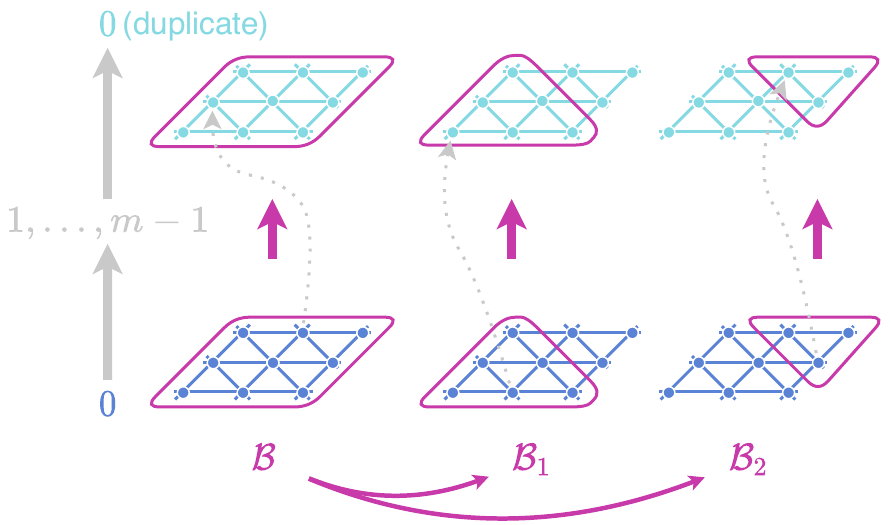}
    \caption{Illustration of the \textbf{branching strategy} in \cref{alg:b-n-b}. First, the shortest path from all vertices in $\branchSet$ on the first layer to the same vertices on the duplicate first layer (which amounts to the last layer) is computed. The resulting shortest path from $\branchSet$ on the first layer to $\branchSet$ on the last layer might not start and end on the same vertex (since we are searching for a shortest path from a set of vertices to a set of vertices). Whenever this is the case, $\branchSet$ is partitioned into two sets $\branchSet_1$ and $\branchSet_2$, for which in subsequent iterations shortest paths are computed analogously as for $\branchSet$.
    }
    \label{fig:layers-b-n-b}
\end{figure}

The main idea of branch-and-bound is to iteratively subdivide the search space, while tightening upper and lower bounds using the results of previous iterations.
In that sense, instead of searching for shortest paths from each vertex on the first layer to each respective vertex on the last layer, we search for the shortest path from a set of vertices $\branchSet\subset\conjProdVerts$ on the first layer to the respective set of vertices $\branchSet$ on the last layer, see \cref{fig:layers-b-n-b} (left).
There is no guarantee that the path $\mathcal{C} = (v_1^*, \dots, v_{|\mathcal{C}|}^*)$ from $\branchSet$ (first layer) to its duplicate $\branchSet$ (last layer) with minimal energy is indeed cyclic, \ie that the final vertex $v_{|\mathcal{C}|}^*$ in the last layer is indeed the same as the starting vertex $v_1^*$ in the first layer.
If $\mathcal{C}$ is not cyclic, we partition $\branchSet$ into smaller, disjunct subsets $\branchSet_1$ and $\branchSet_2$ (with $\branchSet_1 \cup \branchSet_2 = \branchSet$ and $\branchSet_1 \cap \branchSet_2 = \emptyset$) until a cyclic path is found (this is the \emph{branching strategy}, see \cref{fig:layers-b-n-b}).
The partitioning is done by calculating Voronoi cells around edges $e_1^\mesh$ and $e_{|\mathcal{C}|}^\mesh$ on 3D shape assuming $e_1^\mesh$ and $e_{|\mathcal{C}|}^\mesh$ are not identical (where the conjugate product vertex $v_1^* = (e_1^\contour,e_1^\mesh)$ contains edge $e_1^\mesh$ on 3D shape and conjugate product vertex $v_{|\mathcal{C}|}^* = (e_{|\mathcal{C}|}^\contour,e_{|\mathcal{C}|}^\mesh)$ contains edge $e_{|\mathcal{C}|}^\mesh$ on 3D shape).  %
If $e_1^\mesh$ and $e_{|\mathcal{C}|}^\mesh$ are identical we partition according to $\branchSet_1 = \branchSet\setminus\{v_{|\mathcal{C}|}^*\}$ and $\branchSet_2 = \{v_{|\mathcal{C}|}^*\}$.%

The path cost $d_\mathcal{C}$ of non-cyclic paths (\ie $v_{1}^* \neq v_{|\mathcal{C}|}^*$) is a lower bound $b(\cdot)$ on the path cost of the globally optimal cyclic path.
Whenever $v_1^*$ and $v_{|\mathcal{C}|}^*$ are equal %
(meaning that $\mathcal{C}$ is a cyclic path), an upper bound $b_\text{upper}$ is found, which might already be the globally optimal path, but can only be identified as such if all other branches do not yield cyclic paths with lower costs.
Hence, the algorithm has to explore all other branches, which in the worst case are as many as there are vertices on one layer (\ie $2|\meshEdgs| + |\meshVerts|$ many).

While searching for the optimal path, the algorithm only explores paths with cost $d_{\mathcal{C}} < b_\text{upper}$ and thus performance can be improved if tighter upper bounds $b_\text{upper}$ are found as early as possible.
We improve the branch-and-bound algorithm of \cite{lahner2016} by computing all paths $\mathcal{C}_\text{all}$ of a branch, and then search within these for cyclic paths to find lower values of the upper bound $b_\text{upper}$ earlier.
We want to point out that no additional computational effort is required to compute $\mathcal{C}_\text{all}$ using the implementation of~\cite{lahner2016}, since all paths are already available (see \cref{fig:runtime-comparison} for runtime comparisons). 
\begin{algorithm}[htp!]
\newcommand\mycommfont[1]{\footnotesize\ttfamily\textcolor{cGRAY}{#1}}
\SetCommentSty{mycommfont}
\SetKwInOut{Input}{Input}\SetKwInOut{Output}{Output}
\SetKwComment{Comment}{// }{}
\Input{2D shape $\contour = (\contourVerts, \contourEdgs)$, \\ 3D shape $\mesh = (\meshVerts, \meshEdgs)$}
\Output{Optimal Path $\mathcal{C}_\text{opt}\subset\conjProdVerts$}
\BlankLine
\Comment{First branch is complete first layer}
$\branchSet_0\gets \{v^*=(e^\contour,e^\mesh) \;|\; i_0 = 0, e^\contour=(i_0, i_1)\}$\;
\Comment{Initialise bounds and branches}
$b(\branchSet_0) \gets 0$\;
$b_\text{upper} \gets \infty$\;
$\branchSet_\text{Branches} \gets \branchSet_0$\;
\Comment{Run until no branches with a gap between lower and upper bound exist}
\While{$\underset{\branchSet \in \branchSet_\text{Branches}}{\min}b(\branchSet) < b_\text{upper}$}{
    $\branchSet \gets \underset{\branchSet \in \branchSet_\text{Branches}}{\text{argmin }}b(\branchSet)$\;
    $\branchSet_\text{Branches} \gets \branchSet_\text{Branches} \setminus \branchSet$\;
    Compute all paths $\mathcal{C}_\text{all}=\{\mathcal{C}_1,\mathcal{C}_2,\dots\}$ with path cost $d_{\mathcal{C}_i}<b_\text{upper}$ starting and ending in $\branchSet$\;
    \If{$\mathcal{C}_\text{all} = \emptyset$}{
        \Comment{No path which meets $d_{\mathcal{C}}<b_\text{upper}$}
        continue\;
    }
    $\mathcal{C} \gets \underset{\mathcal{C} \in \mathcal{C}_\text{all}}{\text{argmin }} \; d_\mathcal{C}$\;
    \Comment{Check if current path is cyclic}
    \eIf{$v_1^* = v_{|\mathcal{C}|}^*$}{
        \If{$d_\mathcal{C} < b_\text{upper}$}{
            $b_\text{upper} \gets d_\mathcal{C}$\;
            $\mathcal{C}_\text{opt} \gets \mathcal{C}$\;
        }

    }{
        \Comment{Cut current branch into two parts}
        \eIf{$e_1^\mesh = e_{|\mathcal{C}|}^\mesh$}{
            $\branchSet_1 \gets \branchSet \setminus \{v_{|\mathcal{C}|}^*\}$\;
            $\branchSet_2 \gets \{v_{|\mathcal{C}|}^*\}$\;
        } {
            Compute $\branchSet_1, \branchSet_2$ as Voronoi cells around $e_1^\mesh$ and $e_{|\mathcal{C}|}^\mesh$ respectively\;
        }
        \Comment{Add new branches}
        $\branchSet_\text{Branches} \gets \branchSet_\text{Branches} \cup \{\branchSet_1, \branchSet_2$\}\;
        \Comment{Update lower bounds}
        $b(\branchSet_1)=b(\branchSet_2) = d_\mathcal{C}$\;
        \Comment{Try to tighten upper bound}
        \For{$\mathcal{C} \gets \mathcal{C}_\text{all}$}{
            \If{$v_1^* = v_{|\mathcal{C}|}^*$}{
                \If{$d_\mathcal{C} < b_\text{upper}$}{
                    $b_\text{upper} \gets d_\mathcal{C}$\;
                    $\mathcal{C}_\text{opt} \gets \mathcal{C}$\;
                }
            }
        }
    }
}
\caption{\small Branch and Bound for \emph{Cyclic} Shortest Path on (Conjugate) Product Graph\vspace{-0.5cm}}\label{alg:b-n-b}
\end{algorithm}
\section{Number of Conjugate Product Edges}\label{sub:number_cpe}
\begin{figure}[h!]
    \centering
    \includegraphics[height=2.5cm]{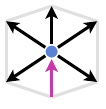}
    \caption{Subset of a triangle mesh. Directed pink edge is connected to all directed black edges via blue vertex.}
    \label{fig:edge-count}
\end{figure}

As mentioned in the main paper, the conjugate product graph $\conjProdGraph$ has $7$ times more vertices than $\prodGraph$ and $c \approx 11$ times more edges.
In the following we derive $c$.
To this end, we count outgoing edges of each conjugate product vertex (which is sufficient since $\conjProdGraph$ is cyclic).
Further, we assume that on average each vertex $j$ of the 3D shape $\mesh$ is connected to $6$ edges~\cite{botsch2010}.
Thus, each (directed) edge on 3D shape is connected to $5$ other directed edges via their shared vertex, see \cref{fig:edge-count}.

In conclusion, each conjugate product vertex is connected to $5$ conjugate product vertices on the same layer (reflecting  \textit{degenerate 2D} conjugate product vertices) and $6$ conjugate product vertices on the next layer (reflecting $5$ \textit{non-degenerate} conjugate product vertices and $1$ \textit{degenerate 3D} conjugate product vertex).
In total, each conjugate product vertex is connected to $c \approx 11$ other conjugate product vertices.
In combination with the number of vertices of the conjugate product graph $|\conjProdVerts|=|\contourVerts| \cdot \big(2|\meshEdgs| + |\meshVerts|\big)$ we obtain the number of edges of $\conjProdGraph$.

\section{Runtime}
\subsection{Runtime Analysis}
In the following we estimate runtime complexity of our branch-and-bound algorithm for conjugate product graphs.
To this end, we use $|\meshEdgs| \approx 3|\meshVerts|$~\cite{botsch2010} to obtain $|\conjProdVerts| \approx 7 \cdot |\contourVerts||\meshVerts|$ and $|\conjProdEdgs| \approx c\cdot 7\cdot |\contourVerts||\meshVerts|$.

The runtime of Dijkstra on an arbitrary graph $\graph = (\graphVerts,\graphEdgs)$ is $\mathcal{O}\big((\vert \graphEdgs \vert + \vert \graphVerts \vert) \cdot \log(\vert \graphVerts \vert)\big)$ where $(\vert \graphEdgs \vert + \vert \graphVerts \vert)$ indicates the number of update steps to be made, and $\log(\vert \graphVerts \vert)$ indicates the complexity to access the priority heap that is used to keep track of the next nodes to be updated. 

In our case, the number of update steps is $(|\conjProdEdgs| %
+ |\conjProdVerts|) \approx c\cdot 14\cdot |\contourVerts||\meshVerts|$ (with $c \approx 11$). We make use of the strictly directed order of the %
$|\contourVerts|$ layers of $\conjProdGraph$, which allows to use a heap that scales with the number of vertices of one layer
$\mathcal{O}(\vert \meshVerts \vert)$ (also see \cite{lahner2016}). %
In summary, the runtime complexity of a single Dijkstra run in our conjugate product graph $\conjProdGraph$ is  $\mathcal{O}(|\contourVerts||\meshVerts|\log(|\meshVerts|))$.

To find the optimal {cyclic} path among all possible cyclic paths, we run Dijkstra not just once but at most $\mathcal{O}(\vert \meshVerts \vert)$ times (without any branch-and-bound optimisation), which leads to a final runtime complexity of $\mathcal{O}(|\contourVerts||\meshVerts|^2\log(|\meshVerts|))$.

\subsection{Runtime Comparison}
\begin{figure}[h!]
    \centering
    \begin{tabular}{cc}
    \setlength{\tabcolsep}{0pt}
    \hspace{-0.8cm}
    \input{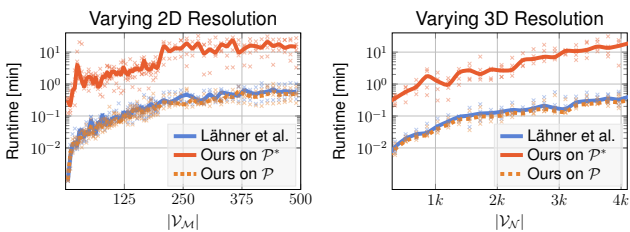} &
    \hspace{-7mm}
    \newcommand{\runtimeLineWidth}{3pt}
\newcommand{\rtplotWidth}{0.75\columnwidth}
\newcommand{\rtplotHeight}{0.5\columnwidth}
\pgfplotsset{%
    label style = {font=\sffamily\large},
    tick label style = {font=\sffamily\large},
    title style =  {font=\Large\sffamily},
    legend style={  fill= gray!10,
                    fill opacity=0.6, 
                    font=\sffamily\large,
                    draw=gray!20, %
                    text opacity=1}
}
\begin{tikzpicture}[scale=0.5, transform shape]
    \begin{axis}[%
        width=\rtplotWidth,
        height=\rtplotHeight,
        title=Varying 3D Resolution,
		title style={yshift=-0.25cm},
        scale only axis,
        grid=major,
        legend style={
			at={(0.98,0.02)},
			anchor=south east,
			legend columns=1,
			legend cell align={left}},
		ylabel={{\sffamily\large Runtime [min]}},
		xlabel={$|\meshVerts|$},
        xmin=300,
        xmax=4100,
        xtick={1000, 2000, 3000, 4000},
        xticklabels={{$1k$}, $2k$, {$3k$}, {$4k$}},
        xtick scale label code/.code={},
        ymode=log,
        ymin=0.0005,
        ymax=40,
        ytick={0.01, 0.1, 1, 10},
        ylabel near ticks,
    ]
\addplot [color=mycolor1, only marks, mark=x, mark options={solid, opacity=0.4}, forget plot]
table[row sep=crcr]{%
174 0.0028767\\
346 0.0069165\\
518 0.018448\\
690 0.037166\\
863 0.057566\\
1035 0.08347\\
1207 0.10621\\
1379 0.094414\\
1551 0.18143\\
1724 0.2335\\
1896 0.15578\\
2068 0.15103\\
2240 0.15379\\
2412 0.15892\\
2585 0.1519\\
2757 0.17747\\
2929 0.22164\\
3101 0.20605\\
3273 0.24569\\
3446 0.29418\\
3618 0.30093\\
3790 0.35175\\
3962 0.26997\\
4134 0.3798\\
174 0.0060621\\
346 0.01575\\
518 0.019259\\
690 0.024675\\
863 0.028345\\
1035 0.041232\\
1207 0.064454\\
1379 0.068214\\
1551 0.045593\\
1724 0.070524\\
1896 0.089297\\
2068 0.081678\\
2240 0.11508\\
2412 0.13507\\
2585 0.14588\\
2757 0.17078\\
2929 0.14971\\
3101 0.15929\\
3273 0.26156\\
3446 0.27385\\
3618 0.27859\\
3790 0.30473\\
3962 0.21414\\
4134 0.22477\\
174 0.0040459\\
346 0.010687\\
518 0.02277\\
690 0.020692\\
863 0.029252\\
1035 0.049513\\
1207 0.067048\\
1379 0.08216\\
1551 0.08482\\
1724 0.096931\\
1896 0.12288\\
2068 0.13555\\
2240 0.14787\\
2412 0.1498\\
2585 0.21739\\
2757 0.20723\\
2929 0.17647\\
3101 0.15682\\
3273 0.20914\\
3446 0.27997\\
3618 0.30753\\
3790 0.25707\\
3962 0.32896\\
4134 0.4046\\
174 0.0047516\\
346 0.011319\\
518 0.019397\\
690 0.028156\\
863 0.030039\\
1035 0.035678\\
1207 0.050938\\
1379 0.093455\\
1551 0.10329\\
1724 0.13669\\
1896 0.11422\\
2068 0.18317\\
2240 0.15491\\
2412 0.18703\\
2585 0.17357\\
2757 0.25672\\
2929 0.17768\\
3101 0.16585\\
3273 0.31154\\
3446 0.39705\\
3618 0.38612\\
3790 0.44547\\
3962 0.57886\\
4134 0.53122\\
174 0.0034394\\
346 0.00795\\
518 0.027028\\
690 0.026602\\
863 0.037868\\
1035 0.047602\\
1207 0.07475\\
1379 0.097914\\
1551 0.11572\\
1724 0.11943\\
1896 0.12831\\
2068 0.11855\\
2240 0.26275\\
2412 0.14946\\
2585 0.3105\\
2757 0.39225\\
2929 0.55857\\
3101 0.39911\\
3273 0.40766\\
3446 0.54365\\
3618 0.467\\
3790 0.48076\\
3962 0.50714\\
4134 0.54755\\
};

\addplot [color=mycolor1, smooth, line width=\runtimeLineWidth]table[row sep=crcr]{%
174 0.0040459\\
346 0.010687\\
518 0.019397\\
690 0.026602\\
863 0.030039\\
1035 0.047602\\
1207 0.067048\\
1379 0.093455\\
1551 0.10329\\
1724 0.11943\\
1896 0.12288\\
2068 0.13555\\
2240 0.15379\\
2412 0.1498\\
2585 0.17357\\
2757 0.20723\\
2929 0.17768\\
3101 0.16585\\
3273 0.26156\\
3446 0.29418\\
3618 0.30753\\
3790 0.35175\\
3962 0.32896\\
4134 0.4046\\
};
\addlegendentry{\textcolor{black}{L{\"a}hner et al.}}

\addplot [color=mycolor3, only marks, mark=x, mark options={solid, opacity=0.4}, forget plot]
table[row sep=crcr]{%
174 0.23176\\
346 0.45094\\
518 0.56891\\
690 0.82198\\
863 1.7511\\
1035 0.6915\\
1207 0.96376\\
1379 2.2675\\
1551 3.2544\\
1724 2.0655\\
1896 2.8503\\
2068 1.683\\
2240 2.8347\\
2412 3.7993\\
2585 4.4334\\
2757 5.2018\\
2929 6.6289\\
3101 8.8853\\
3273 8.2019\\
3446 10.1242\\
3618 10.5132\\
3790 9.1219\\
3962 3.7856\\
4134 3.9494\\
174 0.2427\\
346 0.35554\\
518 0.81266\\
690 0.68809\\
863 2.5401\\
1035 2.2884\\
1207 0.7606\\
1379 2.9719\\
1551 4.3415\\
1724 7.2233\\
1896 2.8195\\
2068 3.9589\\
2240 6.6666\\
2412 5.6298\\
2585 6.1086\\
2757 7.1381\\
2929 12.9875\\
3101 11.6769\\
3273 10.732\\
3446 11.7421\\
3618 13.1117\\
3790 14.8255\\
3962 15.6645\\
4134 21.1874\\
174 0.52393\\
346 0.24356\\
518 0.77917\\
690 0.89874\\
863 0.78252\\
1035 1.2216\\
1207 0.31907\\
1379 2.3694\\
1551 2.6579\\
1724 2.779\\
1896 3.0452\\
2068 3.4413\\
2240 3.7782\\
2412 3.6446\\
2585 12.4119\\
2757 16.9985\\
2929 10.4789\\
3101 11.0738\\
3273 16.6095\\
3446 19.6165\\
3618 20.9\\
3790 31.1146\\
3962 23.7227\\
4134 29.5899\\
174 0.17124\\
346 0.43598\\
518 0.51229\\
690 1.3528\\
863 2.8943\\
1035 1.4321\\
1207 1.2636\\
1379 0.94287\\
1551 1.4735\\
1724 1.6042\\
1896 1.508\\
2068 2.6519\\
2240 1.8148\\
2412 9.1927\\
2585 7.4369\\
2757 6.7565\\
2929 7.0354\\
3101 8.3192\\
3273 11.958\\
3446 21.1366\\
3618 14.8748\\
3790 7.2808\\
3962 20.8039\\
4134 19.1324\\
174 0.17961\\
346 0.2127\\
518 0.26756\\
690 0.48036\\
863 0.44932\\
1035 0.30596\\
1207 1.5371\\
1379 1.4649\\
1551 1.3565\\
1724 1.3773\\
1896 1.3564\\
2068 1.6616\\
2240 2.5234\\
2412 5.5799\\
2585 5.8982\\
2757 10.9639\\
2929 5.9062\\
3101 10.8197\\
3273 5.6515\\
3446 6.4948\\
3618 24.7967\\
3790 13.5975\\
};
\addplot [color=mycolor3, smooth, line width=\runtimeLineWidth]
table[row sep=crcr]{%
174 0.23176\\
346 0.35554\\
518 0.56891\\
690 0.82198\\
863 1.7511\\
1035 1.2216\\
1207 0.96376\\
1379 2.2675\\
1551 2.6579\\
1724 2.0655\\
1896 2.8195\\
2068 2.6519\\
2240 2.8347\\
2412 5.5799\\
2585 6.1086\\
2757 7.1381\\
2929 7.0354\\
3101 10.8197\\
3273 10.732\\
3446 11.7421\\
3618 14.8748\\
3790 13.5975\\
3962 15.6645\\
4134 19.1324\\
};
\addlegendentry{\textcolor{black}{Ours on $\mathcal{P}^*$}}

\addplot [color=mycolor2, only marks, mark=x, mark options={solid, opacity=0.4}, forget plot]
table[row sep=crcr]{%
174 0.0025221\\
346 0.0059873\\
518 0.015841\\
690 0.030828\\
863 0.047089\\
1035 0.066997\\
1207 0.08662\\
1379 0.077252\\
1551 0.1469\\
1724 0.1896\\
1896 0.11762\\
2068 0.12092\\
2240 0.12524\\
2412 0.12849\\
2585 0.11724\\
2757 0.14085\\
2929 0.182\\
3101 0.16342\\
3273 0.19423\\
3446 0.23604\\
3618 0.24087\\
3790 0.28279\\
3962 0.21011\\
4134 0.31181\\
174 0.0052422\\
346 0.013845\\
518 0.016047\\
690 0.020913\\
863 0.023446\\
1035 0.035615\\
1207 0.053274\\
1379 0.055126\\
1551 0.03606\\
1724 0.056966\\
1896 0.067402\\
2068 0.066252\\
2240 0.092318\\
2412 0.11142\\
2585 0.12325\\
2757 0.14138\\
2929 0.12691\\
3101 0.12731\\
3273 0.21622\\
3446 0.22214\\
3618 0.23423\\
3790 0.25307\\
3962 0.17441\\
4134 0.1857\\
174 0.0036565\\
346 0.0090635\\
518 0.019026\\
690 0.017737\\
863 0.025366\\
1035 0.041088\\
1207 0.05388\\
1379 0.065437\\
1551 0.067375\\
1724 0.07565\\
1896 0.096199\\
2068 0.10386\\
2240 0.11524\\
2412 0.11862\\
2585 0.17237\\
2757 0.16512\\
2929 0.14045\\
3101 0.12578\\
3273 0.17106\\
3446 0.22575\\
3618 0.25259\\
3790 0.20643\\
3962 0.2705\\
4134 0.32465\\
174 0.0041821\\
346 0.0095179\\
518 0.01634\\
690 0.024246\\
863 0.025691\\
1035 0.029772\\
1207 0.043558\\
1379 0.07864\\
1551 0.082386\\
1724 0.10992\\
1896 0.088066\\
2068 0.14221\\
2240 0.12764\\
2412 0.14765\\
2585 0.15451\\
2757 0.20708\\
2929 0.14519\\
3101 0.13423\\
3273 0.24415\\
3446 0.31644\\
3618 0.30739\\
3790 0.35459\\
3962 0.473\\
4134 0.42712\\
174 0.0032358\\
346 0.00693\\
518 0.022741\\
690 0.022509\\
863 0.03194\\
1035 0.039921\\
1207 0.061374\\
1379 0.079062\\
1551 0.093368\\
1724 0.096522\\
1896 0.10059\\
2068 0.094725\\
2240 0.2127\\
2412 0.12229\\
2585 0.25343\\
2757 0.30848\\
2929 0.44817\\
3101 0.33524\\
3273 0.33372\\
3446 0.42992\\
3618 0.36944\\
3790 0.39919\\
3962 0.40658\\
4134 0.44549\\
};
    
\addplot [color=mycolor2, smooth, dashed, line width=\runtimeLineWidth]
  table[row sep=crcr]{%
174 0.0036565\\
346 0.0090635\\
518 0.01634\\
690 0.022509\\
863 0.025691\\
1035 0.039921\\
1207 0.05388\\
1379 0.077252\\
1551 0.082386\\
1724 0.096522\\
1896 0.096199\\
2068 0.10386\\
2240 0.12524\\
2412 0.12229\\
2585 0.15451\\
2757 0.16512\\
2929 0.14519\\
3101 0.13423\\
3273 0.21622\\
3446 0.23604\\
3618 0.25259\\
3790 0.28279\\
3962 0.2705\\
4134 0.32465\\
};
\addlegendentry{\textcolor{black}{Ours on $\mathcal{P}$}}
\end{axis}
\end{tikzpicture}%
    \end{tabular}
    \vspace{-0.3cm}
    \caption{\textbf{Runtime comparison} of the approach by \laehneretal \cite{lahner2016} and ours (on conjugate product graph $\conjProdGraph$ as well as product graph $\prodGraph$ for a fair comparison). The vertical axis shows the runtime in minutes. Points (light colours) are individual experiments, while thick lines are median runtimes. Spikes in computation time stem from a varying number of branches needed to compute the optimal path. \textbf{Left: } %
    We fix the size of various 3D shapes and gradually increase the number of vertices (horizontal axis) of respective 2D shapes (by subsampling). %
    \textbf{Right: } We fix the size of various 2D shapes and gradually increase the number of vertices (horizontal axis) of respective 3D shapes (also by subsampling). %
    }
    \label{fig:runtime-comparison}
\end{figure}

In \cref{fig:runtime-comparison}, we show the median runtime for the approach by \laehneretal \cite{lahner2016} and our approach. The plot shows that both approaches have the same asymptotic behaviour.
Due to the use of the \emph{larger} conjugate product graph $\conjProdGraph$ in comparison to product graph $\prodGraph$ (see also~\ref{sub:number_cpe}), our approach takes by a constant factor more time to compute results.
For a fair comparison with equal graph sizes, we additionally include computation times of our approach on the product graph $\prodGraph$ which shows the improved performance when using \cref{alg:b-n-b}.
Nevertheless, we emphasise that our approach (on $\conjProdGraph$) still requires polynomial time while being the only one that is able to compute 2D-3D matchings without the need for pre-matching.
\section{2D to 3D Deformation Transfer}
We compute 2D to 3D deformation transfer by applying the following steps:

\paragraph{2D-3D Matching} We find a matching between 2D and 3D shape using our approach.

\paragraph{2D Deformation} We deform the 2D shape by using a skeleton which allows for different articulation of arms, legs and head. In combination with biharmonic weights~\cite{wang2015linear,gptoolbox}, we obtain a smooth deformation of the 2D shape 
(we tessellate the interior of the contour for biharmonic weight computation \cite{persson2004simple}).

\paragraph{2D-3D Alignment} We find the optimal alignment $T_{2D}^{3D}$ of 2D shape and matched vertices on 3D shape by introducing a third, constant coordinate for 2D vertices and solving the (orthogonal) Procrustes problem \cite{ten1977orthogonal}.%

\paragraph{3D Deformation} We apply the deformation to the 3D shape by transforming the deformation on the 2D shape using $T_{2D}^{3D}$, applying the transformed deformation to a small subset of 3D vertices (chosen by furthest distance) and using their new positions as a constraint when deforming all other vertices of the 3D shape with the as-rigid-as-possible method of \cite{sorkine2007rigid}.

\section{Ablation Studies}
\subsection{Cost Function}
We evaluate the performance of different parts of our cost function in \cref{tab:ablation-study} as well as the performance of local rigidity when using multidimensional spectral features.
\begin{table}[h]
    \vspace{-0.2cm}
    \small
    \centering
    \begin{tabular}{lc}
      \toprule
        \textbf{Method} & \textbf{AUC} $\uparrow$\\
        \midrule
        Local Rigidity \& Spectral & 0.95\\
        Local Rigidity& 0.76\\
        Local Thickness & 0.92\\
        Local Rigid. \& Local Thick., ($\psi_1(x) = \psi_2(x) = |x|$) & 0.89\\
        Ours & \textbf{0.98}\\
      \bottomrule
    \end{tabular}
    \caption{\textbf{Various cost functions} on FAUST. %
    The score is the area under the curve (AUC) of the cumulative segmentation errors. All introduced components increase performance. Our one-dimensional local thickness outperforms the multi-dimensional spectral features due to different intrinsic properties of 2D and 3D shapes.}
    \label{tab:ablation-study}
\end{table}

\subsection{Discretisation}
In \cref{tab:ablation-study-discretisation} we evaluate the robustness of our method w.r.t. to different discretisations.
\begin{table}[h]
    \small
    \centering
    \setlength{\tabcolsep}{0pt}
    \begin{tabular}{rlc}
      \toprule
        \multicolumn{2}{c}{\textbf{Mean Edge Length 2D Shape} $\quad$} & \textbf{AUC} $\uparrow$\\
        \midrule
        $0.5\cdot$ & $\bar{e}$ & 0.96\\
        $0.75\cdot$ & $\bar{e}$ & 0.97\\
        $1\cdot$ &$\bar{e}$ & \textbf{0.98}\\
        $\;1.25\cdot$ & $\bar{e}$ & 0.97\\
        $1.5\cdot$ & $\bar{e}$ & 0.95\\
      \bottomrule
    \end{tabular}
    \caption{Ablation study on the sensitivity of our approach to \textbf{different discretisations}. The score is the area under the curve (AUC) of the cumulative segmentation errors. We fix the discretisation of 3D shape and vary edge lengths of 2D shape. $\bar{e}$ depicts the mean edge length of 3D shape.}
    \label{tab:ablation-study-discretisation}
\end{table}
For all our experiments in the main paper we reduce influence of discretisation by decimating 3D shapes to half of their original resolution, which results in more uniform edge lengths \cite{gptoolbox}.
Additionally, we re-sample 2D shapes with edge lengths according to the mean edge length of the decimated 3D shape.

\begin{figure}[h]
    \centering
    \newcommand{\pckNULineWidth}{4pt}
\newcommand{\pckNUplotWidth}{\columnwidth}
\newcommand{\pckNUplotHeight}{0.65\columnwidth}
\newcommand{\pckNUTitle}{FAUST: Mesh Discretisation}
\pgfplotsset{%
    every axis/.style={line width=0.01pt},
    label style = {font=\sffamily\large},
    tick label style = {font=\sffamily\large},
    title style =  {font=\Large\sffamily},
    legend style={  fill= gray!10,
                    fill opacity=0.6, 
                    font=\sffamily\large,
                    draw=gray!20, %
                    text opacity=1}
}
\begin{tikzpicture}[scale=0.5, transform shape]
	\begin{axis}[
		width=\pckNUplotWidth,
		height=\pckNUplotHeight,
		grid=major,
		title=\pckNUTitle,
		title style={yshift=-0.2cm},
		legend style={
			at={(0.97,0.03)},
			anchor=south east,
			legend columns=1},
		legend cell align={left},
		ylabel={{\sffamily\large $\%$ Correct Segment}},
        xlabel={Geodesic Error Threshold},
		xmin=0,
        xmax=1,
        ylabel near ticks,
        xtick={0, 0.25, 0.5, 0.75, 1},
		ymin=0,
        ymax=103,
        ytick={0, 20, 40, 60, 80, 100},
	]
	
    \addplot [color=mycolor1, dotted, smooth, line width=\pckNULineWidth]
          table[row sep=crcr]{%
0 72.0955\\
     0.01 73.2473\\
     0.02 75.7392\\
     0.03 77.8438\\
     0.04 79.4603\\
     0.05 80.9968\\
     0.06 82.3112\\
     0.07 83.5739\\
     0.08 84.8547\\
     0.09 86.0684\\
     0.1 87.2124\\
     0.11 88.1859\\
     0.12 88.7928\\
     0.13 89.2034\\
     0.14 89.4668\\
     0.15 89.5778\\
     0.16 89.6733\\
     0.17 89.7301\\
     0.18 89.8386\\
     0.19 90.0219\\
     0.2 90.1924\\
     0.21 90.3783\\
     0.22 90.4842\\
     0.23 90.5849\\
     0.24 90.7372\\
     0.25 90.8767\\
     0.26 91.1246\\
     0.27 91.3828\\
     0.28 91.6436\\
     0.29 91.9225\\
     0.3 92.2402\\
     0.31 92.6043\\
     0.32 92.9632\\
     0.33 93.2679\\
     0.34 93.5313\\
     0.35 93.836\\
     0.36 94.102\\
     0.37 94.4506\\
     0.38 94.7734\\
     0.39 95.1788\\
     0.4 95.6746\\
     0.41 96.0955\\
     0.42 96.5552\\
     0.43 97.0484\\
     0.44 97.4926\\
     0.45 97.8567\\
     0.46 98.2802\\
     0.47 98.6753\\
     0.48 99.0058\\
     0.49 99.2899\\
     0.5 99.4732\\
     0.51 99.6462\\
     0.52 99.7728\\
     0.53 99.8115\\
     0.54 99.8296\\
     0.55 99.8554\\
     0.56 99.8683\\
     0.57 99.8941\\
     0.58 99.9199\\
     0.59 99.9354\\
     0.6 99.9484\\
     0.61 99.9716\\
     0.62 99.9768\\
     0.63 99.9819\\
     0.64 99.9871\\
     0.65 100\\
     0.99 100\\
        };
        \addlegendentry{\textcolor{black}{Ours (non-uni.): 0.95}}
    
    \addplot [color=mycolor3, smooth, line width=\pckNULineWidth]
          table[row sep=crcr]{%
0 78.2217\\
0.01 79.3683\\
0.02 81.924\\
0.03 84.1805\\
0.04 86.0533\\
0.05 87.5812\\
0.06 88.9173\\
0.07 90.2306\\
0.08 91.5255\\
0.09 92.768\\
0.1 93.8985\\
0.11 94.8829\\
0.12 95.5863\\
0.13 96.0637\\
0.14 96.3583\\
0.15 96.5182\\
0.16 96.6141\\
0.17 96.7055\\
0.18 96.7922\\
0.19 96.8767\\
0.2 96.9795\\
0.21 97.0663\\
0.22 97.144\\
0.23 97.2125\\
0.24 97.2719\\
0.25 97.3244\\
0.26 97.4112\\
0.27 97.5117\\
0.28 97.5996\\
0.29 97.6521\\
0.3 97.7321\\
0.31 97.8074\\
0.32 97.9057\\
0.33 98.0381\\
0.34 98.1272\\
0.35 98.2163\\
0.36 98.319\\
0.37 98.4241\\
0.38 98.5977\\
0.39 98.6936\\
0.4 98.8717\\
0.41 98.9905\\
0.42 99.1344\\
0.43 99.2349\\
0.44 99.3354\\
0.45 99.477\\
0.46 99.5957\\
0.47 99.7145\\
0.48 99.8219\\
0.49 99.9064\\
0.5 99.9543\\
0.51 99.9817\\
0.52 99.9931\\
0.53 100\\
0.99 100\\
        };
        \addlegendentry{\textcolor{black}{Ours (uniform): 0.98}}
	\end{axis}
\end{tikzpicture}%
    \caption{Comparison of the matching performance of a \textbf{uniformly} sampled 3D shape (red line, edge lengths approx.~equal) vs.~a 3D shape with \textbf{non-uniform} density (blue line, regions with high curvature have smaller edge lengths) on the entire FAUST dataset. The vertical axis shows $\%$ of points in correct segment ($\uparrow$) while the horizontal axis shows the geodesic error threshold.}
    \label{fig:abla-pck-uni-vs-nonuni}
\end{figure}
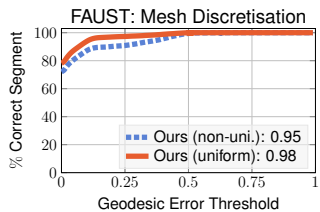
In \cref{fig:abla-pck-uni-vs-nonuni}, we compare the performance of our method w.r.t to different discretised 3D shapes.
We observe  that our method is relatively robust to different discretisation.

\begin{figure}[h]
    \centering
    \begin{tabular}{cc}
        \small
        \setlength{\tabcolsep}{0pt}
            \begin{tabular}{cccccc}
            \setlength{\tabcolsep}{0pt}
            & & \multicolumn{4}{c}{$|V_\contour|$}\\
            & \cellcolor{gray!10}& \cellcolor{gray!10}$39$ &\cellcolor{gray!10}$79$ & \cellcolor{gray!10}$159$& \cellcolor{gray!10}$318$\\
            \multirow{4}{*}{\rotatebox{90}{$|V_\mesh|$}}
            &\cellcolor{gray!10}$432$ &\colorbox[rgb]{0.377760, 0.599355, 0.299538}{$95.1$}  &\colorbox[rgb]{0.473182, 0.615925, 0.278397}{$87.8$}  &\colorbox[rgb]{0.363080, 0.596806, 0.302791}{$96.2$}  &\colorbox[rgb]{0.377113, 0.599243, 0.299682}{$95.1$} \\
            &\cellcolor{gray!10}$863$ &\colorbox[rgb]{0.806563, 0.673818, 0.204534}{$62.2$}  &\colorbox[rgb]{0.330376, 0.591127, 0.310036}{$98.7$}  &\colorbox[rgb]{0.360765, 0.596404, 0.303304}{$96.4$}  &\colorbox[rgb]{0.386007, 0.600787, 0.297711}{$94.5$} \\
            &\cellcolor{gray!10}$1724$ &\colorbox[rgb]{0.845785, 0.680629, 0.195844}{$59.2$}  &\colorbox[rgb]{0.560553, 0.631098, 0.259039}{$81.1$}  &\colorbox[rgb]{0.330114, 0.591081, 0.310094}{$98.7$}  &\colorbox[rgb]{0.340238, 0.592839, 0.307851}{$98.0$} \\
            &\cellcolor{gray!10}$3446\;$ &\colorbox[rgb]{0.838895, 0.679432, 0.197371}{$59.7$}  &\colorbox[rgb]{0.729401, 0.660418, 0.221630}{$68.1$}  &\colorbox[rgb]{0.357507, 0.595838, 0.304025}{$96.6$}  &\colorbox[rgb]{0.326784, 0.590503, 0.310832}{$99.0$} \\
            \end{tabular}
         &
         \hspace{-0.4cm}
         \def\subsamplFigHeight{2.5cm}
         \begin{tabular}{cc}
            \includegraphics[height=\subsamplFigHeight]{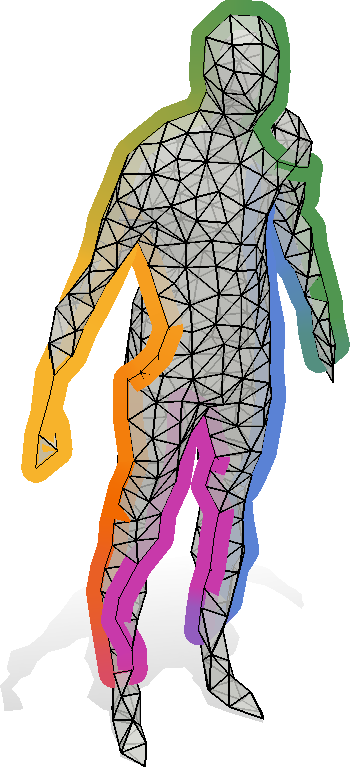} &
            \includegraphics[height=\subsamplFigHeight]{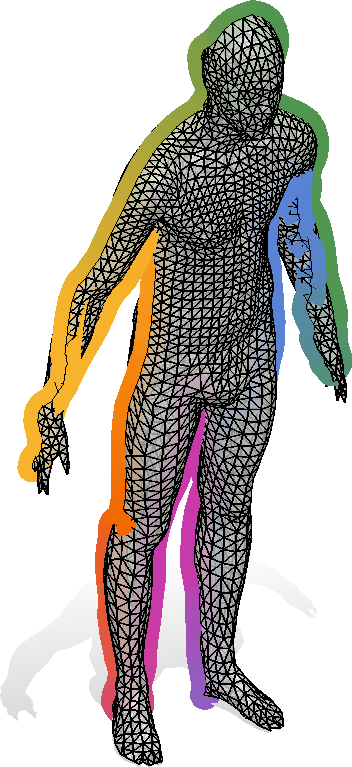}\\
         \end{tabular}
    \end{tabular}
    \caption{\textbf{Mesh resolution.} We downsample various 2D and 3D shapes (FAUST). The values with green and yellow background are AUC ($\uparrow$) of $\%$ of points in correct segment. The values with grey background are the number of vertices of respective 2D and 3D shapes. On the right we show that matchings from coarsest ($|V_\mesh|=432, |V_\contour|=39$) and finest resolution ($|V_\mesh|=3446, |V_\contour|=318$) are consistent.}
    \label{fig:abla-discretisation-table}
\end{figure}

In \cref{fig:abla-discretisation-table} we show results for matchings across different resolution of 2D and 3D shapes.
The experiment confirms that our method is robust in reasonable settings.
In addtion, we can see that matchings on the coarsest and on the finest level result in similar correspondences
(cf. shape visualisations on the right of \cref{fig:abla-discretisation-table}).

\subsection{Shape Discrepancies}
In \cref{fig:abl-wrong-surface}, we show results of cross-category matchings.
\def\qheight{1.6cm}
\def\qwidth{1.5cm}
\def\qhspaceCols{0cm}
\begin{figure}[!h]
    \centering
    \footnotesize
    \vspace{-0.2cm}
    \begin{tabular}{cccc}
        &
        \textcolor{gray!80}{$87.1$}&
        \textcolor{gray!80}{$118.5$}&
        \textcolor{gray!80}{$124.8$}\\
        \includegraphics[height=\qheight, width=\qwidth]{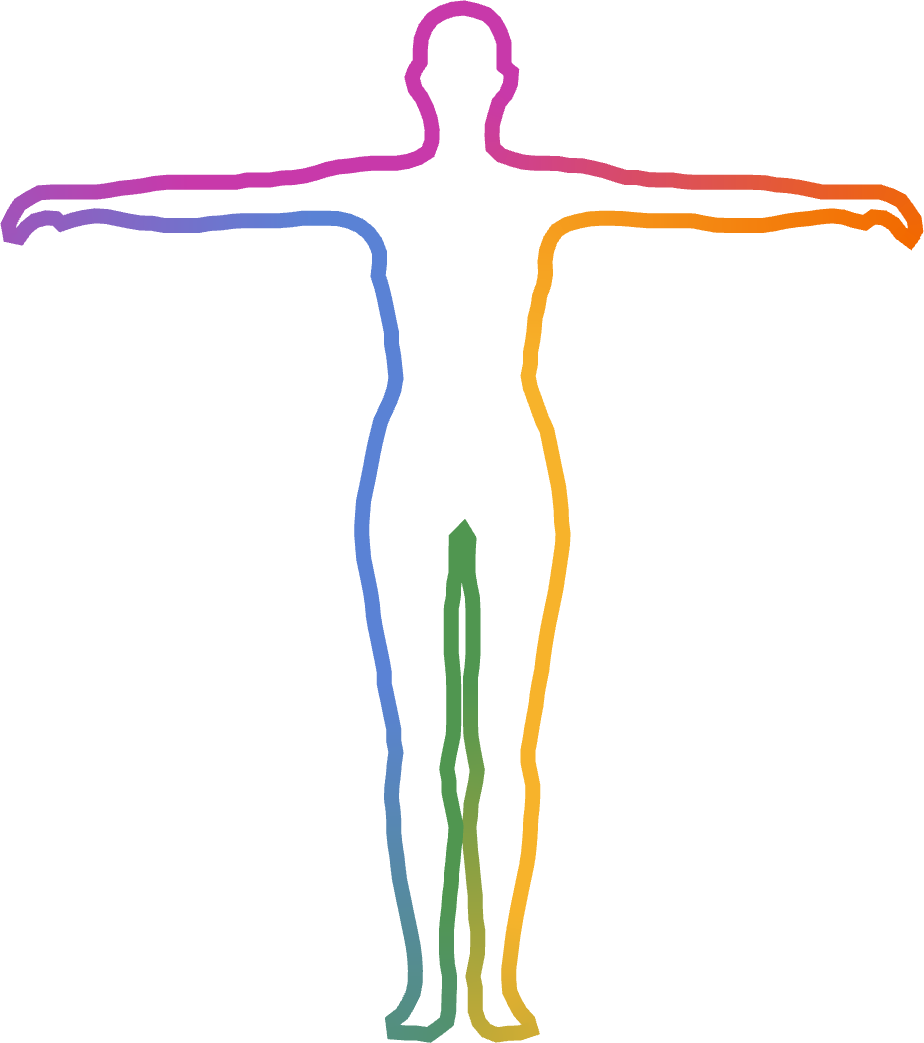}&
        \hspace{\qhspaceCols}
        \includegraphics[height=\qheight, width=\qwidth]{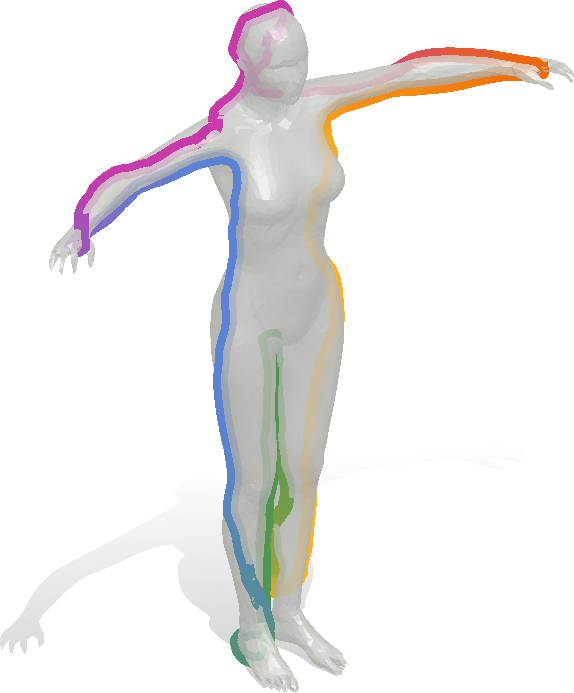}&
        \hspace{\qhspaceCols}
        \includegraphics[height=\qheight, width=\qwidth]{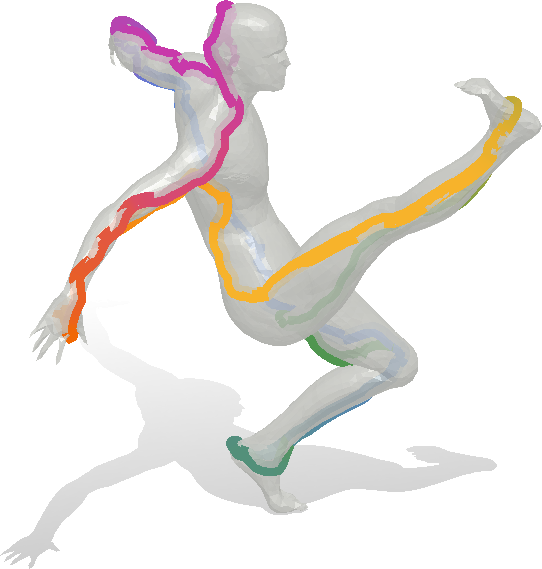}&
        \hspace{\qhspaceCols}
        \includegraphics[height=\qheight, width=\qwidth]{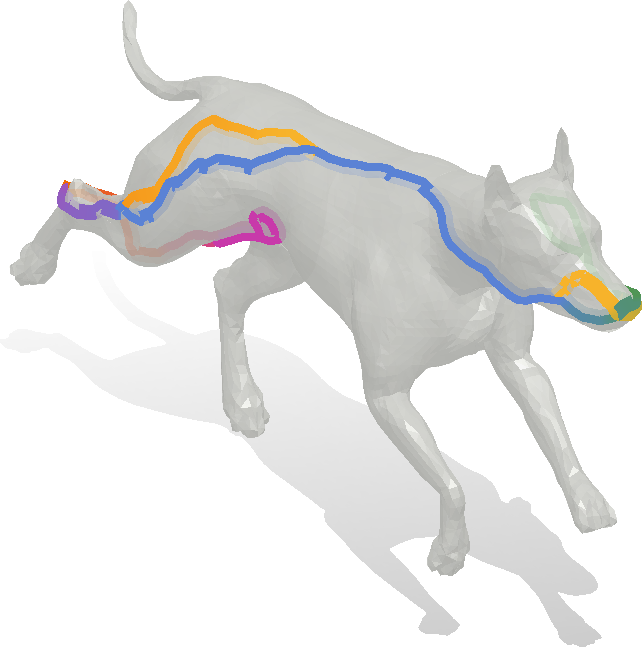}
    \end{tabular}
    \vspace{-0.2cm}
    \caption{
    \textbf{Cross-category matching.} We match the 2D shape of woman to the 3D shape of a woman and man, as well as to the 3D shape of dog. We can see that matching of woman to man results in a plausible matching, while matching of woman to dog does not yield meaningful results (as expected). The values are the resulting path-costs for each pair. %
    }
    \label{fig:abl-wrong-surface}
\end{figure}

\subsection{Noise}
In \cref{fig:abl-noise}, we evaluate our method's  robustness to noise when the 3D shape is disturbed by Gaussian noise.
We plot the AUC ($\uparrow$) of $\%$ of points in correct segment for each noise level.
\def\heightNoise{2.2cm}
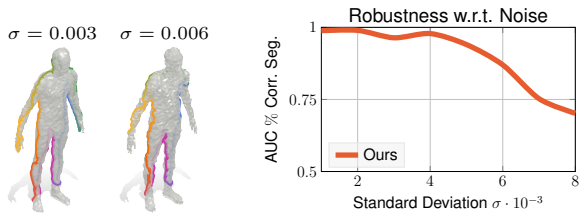
\begin{figure}[h]%
    \centering%
    \begin{tabular}{cc}
    \begin{overpic}[height=\heightNoise]
    {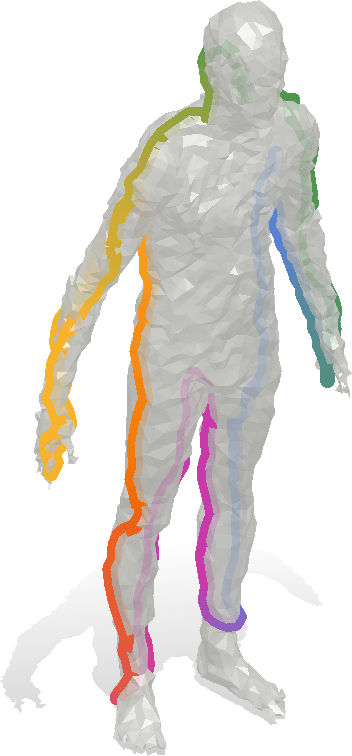}
    \put(0,105){\scriptsize$\sigma=0.003$}
    \end{overpic}
    \hspace{0.3cm}
    \begin{overpic}[height=\heightNoise]
    {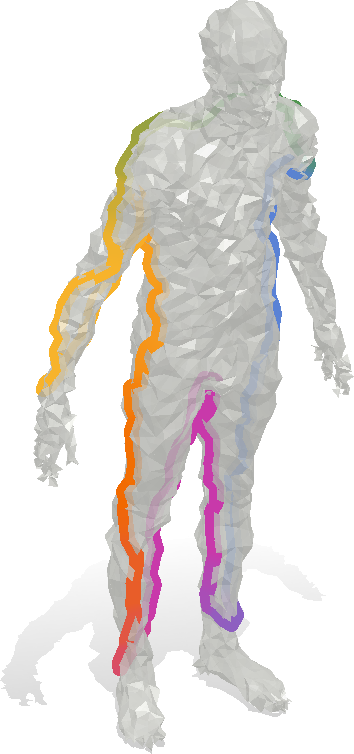}
    \put(0,105){\scriptsize$\sigma=0.006$}
    \end{overpic}
    &
    \newcommand{\discTwoDLineWidth}{4pt}
\newcommand{\discTwoDPlotWidth}{\columnwidth}
\newcommand{\discTwoDPlotHeight}{0.65\columnwidth}
\newcommand{\discTwoDTitle}{Robustness w.r.t. Noise}

\pgfplotsset{%
    every axis/.style={line width=0.01pt},
    label style = {font=\sffamily\large},
    tick label style = {font=\sffamily\large},
    title style =  {font=\Large\sffamily},
    legend style={  fill= gray!10,
                    fill opacity=0.6, 
                    font=\sffamily\large,
                    draw=gray!20, %
                    text opacity=1}
}
\begin{tikzpicture}[scale=0.5, transform shape]
	\begin{axis}[
		width=\discTwoDPlotWidth,
		height=\discTwoDPlotHeight,
		grid=major,
		title=\discTwoDTitle,
		title style={yshift=-0.2cm},
		legend style={
			at={(0.03,0.03)},
			anchor=south west,
			legend columns=1},
		legend cell align={left},
		ylabel={{\sffamily\large AUC $\%$ Corr. Seg.}},
        xlabel={Standard Deviation $\sigma\cdot10^{-3}$},
		xmin=0.001,
        xmax=0.008,
        ylabel near ticks,
        xtick={0.002, 0.004, 0.006, 0.008},
        xtick scale label code/.code={},
		ymin=0.5,
        ymax=1,
        ytick={0, 0.25, 0.5, 0.75, 1},
	]

    \addplot [color=mycolor3, smooth, line width=\discTwoDLineWidth]
          table[row sep=crcr]{%
        0.001 0.98844 \\
        0.002 0.98963 \\
        0.003 0.96387 \\
        0.004 0.97844 \\
        0.005 0.94036 \\
        0.006 0.86993 \\
        0.007 0.75262 \\
        0.008 0.70189 \\
        };
        \addlegendentry{\textcolor{black}{Ours}}

	\end{axis}
\end{tikzpicture}
    \end{tabular}
    \caption{\textbf{Robustness w.r.t noise.} We apply Gaussian noise with $\sigma=\{0.001,...,0.008\}$ to 3D shapes and subsequently match them to respective 2D shapes. Even under severe noise (see qualitative example on the left) our method is able to compute meaningful results which is confirmed by the plot on the right which shows AUC ($\uparrow$) of $\%$ of points in correct segment (vertical axis) for increasing standard deviations (horizontal axis).}
    \label{fig:abl-noise}
\end{figure}
\newpage
\section{Qualitative Results on FAUST}
\begin{figure}[ht!]
    \hspace{-0.6cm}
    \newcommand{\widthQualF}{2.3cm}
\newcommand{\heightQualF}{2.2cm}
\newcommand{\lrflip}{{\color{cPINK} \hspace{-0.1cm}$\nwarrow$\hspace{-0.3cm} }}
\begin{tabular}{c}
    \begin{tabular}{ccccc}%
        \includegraphics[height=\heightQualF,width=\widthQualF,keepaspectratio]{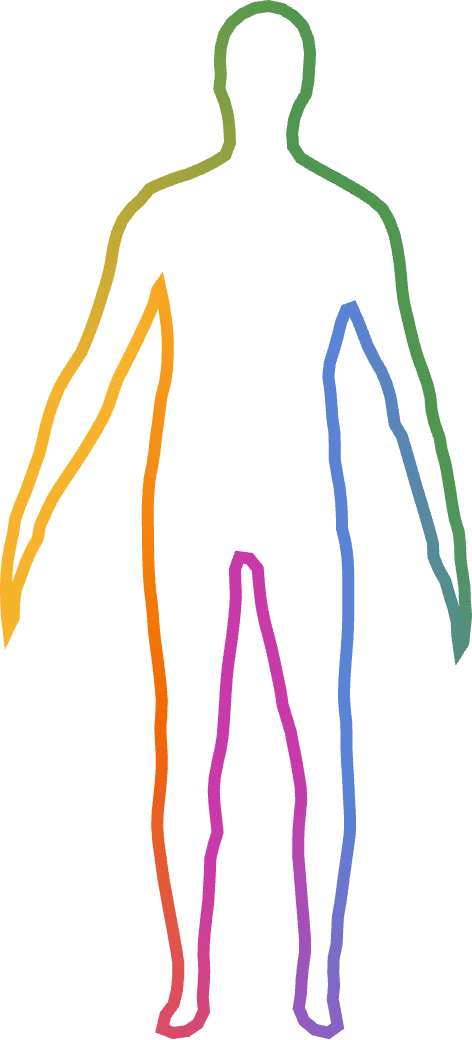} &
        \includegraphics[height=\heightQualF,width=\widthQualF,keepaspectratio]{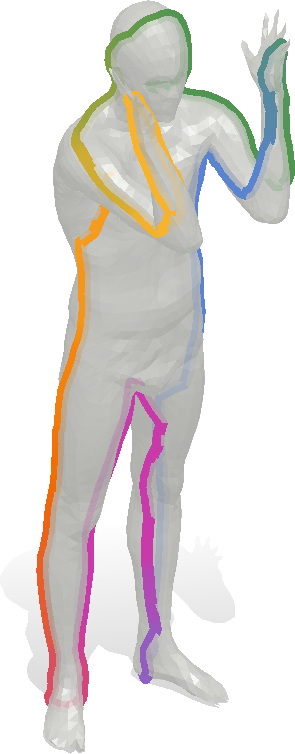} &
        \includegraphics[height=\heightQualF,width=\widthQualF,keepaspectratio]{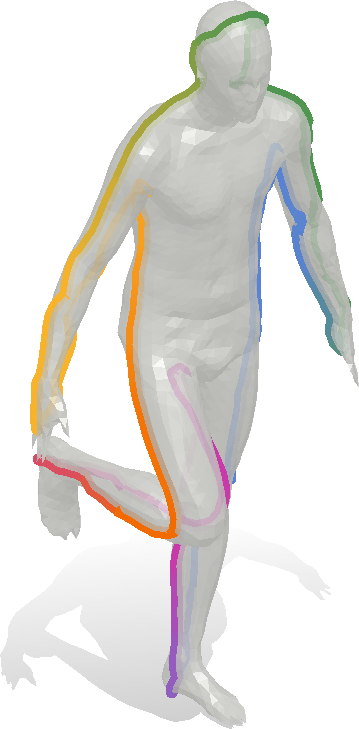} &
        \includegraphics[height=\heightQualF,width=\widthQualF,keepaspectratio]{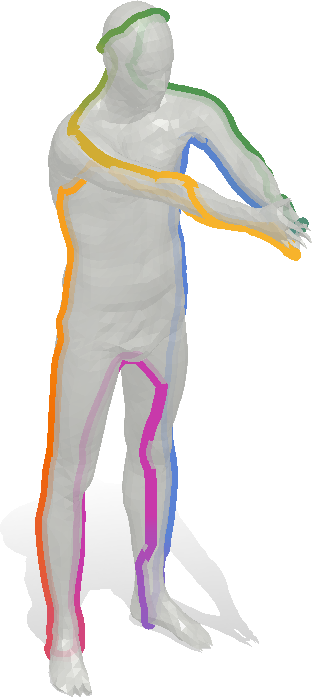} &
        \hspace{-0.8cm}
        \includegraphics[height=\heightQualF,width=\widthQualF,keepaspectratio]{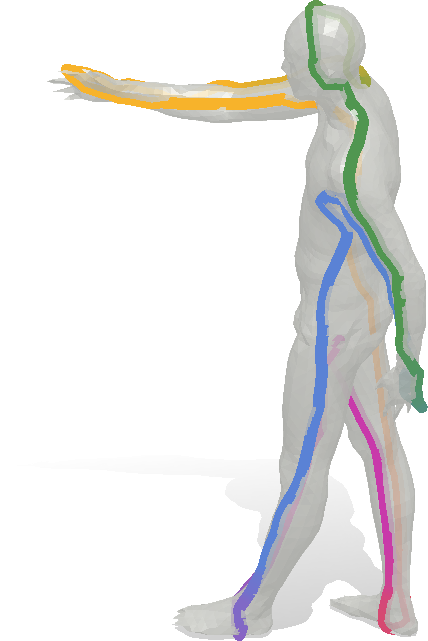} \\
        \includegraphics[height=\heightQualF,width=\widthQualF,keepaspectratio]{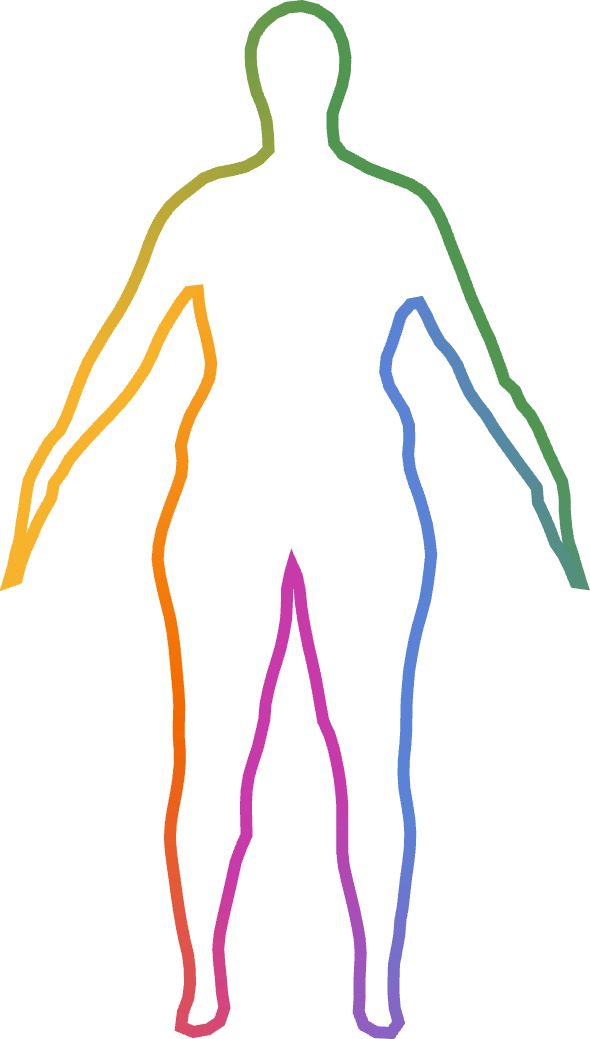} &
        \includegraphics[height=\heightQualF,width=\widthQualF,keepaspectratio]{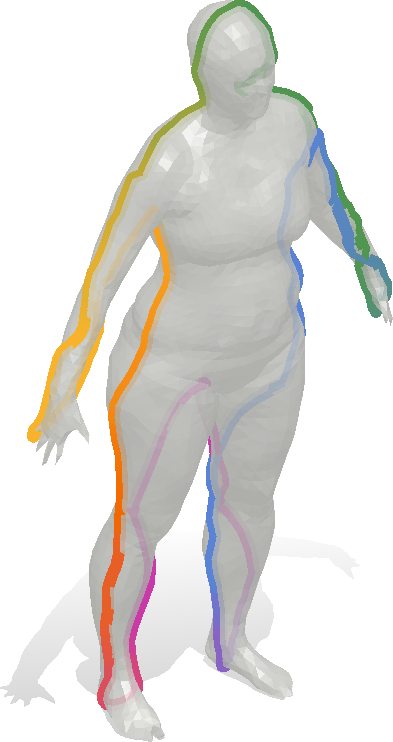} &
        \includegraphics[height=\heightQualF,width=\widthQualF,keepaspectratio]{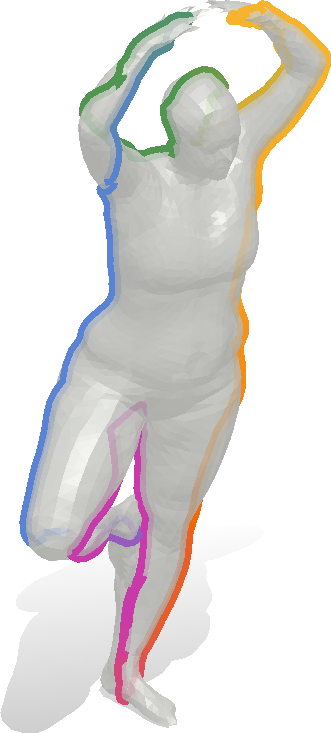}\lrflip&
        \includegraphics[trim={0 0 0 0.7cm}, height=\heightQualF,width=\widthQualF,keepaspectratio]{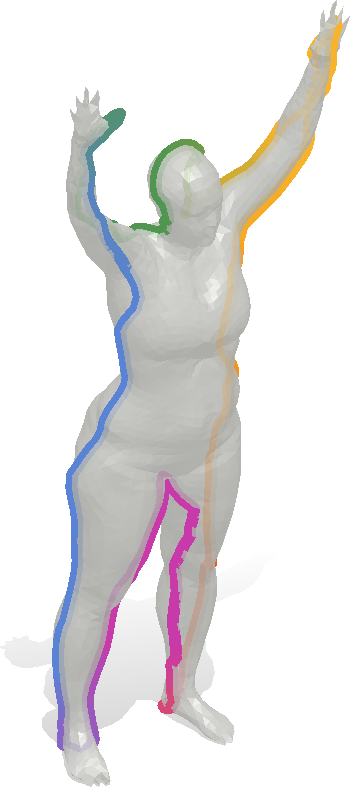}\lrflip &
        \includegraphics[height=\heightQualF,width=\widthQualF,keepaspectratio]{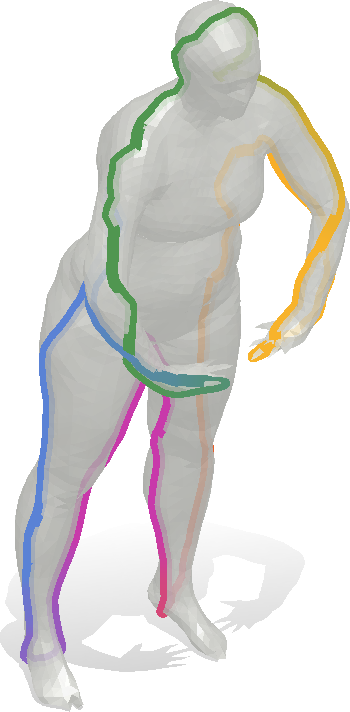}\lrflip\\
        \includegraphics[height=\heightQualF,width=\widthQualF,keepaspectratio]{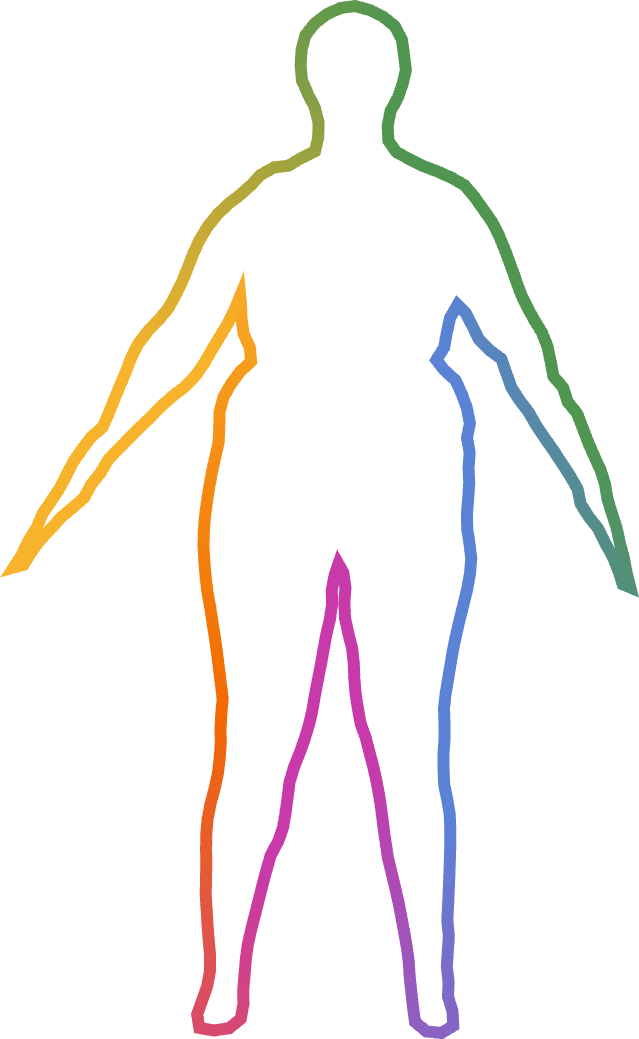} &
        \includegraphics[height=\heightQualF,width=\widthQualF,keepaspectratio]{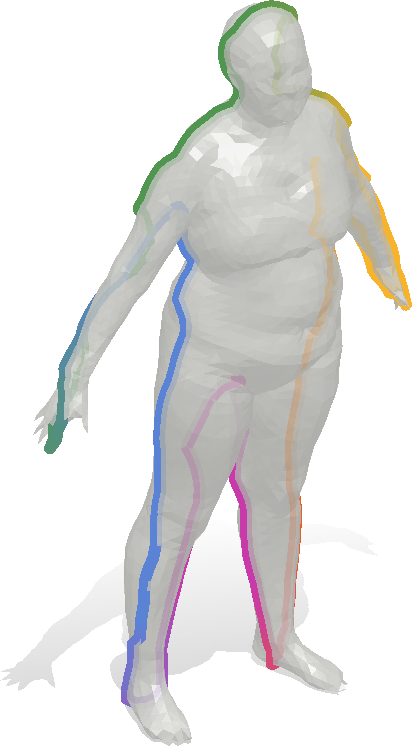}\lrflip &
        \includegraphics[height=\heightQualF,width=\widthQualF,keepaspectratio]{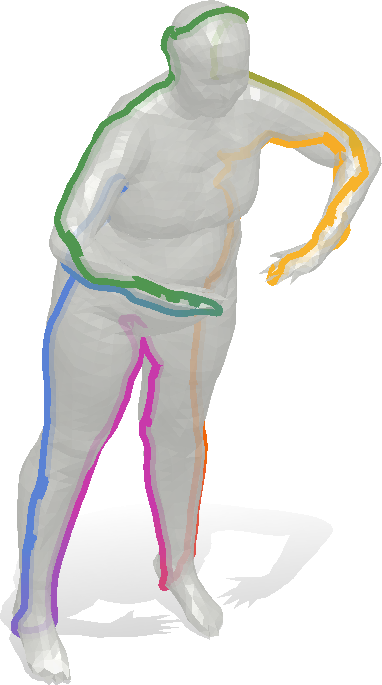}\lrflip&
        \includegraphics[trim={0 0 0 0.7cm}, height=\heightQualF,width=\widthQualF,keepaspectratio]{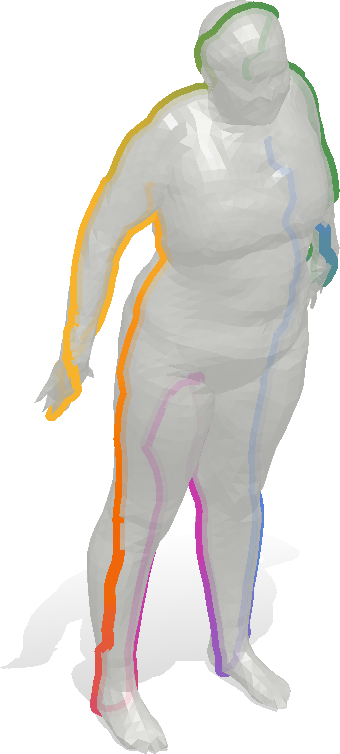} &
        \includegraphics[height=\heightQualF,width=\widthQualF,keepaspectratio]{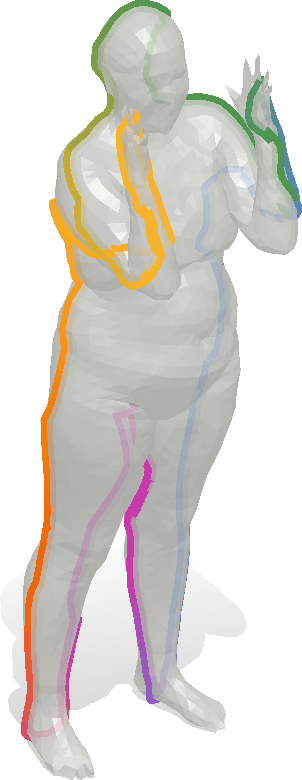}
    \end{tabular}\\ 
\end{tabular}
    \vspace{-0.3cm}
    \caption{\textbf{Qualitative results} on instances of FAUST dataset. We can see that left-right-flips occur (indicated with {\color{cPINK} $\nwarrow$}) which nevertheless are plausible matchings.}
    \label{fig:qual-faust-appendix}
\end{figure}

In \cref{fig:qual-faust-appendix} we show additional qualititve results.

\end{document}